%% file: main.tex
\documentclass[10pt]{article} % For LaTeX2e
\usepackage[preprint]{tmlr}

\input{math_commands.tex}

\usepackage{hyperref}
\usepackage{url}
\usepackage{graphicx}
\usepackage{subcaption}
\usepackage{wrapfig}

\title{On the special role of class-selective neurons in early training}

\author{\name Omkar Ranadive\thanks{Corresponding author} 
      \email omkar.ranadive@u.northwestern.edu \\
      \addr Alchera X \\
      \addr Northwestern University
      \AND
      \name Nikhil Thakurdesai 
      \email nikhilthakurdesai@gmail.com \\
      \addr Independent Researcher 
      \AND
      \name Ari S Morcos\thanks{Co-last authors}
      \email arimorcos@meta.com\\
      \addr Meta AI (FAIR)
      \AND 
      \name Matthew Leavitt\footnotemark[2]  
      \email matthew@mosaicml.com \\ 
      \addr MosaicML
      \AND 
      \name Stéphane Deny\footnotemark[2]
      \email stephane.deny@aalto.fi \\
      \addr Aalto University 
     }

\def\month{MM}  % Insert correct month for camera-ready version
\def\year{YYYY} % Insert correct year for camera-ready version
\def\openreview{\url{https://openreview.net/forum?id=XXXX}} % Insert correct link to OpenReview for camera-ready version

\begin{document}

\maketitle

\begin{abstract}

It is commonly observed that deep networks trained for classification exhibit class-selective neurons in their early and intermediate layers. Intriguingly, recent studies have shown that these class-selective neurons can be ablated without deteriorating network function. But if class-selective neurons are not necessary, why do they exist? We attempt to answer this question in a series of experiments on ResNet-50s trained on ImageNet. We first show that class-selective neurons emerge during the first few epochs of training, before receding rapidly but not completely; this suggests that class-selective neurons found in trained networks are in fact vestigial remains of early training. With single-neuron ablation experiments, we then show that class-selective neurons are important for network function in this early phase of training. We also observe that the network is close to a linear regime in this early phase; we thus speculate that class-selective neurons appear early in training as quasi-linear shortcut solutions to the classification task. Finally, in causal experiments where we regularize against class selectivity at different points in training, we show that the presence of class-selective neurons early in training is critical to the successful training of the network; in contrast, class-selective neurons can be suppressed later in training with little effect on final accuracy. It remains to be understood by which mechanism the presence of class-selective neurons in the early phase of training contributes to the successful training of networks.

\end{abstract}

\section{Introduction}
A significant body of research has attempted to understand the role of single neuron class-selectivity in the function of artificial \citep{zhou_object_2015, radford_sentiment_neuron_2017, bau_network_2017, s.2018on, olah_building_blocks_2018, rafegas_indexing_selectivity_2019, dalvi_grain_of_sand_2019, meyes_ablation_2019, dhamdhere_conductance_2019, leavitt2020selectivity, kanda_deleting_2020, leavitt_linking_2021}, and biological \citep{sherrington_integrative_1906, adrian_impulses_1926, granit_receptors_1955, hubel_receptive_1959, barlow_single_1972} neural networks. Neurons responding selectively to specific classes are typically found throughout networks trained for image classification, even in early and intermediate layers. Interestingly, these class-selective neurons can be ablated (i.e. their activation set to 0; \citealt{s.2018on}) or class selectivity substantially reduced via regularization \cite{leavitt2020selectivity} with little consequence to overall network accuracy—sometimes even improving it. These findings demonstrate that class selectivity is not necessary for network function, but it remains unknown \textit{why} class selectivity is learned if it is largely not necessary for network function.

\begin{figure}
     \centering
     \begin{subfigure}{0.55\linewidth}
         \centering
         \includegraphics[width=\linewidth]{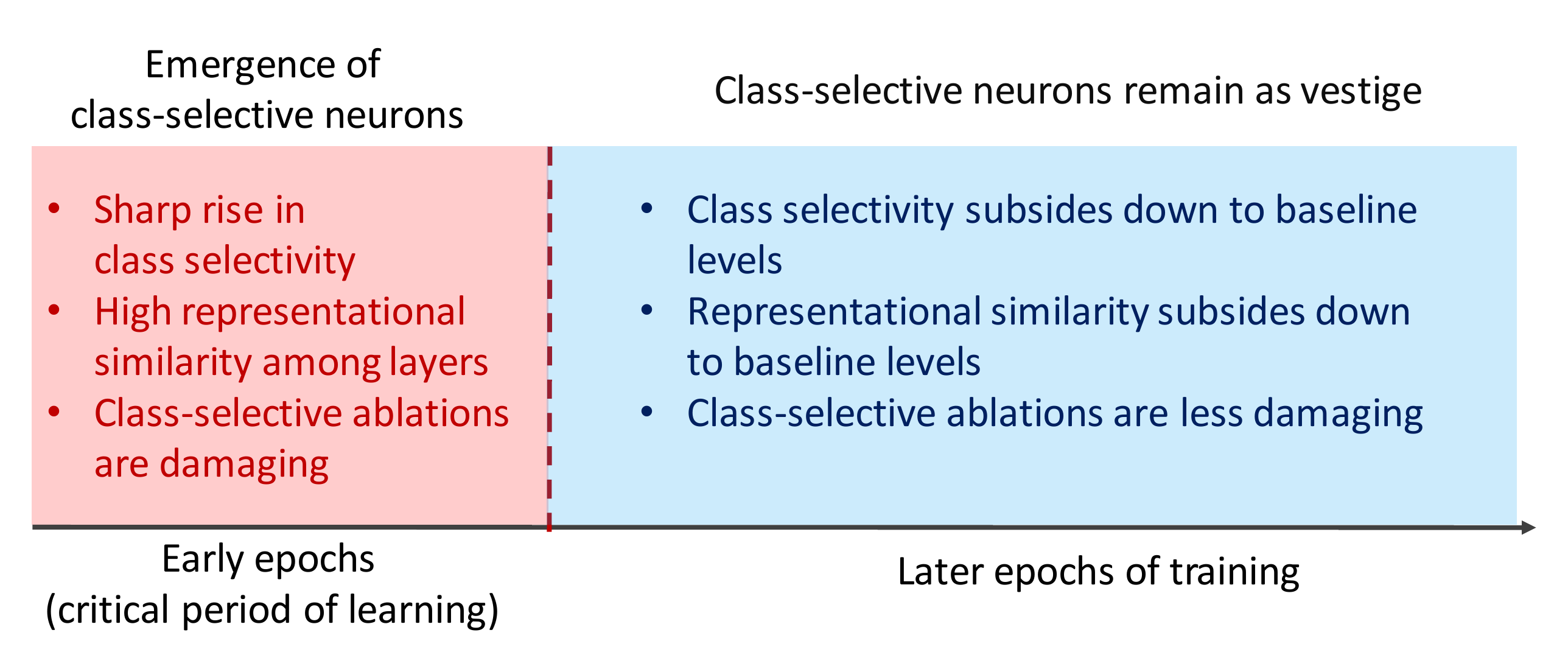}
         \caption{Dynamics of class selectivity throughout training}
         \label{resIntro1}
     \end{subfigure}
     %\vspace{1cm}
    \hspace{0.02\linewidth}
     \begin{subfigure}{0.4\linewidth}
        \centering
         \includegraphics[width=\linewidth]{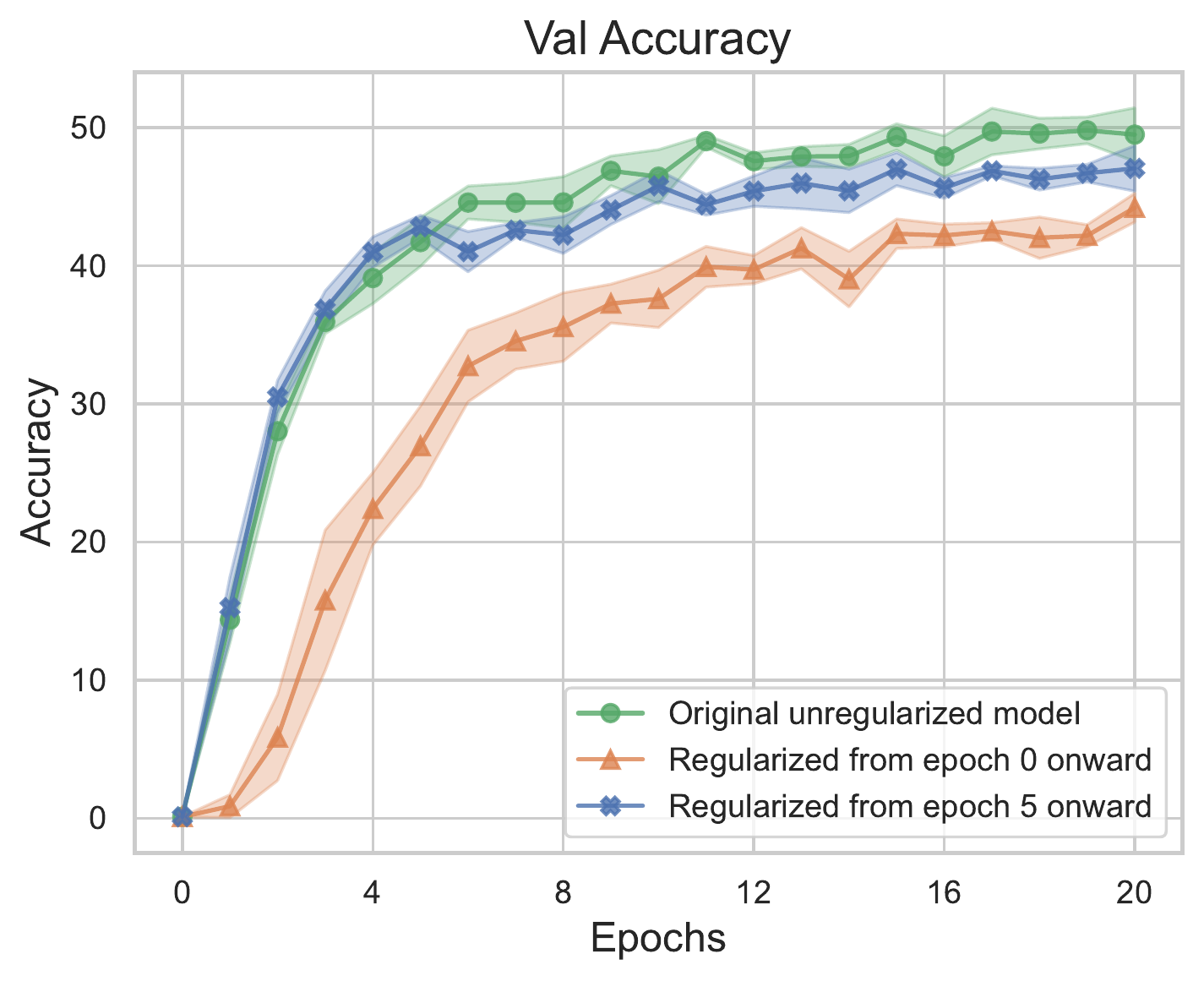}
         \caption{Suppressing selectivity in early epochs impairs learning}
         \label{resIntro2}
     \end{subfigure}     
     \caption{\textbf{Summary of results.} \textbf{(a)} Class-Selective neurons emerge during the early epochs of training possibly due to the high representational similarity among layers during the early epochs. In the later epochs, they remain as a vestige of their emergence during the early epochs. Ablating class-selective neurons during the early epochs is more damaging to the network accuracy compared to later epochs. \textbf{(b)} Model regularized against selectivity from epoch 0 onward performs significantly worse than model with no regularization, while model regularized against selectivity from epoch 5 onward is not substantially different from the model with no regularization. This indicates that class selectivity is important for learning during the early epochs of training. Shaded region denotes 95\% confidence interval of the mean for \emph{N} = 5 seeds.}   
     \label{res_sum}
\end{figure}

One notable limitation of many previous studies examining selectivity is that they have largely overlooked the temporal dimension of neural network training; single unit ablations are performed only at the end of training \citep{s.2018on, amjad_understanding_2018, zhou_revisiting_2018, meyes_ablation_2019, kanda_deleting_2020}, and selectivity regularization is mostly constant throughout training \citep{leavitt2020selectivity, leavitt_linking_2021}. However, there are numerous studies demonstrating substantial differences in training dynamics during the early vs. later phases of neural network training \citep{sagun_empirical_early_2018, gur-ari_gradient_2018, golatkar_time_early_2019, frankle_early_2020, jastrzebski_break-even_early_2020}. Motivated by these studies, we asked a series of questions about the dynamics of class selectivity during training in an attempt to elucidate \textit{why} neural networks learn class selectivity: When in training do class-selective neurons emerge? \textit{Where} in networks do class-selective neurons first emerge? Is class selectivity uniformly (ir)relevant for the entirety of training, or are there critical periods during which class selectivity impacts later network function? We addressed these questions in experiments conducted in ResNet-50 trained on ImageNet, which led to the following results (Fig \ref{res_sum}):

% Does the emergence of class-selective neurons play a role in the correct training of the network, or is it an epiphenomenon with little consequence to the final accuracy of the network?
\begin{itemize}
    \item The emergence of class-selective neurons in early and intermediate layers follow a non-trivial and surprising path: after a prominent rise during the first few epochs of training, class selectivity subsides quickly during the next few epochs, before returning to a baseline level specific to each layer.
    
    \item During this early training phase where average class selectivity is high in early and intermediate layers, class-selective neurons in these layers are much more important for network function compared to later in training, as assessed with single-unit ablation.
    
    \item During this early, high-selectivity phase of training, the representations of early and latter layers are much more similar than during later in training, implying that selectivity in early layers could be leveraged to solve the classification problem by transmission to the latter layers via skip connections.
    
    \item In a causal experiment where we prevent class selectivity from rising sharply in early and intermediate layers during the first epochs of training, we show the network training accuracy suffers from the suppression of this phenomenon. This indicates that the rapid emergence of class-selective neurons in early and intermediate layers during the first phase of training plays a useful role in the successful training of the network.
\end{itemize}
%Consistent with theoretical results of random networks (REF),

Together, our results demonstrate that class-selective neurons in early and intermediate layers of deep networks are a vestige of their emergence during the first few epochs of training, during which they play a useful role to the successful training of the network.
%We first reproduce this result. For this we used a ResNet50 pretrained on ImageNet. 

\section{Related Work}\label{RWork}

\subsection{The Role of Selectivity in Deep Networks}
Numerous studies have examined the causal role of class selectivity for network performance, nearly all of which have relied on single unit ablation as their method of choice. \citealt{s.2018on} examined a number of different CNN architectures trained to perform image classification and found that class selectivity for individual units was uncorrelated (or negatively correlated) with test-set generalization. This finding was replicated by \citealt{kanda_deleting_2020} and \citealt{amjad_understanding_2018}, though the latter study also observed that the effects can vary when ablating groups of neurons, in which case, selectivity can be beneficial. Furthermore, \citealt{zhou_revisiting_2018} found that ablating class-selective units can impair accuracy for specific classes, but a corresponding increase in accuracy for other classes can leave overall accuracy unaffected. 

Studies using NLP models have also shown varied results. \citealt{donnelly_interpretability_2019} ablated the "sentiment neuron" reported by \citealt{radford_sentiment_neuron_2017} and found mixed effects on performance, while \citealt{dalvi_grain_of_sand_2019} found networks were more negatively impacted when class-selective units were ablated.

Of particular importance is the work of \citealt{leavitt2020selectivity}, which introduced a regularizer for fine-grained control over the amount of selectivity learned by a network. This regularizer makes it possible to test whether the presence of selectivity is beneficial, and whether networks need to learn selectivity, two questions that ablation methods cannot test. By regularizing to discourage or promote the learning of selectivity, they found that selectivity is neither strictly necessary nor sufficient for network performance. In follow-up work they also showed that promoting class selectivity with their regularizer confers robustness to adversarial attacks, while reducing selectivity confers robustness to naturalistic perturbations \citep{leavitt_linking_2021}. However, they did not scale their experiments beyond ResNet18 \citep{he2016deep} and Tiny-ImageNet \citep{tiny_imagenet}. Additionally, with the exception of one set of controls in which the regularizer was linearly warmed up during the first five epochs of training \citep{leavitt2020selectivity}, they did not examine the dynamics of class selectivity's importance over the training process. Most importantly, they did not attempt to address \textit{why} selectivity is learned.

\subsection{The Early Phase of Neural Network Training}
A breadth of approaches have been used to characterize the differences between the early and later phases of neural network training and highlight the impact of early-phase interventions on late-phase behavior. The application of iterative magnitude pruning in the Lottery Ticket approach \citep{frankle_lottery_2018} requires rewinding to a sufficiently early point in training—within the first few thousand iterations—to prune without negatively impacting model quality \citep{frankle_linear_2020,frankle_early_2020}. The local loss landscape also changes rapidly early in training \citep{sagun_empirical_early_2018}; the subspace in which gradient descent occurs quickly shrinks into a very restricted subspace \citep{gur-ari_gradient_2018}. \citealt{achille_critical_2018} characterized a critical period of training by perturbing the training process with corrupted data labels early in training and found that the network's final performance was irreparably impaired. Similarly, \citealt{golatkar_time_early_2019} found that removing some forms of regularization after the early phase of training, or imposing them after the early phase, had little effect on network performance. These results emphasize the outsize importance of interventions applied during the early phase of neural network training, and the relevance of the early phase of training for understanding why networks learn class selectivity.

\begin{figure}[ht]
     \centering
     \begin{subfigure}{0.9\linewidth}
         \centering
         \includegraphics[width=\linewidth]{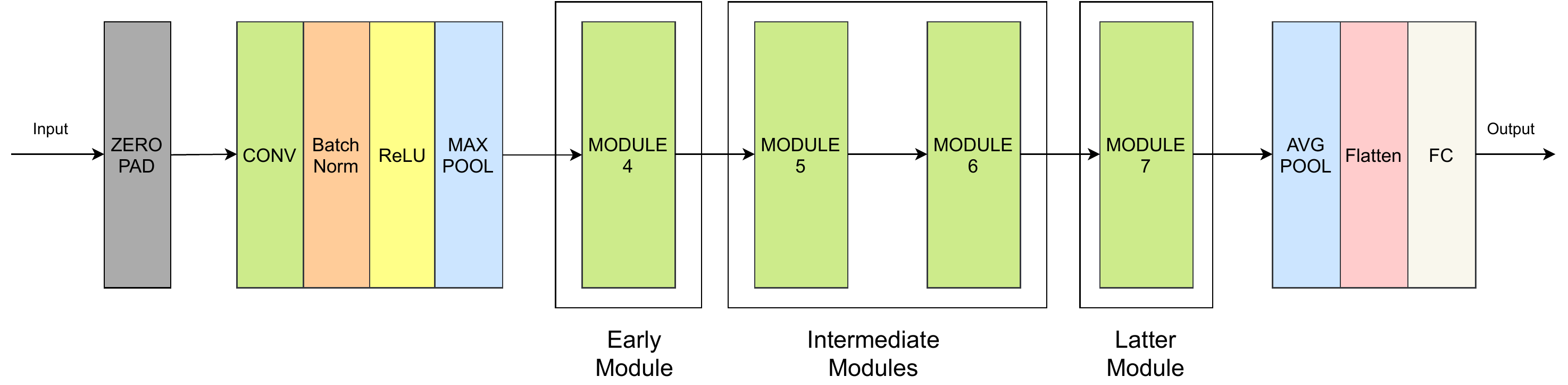}
         \caption{High-level modular design of a ResNet-50.}
         \label{r50_overview}
     \end{subfigure}
     \hfill
     %\vspace{1cm}
     \begin{subfigure}{0.49\linewidth}
        \centering
         \includegraphics[width=\linewidth]{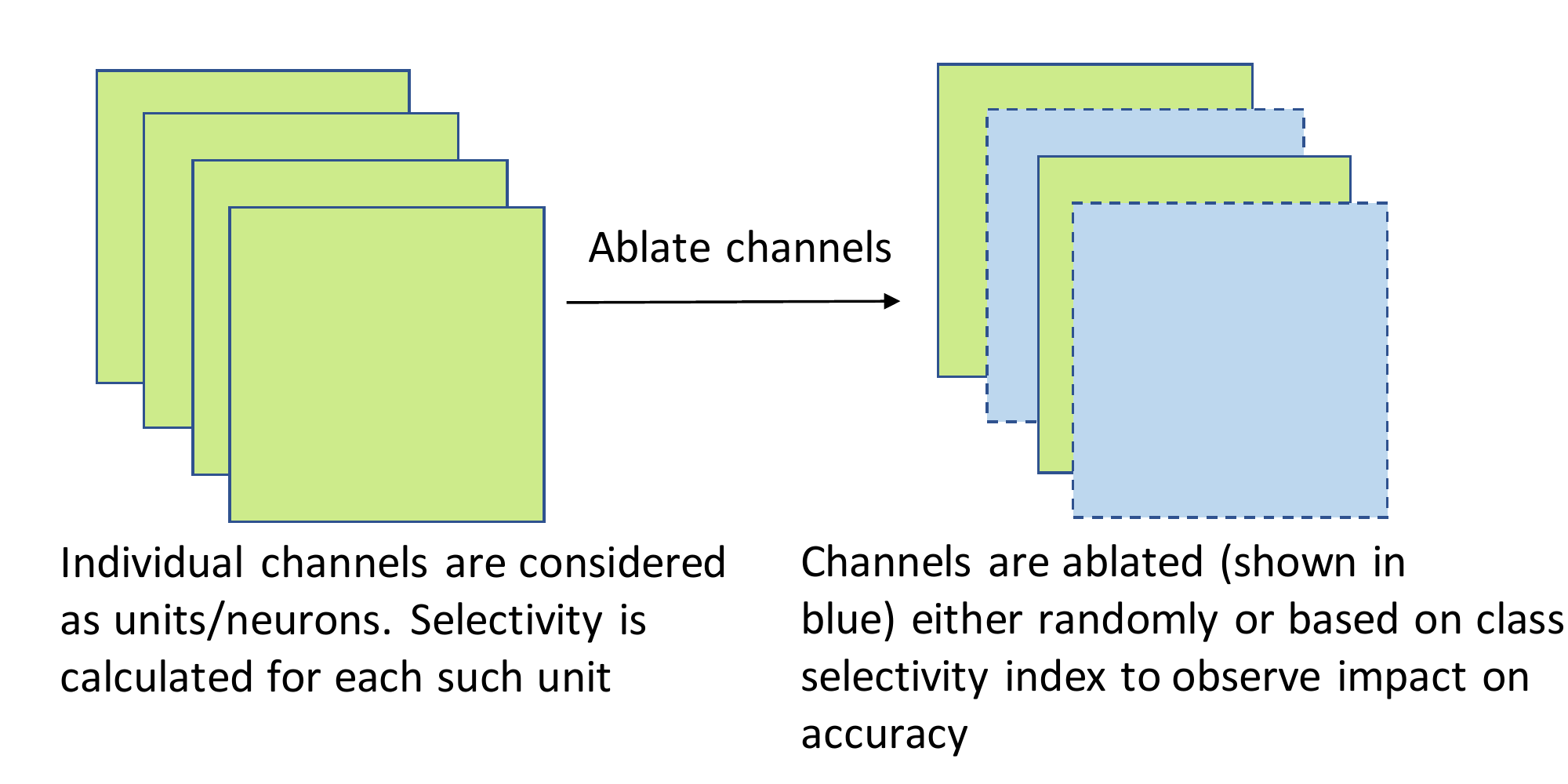}
         \caption{Class Selectivity Index}
         \label{csi}
     \end{subfigure}
     \hspace{0.02\linewidth}
     \begin{subfigure}{0.45\linewidth}
        \centering
         \includegraphics[width=\linewidth]{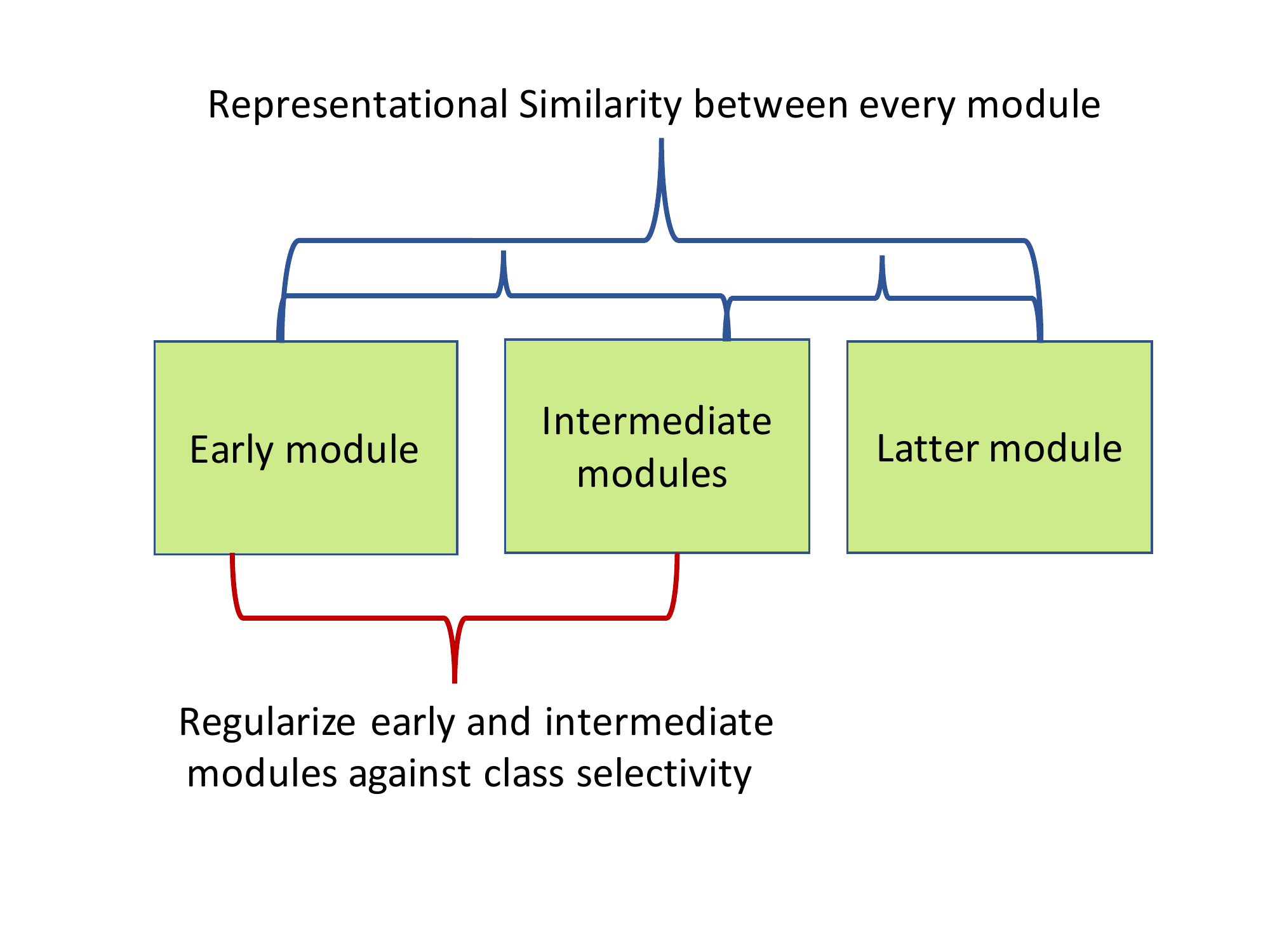}
         \caption{Regularization and Representational Similarity}
         \label{regsim}
     \end{subfigure}
     
     \caption{\textbf{Overview of our methods. } \textbf{(a)} We define the early and intermediate layers as the layers present in the early and intermediate modules of a ResNet-50. \textbf{(b)} Class Selectivity Index is calculated on individual channels of layers present inside each module. \textbf{(c)} We regularize early and intermediate modules against selectivity at different stages of training to observe impact on accuracy. Representational similarity is calculated between all modules throughout training to understand the emergence of class-selective neurons.}   
     \label{overview_methods}
\end{figure}

\section{Approach}
\subsection{Model Architecture}
We use ResNet-50s \citep{he2016deep} trained on ImageNet for all our experiments. ResNet-50s are formed of a convolutional stem followed by four successive modules. We define the early and intermediate layers as the layers present inside the early and intermediate modules shown in Fig \ref{r50_overview}. Each module is composed of multiple bottleneck layers. Each bottleneck layer contains one residual (or "skip") connection to the next bottleneck layer. Further details on the structure of the model can be found in Appendix \ref{appStruc}.  Details on model training can be found in Appendix \ref{appTrain}.

\subsection{Class Selectivity Index}
We refer to the individual channels of the bottleneck layers as "units" or "neurons" for the purpose of our experiments (Fig \ref{csi}). We compute a class selectivity index for each unit of each bottleneck layer of every module. This index is a measure of how much the unit favors a specific ImageNet class. It is defined as \citep{leavitt2020selectivity}:

\begin{equation}
    SI = \frac{\mu_{max} - \mu_{-max}}{\mu_{max} + \mu_{-max} + \epsilon}
\end{equation}

where $\mu_{max}$ is the largest class-conditional mean activation over all possible classes, averaged within samples of a class, and averaged over spatial locations for that channel, $\mu_{-max}$ is the mean response to the remaining (i.e. non-$\mu_{max}$) classes, and $\epsilon$ is a small value to prevent division by zero (here $10^{-6}$) in the case of a dead unit. All activations are assumed to be positive because they are extracted after a ReLU layer. The selectivity index can range from 0 to 1. A unit with identical average activity for all classes would have a selectivity of 0, and a unit that only responds to a single class would have a selectivity of 1. We compute such class selectivity indices across all epochs of training and then we perform various experiments to see how class selectivity affects learning at different stages of training.

\subsection{Quantifying the Relevance of Class-Selective Neurons}\label{mainAb}

We used the following methods to determine whether class-selective neurons are relevant to network function. 

\paragraph{Single neuron ablation:}This method consists in setting the channel outputs of the bottleneck layers to 0. In our experiments, we perform progressive ablation of the channels in all bottleneck layer of a given module, ordered by class selectivity index and starting with the most class-selective channels, or by random ordering of channels (random control condition). On a fully-trained ResNet-50, we found that ablation of class-selective neurons is less damaging to network accuracy than ablation of random neurons (Appendix \ref{appAB}), consistent with previous reports.

\paragraph{Class-selectivity regularizer:} We used \cite{leavitt2020selectivity}'s selectivity regularizer to either promote or suppress the emergence of class-selective neurons in different modules and at different epochs of training. The regularization term is as follows: 
\begin{equation}
    \mu_{SI} = \frac{1}{M}\sum_{m}^{M}\frac{1}{B}\sum_{b}^{B}SI_{b}
\end{equation}
where M is the total number of modules, m is the current module, B is the total number of bottleneck layer in current module M and b is the current bottleneck layer. $SI_{b}$ represents the selectivity indices of the channels in the current bottleneck layer b.

This term can be added to the loss function as follows: 
\begin{equation}
    loss = -\sum_{c}^{C}y_{c}\log(\hat{y_{c}}) - \alpha\mu_{SI}
\end{equation}

The left hand term of the loss function is the cross-entropy loss where C is the total number of classes, c is class index, $y_{c}$ is true class label and $\hat{y_{c}}$ is the predicted class probability. The right hand side of the loss function is the regularization term ($-\alpha\mu_{SI}$). The $\alpha$ parameter controls the strength of the regularizer. A negative value of $\alpha$ will regularize against selectivity and a positive value of $\alpha$ will regularize in favor of selectivity.

\section{Results}
\subsection{Class-selective neurons emerge during early epochs of training}

\begin{figure}{}
  \centering
  \includegraphics[width=0.45\linewidth]{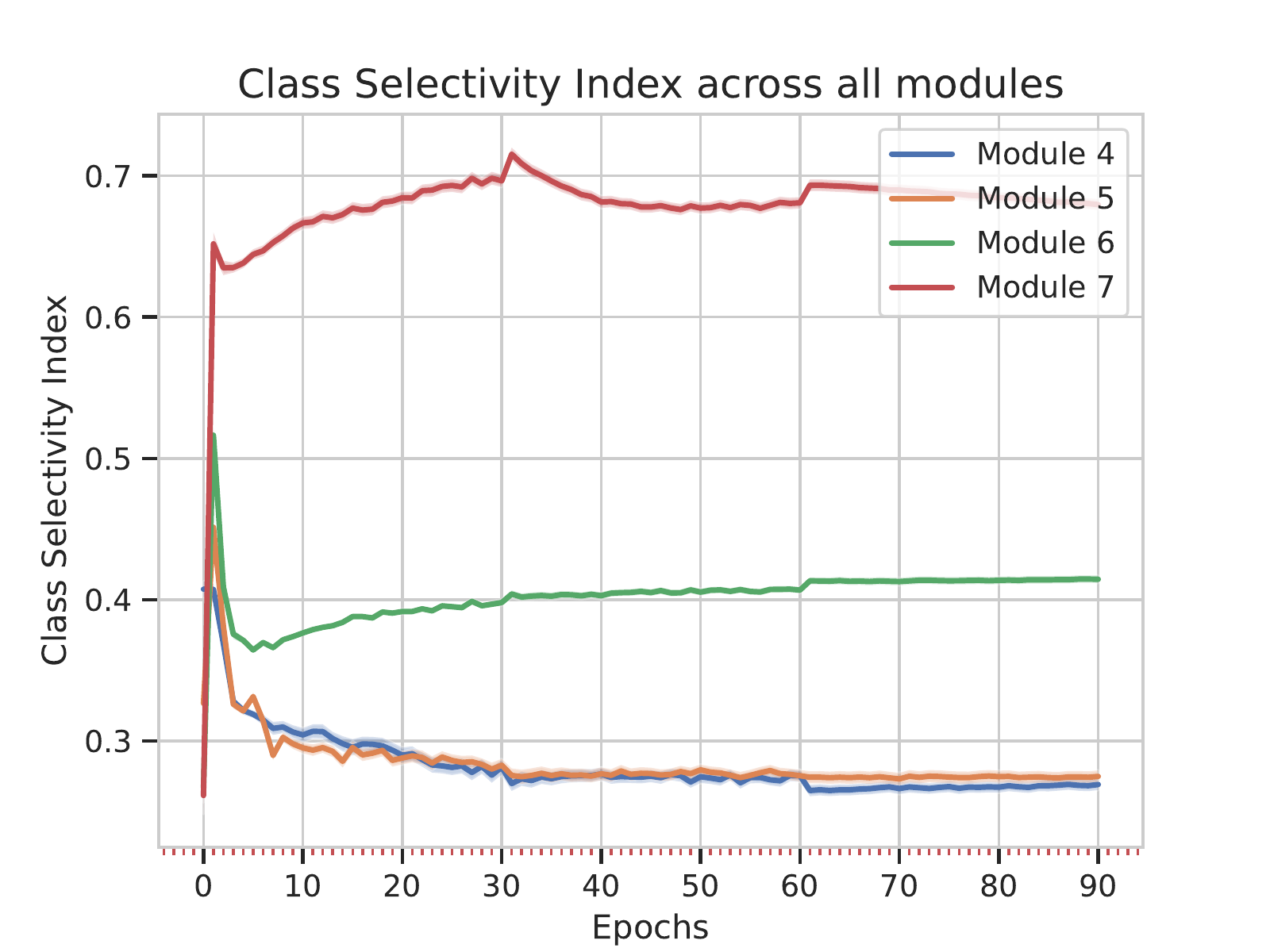}
  \caption{\textbf{Evolution of class selectivity throughout training in all modules of a ResNet-50.} Each line represents the evolution of average class selectivity of neurons of a given module throughout training for a ResNet-50 trained on ImageNet. A sharp rise in selectivity can be seen in all modules during the first few epochs of training, which recedes rapidly for the intermediate modules (modules 4, 5, 6) but not for the last module (module 7).}
  \label{avg_selectivity}
\end{figure}

We began by examining the dynamics of class selectivity in each module over the course of training. It is possible that selectivity follows a similar pattern as accuracy: increasing quickly over the first few epochs, then gradually saturating over the remainder of training. We consider this to be the most likely scenario, as it is follows the dynamics of early and rapid change followed by stabilization observed in many other quantities during training. Indeed, class selectivity sharply rises during the first epoch of training (Fig \ref{avg_selectivity}; analysis of evolution of selectivity throughout training for individual bottleneck layers can be found in Appendix \ref{appBott}), but after that, the effect varies depending on layer depth: earlier layers (Modules 4-6) show rapid decreases in selectivity over 3-5 epochs, followed by slow approach to an asymptote over the remainder of training. In contrast, latter layers (Module 7) do not exhibit any substantial decrease in selectivity after the initial jump, and selectivity remains high in latter layers throughout training. These dynamic, depth-dependent changes in selectivity during the first few epochs of training indicate that the role of selectivity could also vary during training, and in a layer-dependent manner.

% \begin{wrapfigure}{}{0.5\linewidth}
% \vspace{-5mm}
% % \begin{figure}
% %\begin{figure}
%   \centering
%   \includegraphics[width=\linewidth]{Figures/selectivity_over_cps_avg.pdf}
%   \caption{\textbf{Evolution of class selectivity throughout training in all modules of a ResNet-50.} Each line represents the evolution of average class selectivity of neurons of a given module throughout training for a ResNet-50 trained on ImageNet. A sharp rise in selectivity can be seen in all modules during the first few epochs of training, which recedes rapidly for the intermediate modules (modules 4, 5, 6) but not for the last module (module 7).}
%   \label{avg_selectivity}
% %\end{figure}
% \vspace{-1mm}
% \end{wrapfigure}

\subsection{Class-selective neurons support the decision of the network more during early epochs of training than latter epochs} 

\begin{figure}
  \centering
  \includegraphics[width=0.6\linewidth]{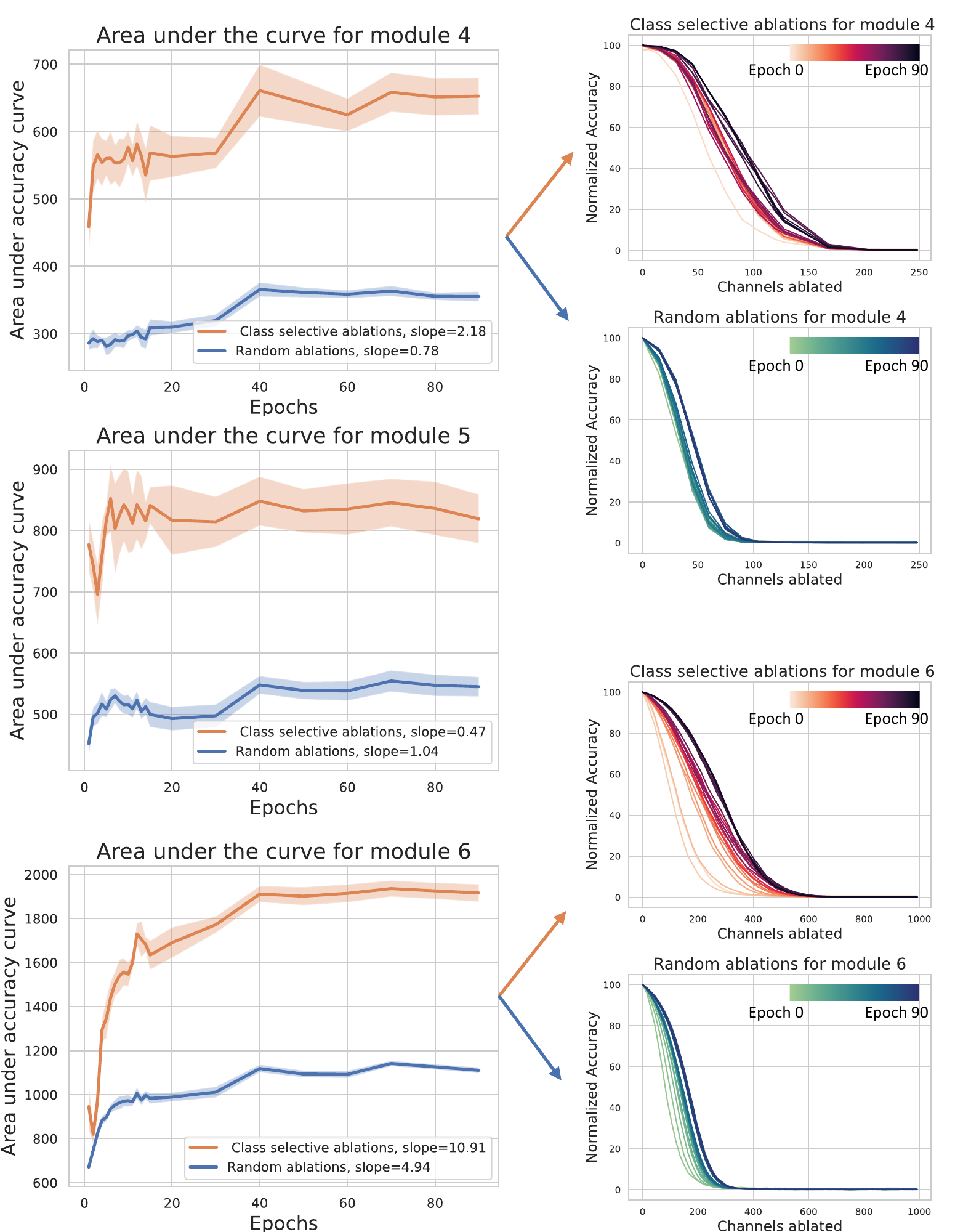}
  \caption{\textbf{Effect of class-selective ablations and random ablations on normalized network accuracy throughout training for early and intermediate modules.} The right-hand side of the figure depicts the effect of class-selective neurons ablations on network accuracy throughout epochs. The effect on accuracy is more damaging during the earlier epochs of class-selective ablations compared to latter epochs. The random ablation curves do not show this effect (curves are close to each other). The left-hand side of the figure depicts the area under the ablation curves (AUC), as calculated by the sum of accuracies for each epoch-specific curve shown on the right-hand side, summed over ablations. The positive slopes of the AUC curves for the class-selective ablations, especially visible during the early epochs of training, indicate that class-selective ablations are more important for network decision during these early epochs compared to later epochs. Error shades indicate the 95\% CI of the mean, calculated over 10 networks trained from different random initial conditions.}
  \label{auc}
  \medskip 
  \centering
  \includegraphics[width=0.8\linewidth]{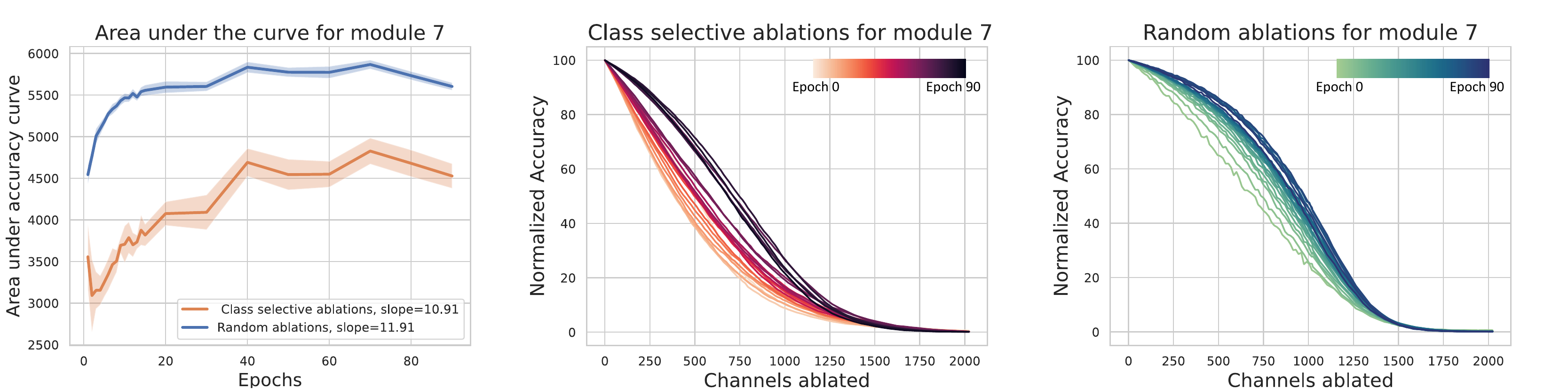}
  \caption{\textbf{Effect of class-selective ablations on the latter module.} The effect of class-selective ablations is different in module 7 than in the intermediate modules: class-selective neurons remain important throughout training, and not just in the early epochs. Error shades indicate the 95\% CI of the mean, calculated over 10 networks trained from different random initial conditions.}
  \label{auc2}

\end{figure}

In order to determine whether the role of class selectivity in network behavior varies over the course of training, we performed class-selective and random ablations (Section \ref{mainAb}) at different points during training and measured the effect on accuracy. We normalize accuracy between 0 and 100 so that the curves for each epoch can be plotted together and compared. The right-hand side of Fig \ref{auc} shows that the effect on accuracy is more prominent in the earlier epochs of class-selective ablations. Compared to this, random ablation curves do not show any significant difference across epochs. 

To quantify this phenomenon further, we calculated the area under the ablation curves for each epoch, obtaining one value per epoch summarizing the sensitivity of the network to ablation, and plotted these values across epochs. If class-selective ablations were more damaging in the earlier epochs, then this would correspond to less area under the curve (AUC) in the earlier epochs compared to the latter epochs i.e., the curve should show a positive slope across epochs. As illustrated in Fig \ref{auc}, the class-selective curves for module 4 and module 6 indeed show a positive slope which is steeper especially at early epochs compared to the random ablation curve which is relatively flat.  These results indicate that class-selective neurons are more important in the earlier epochs of training compared to the latter epochs as they are more damaging to the normalized accuracy in the earlier epochs compared to latter epochs. Interestingly, module 5 doesn't exhibit this trend as clearly and the effect is much stronger in module 6.
%\ari{effect is much stronger for module 6, I might emphasize that one more}
%\textit{weakly}

However, the AUC curves in Fig \ref{auc} also show that random ablations remain more damaging than class-selective ablations at all epochs, as they exhibit overall less area under the curve. The important distinction is that class-selective ablations are more damaging in the earlier epochs \textit{relative to the class-selective ablations in latter epochs} while random ablations are roughly equally damaging throughout training. This phenomenon reverses in module 7 (Fig \ref{auc2}), where class-selective ablations are more damaging throughout training, which is expected from the role of these layers to provide class-selective information for the network decision. %which is consistent with our previous findings. 

\begin{figure}
  \centering
  \includegraphics[width=0.8\linewidth]{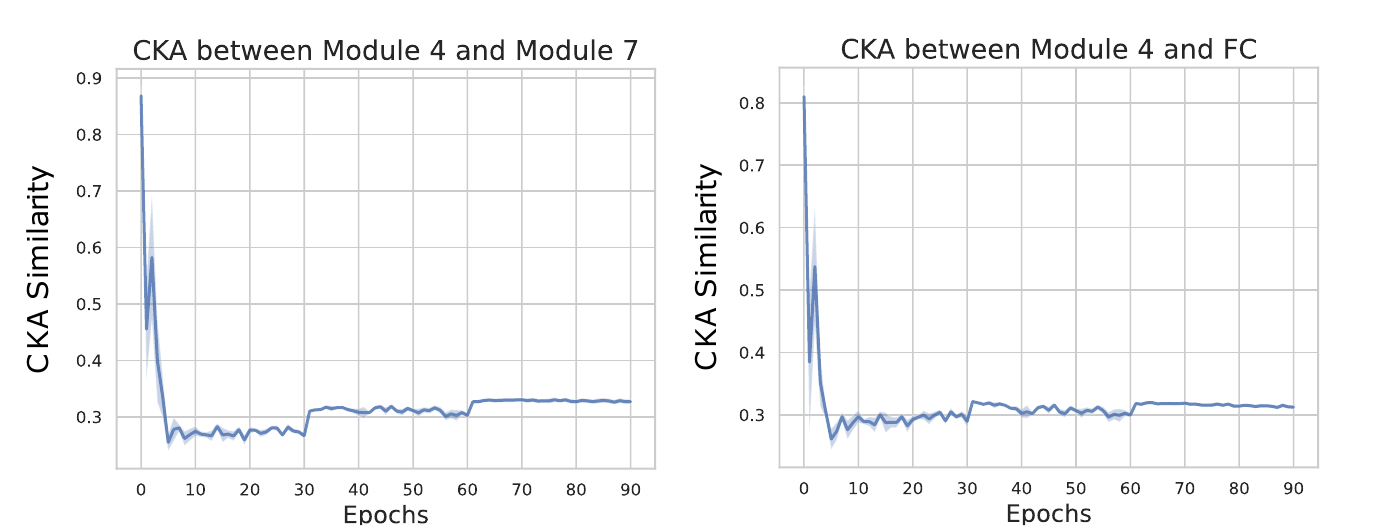}
  \caption{\textbf{Representational similarity between modules throughout training.} We measure the representational similarity between the early module 4 and latter layers (module 7 layers, fully connected layer (FC)) using CKA. We find that the similarity between these layers is high at the early epochs of training, indicating that the network is close to linear at this stage. This linearity might be key to explain the emergence of class-selective neurons in all layers during the early phase of training (see text). Error shades indicate the 95\% CI of the mean, calculated over 6 networks trained from different random initial conditions.}
  \label{cka}
\end{figure}
%which might explain the emergence of class-selective neurons in the early epochs.

\subsection{The network is linear at initialization and in early epochs of training, which might explain the emergence of shortcut decision strategies}

Why are neurons learning class-selective features in early and intermediate modules during the early epochs of training? Here, we postulated that the network is close to being linear at random weight initialization, such that the neurons of all layers jointly learn to be class-selective, with neurons from latter layers relying on class-selective features emerging in early and intermediate layers, effectively producing crude shortcut decision strategies at these early epochs of training. 
%the information from the latter layers making them more class-selective i.e, if their representations are similar to latter layers then they would be class-selective like the latter layers. We know that layers at the end (i.e, module 7 and fully connected layer (FC)) are highly class-selective as class selectivity is important near the end of the network for prediction. 

To test out this hypothesis, we calculated the Centered Kernel Alignment (CKA) \citep{kornblith2019similarity, nguyen2020wide} similarity between all modules. CKA can be used to find representational similarity between different components of a network. The CKA similarity between early module and latter module and between early module and the fully connected layer (FC) is indeed high during the first few epochs, and then recedes to a low level for the rest of the training (Fig \ref{cka}). The CKA similarity across all other pairs of modules shows a similar trend (Appendix \ref{appCKA}). 

From these results, the following picture emerges: at the early stages of training where the network is close to being linear, all layers jointly learn class-selective features, with latter layers piggy-backing on class-selective features learned by early layers.
%which might explain the emergence and significance of class-selective neurons at this early stage.

\subsection{The emergence of class-selective neurons early in training is essential to successful training: suppressing it impairs learning}\label{mainReg}

Is the emergence of class-selective neurons early in training crucial to learning? To test whether the spike observed in selectivity during the early epochs (Fig \ref{avg_selectivity}) is important to learning, we conducted a series of regularization experiments. The question which we want to answer is - Will the model be able to learn if we regularize against selectivity and suppress that spike in selectivity in the early epochs? 

%\cite{leavitt2020selectivity} had conducted regularization experiments showing that regularizing against selectivity doesn't impair learning. However, they had only tested this on the trained model and not across different epochs of training. Also, they had regularized the entire network, while we only regularize the early and intermediate modules.  

We performed experiments to test two scenarios: regularizing against class-selectivity from epoch 0 onward VS regularizing from epoch 5 onward. Epoch 0 is the randomly initialized model and epoch 5 is approximately after the spike in selectivity (Fig \ref{avg_selectivity}). We hypothesized that if selectivity is important to learning in those early epochs, then regularizing from epoch 0 onward should lead to a greater decrease in the model accuracy later on in training as compared to regularizing from epoch 5 onward. As we already know that selectivity is important to learning in the latter module and as we are interested in understanding the emergence of class-selective neurons in early and intermediate modules, we only regularized the early and intermediate modules against selectivity. 

\begin{figure}[t]
     \centering
     \begin{subfigure}{\linewidth}
        \centering
         \includegraphics[width=0.32\linewidth]{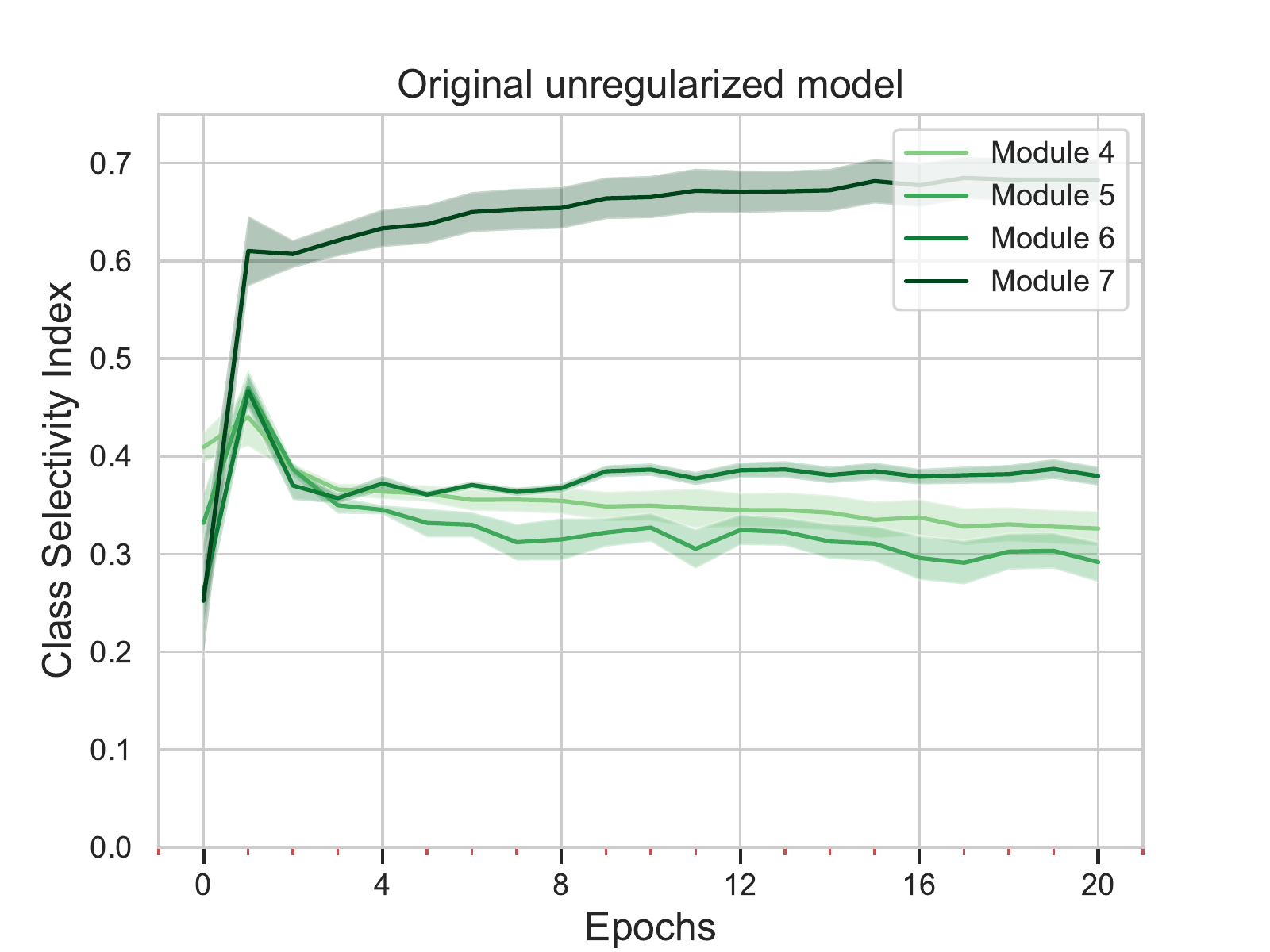}
         \includegraphics[width=0.32\linewidth]{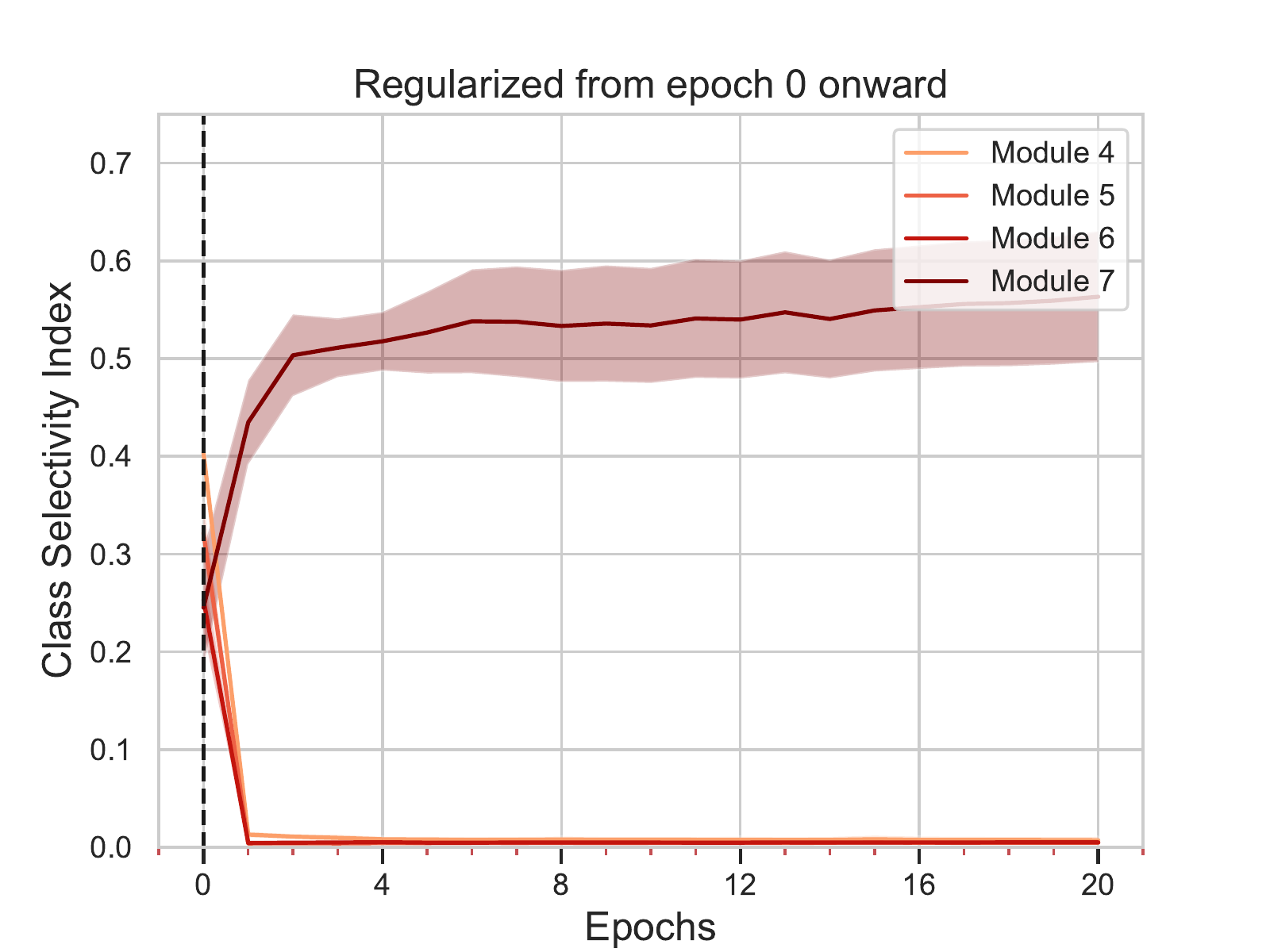}
         \includegraphics[width=0.32\linewidth]{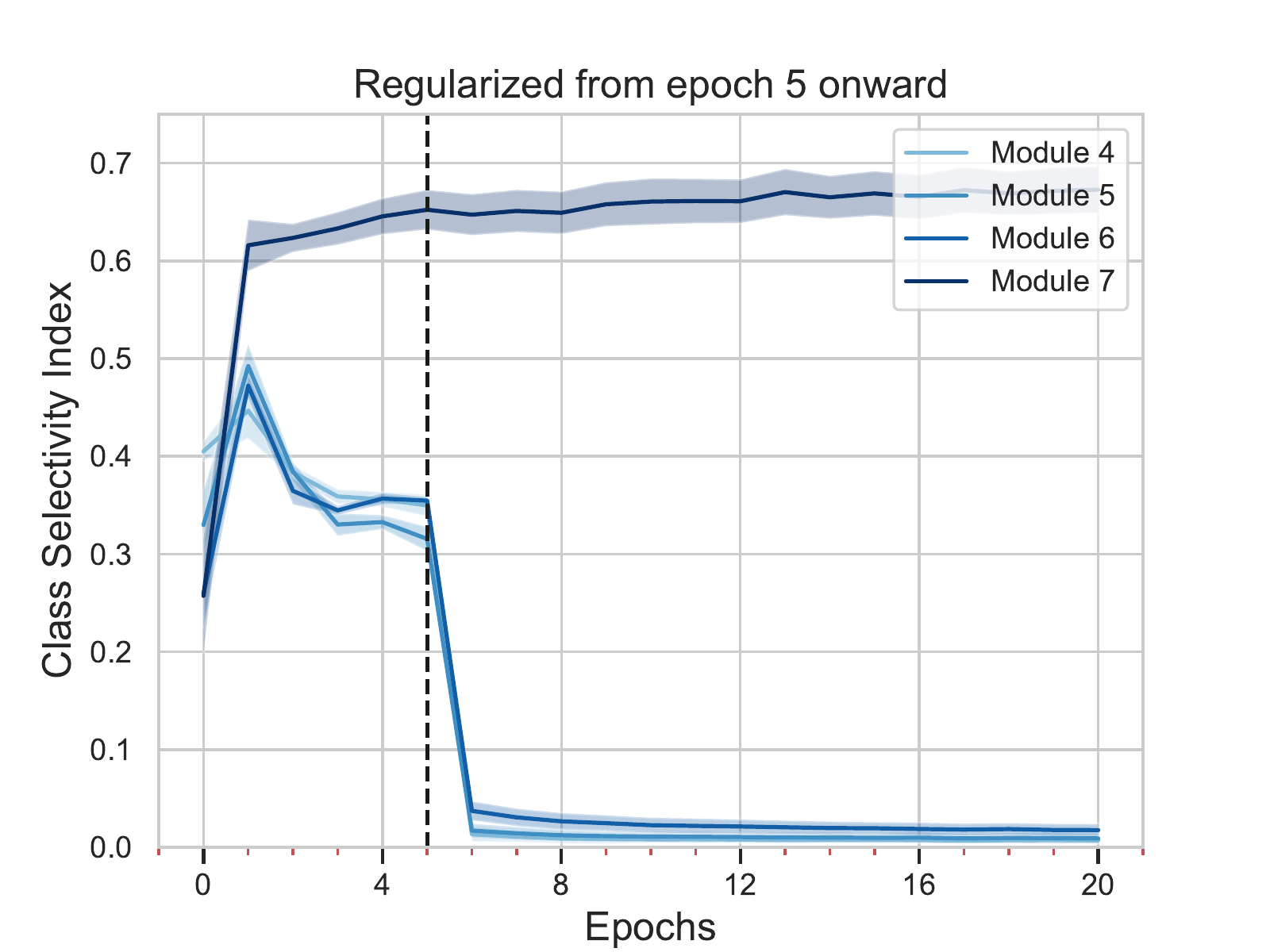}
         \caption{}
     \end{subfigure}
     
    \begin{subfigure}{\linewidth}
        \centering
         \includegraphics[width=0.45\linewidth]{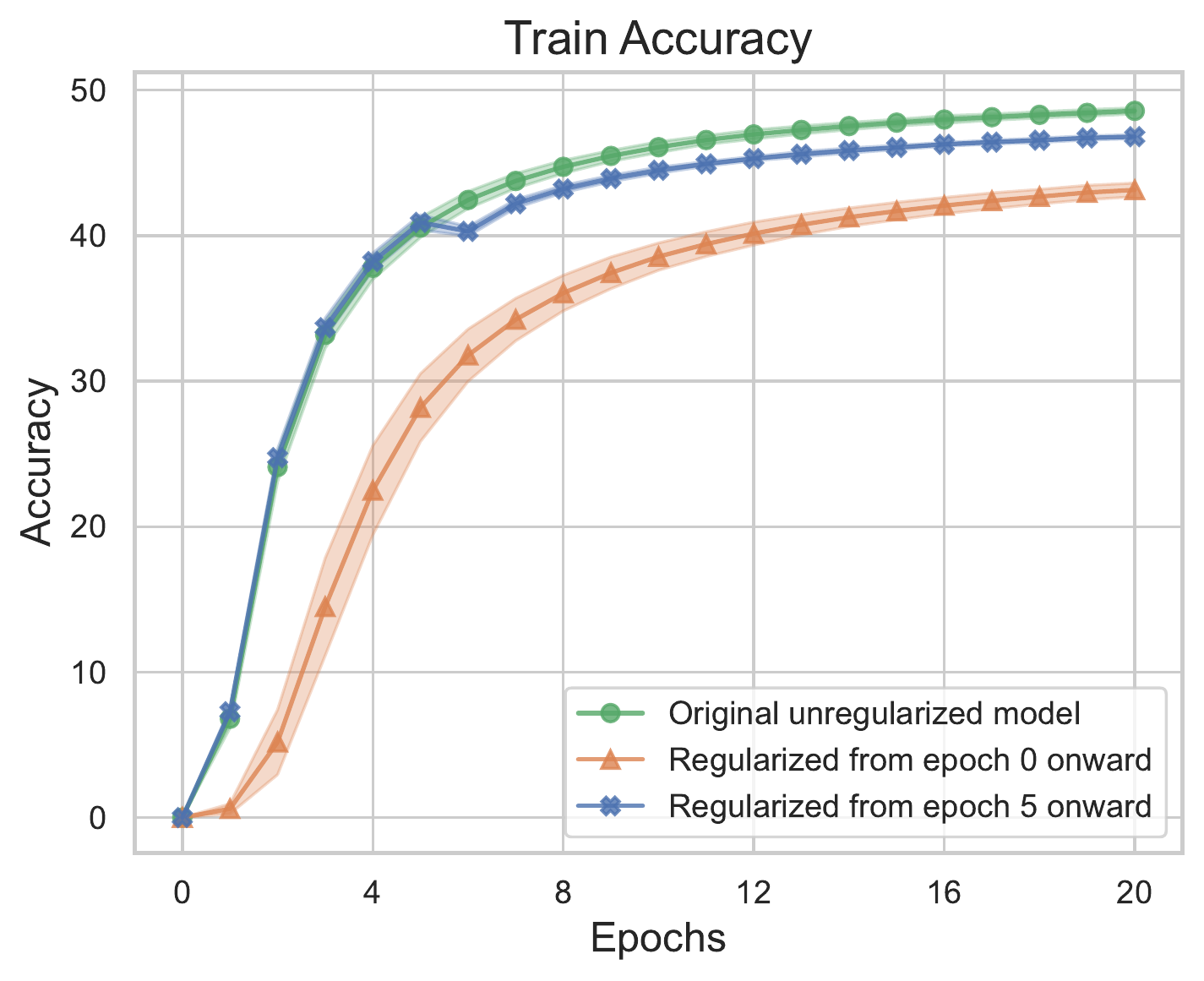}
         \includegraphics[width=0.45\linewidth]{Figures/reg_a20_comb_val.pdf}
          \caption{}
     \end{subfigure}

  \caption{\textbf{Regularization against selectivity.} \textbf{(a)} Class selectivity indices throughout training for three models, respectively unregularized, regularized from epoch 0 onward, and regularized from epoch 5 onward. The effect of regularization on class-selectivity is clearly visible in the second and third plots. Note that only early and intermediate modules are regularized. The latter module (module 7) is not regularized against selectivity. Error shades indicate the 95\% CI of the mean, calculated over all the bottleneck layers for each module, averaged over 5 trained instances in each case.
  \textbf{(b)} Train and Validation accuracies for each model throughout training averaged over 5 trained instances. The model regularized against selectivity from epoch 5 onward performs almost as well as the original unregularized model. The model regularized against selectivity from epoch 0 onward performs significantly worse than the other two models. This result shows the importance of class selectivity in the early epochs of training. }
  \label{reg_a20}

\end{figure}

We used a regularizing strength of $\alpha=-20$ and found that the regularizer prevents the peak of emergence of class-selective neurons in early epochs of training and that it impairs training and validation accuracy throughout training if regularized from epoch 0 (Fig \ref{reg_a20}). However, in the second scenario where the regularizer was turned on from epoch 5 onward, the model performed almost as well as the original unregularized model. These results indicate that the emergence of class-selective neurons in intermediate modules in early epochs of training is a phenomenon that plays a crucial role in the correct training of the network. Other sets of experiments with different values of $\alpha$ can be found in Appendix \ref{appReg1}. It is worth noting that \cite{leavitt2020selectivity} found that regularizing beyond the range of $\alpha=[-1,-5]$—depending on the model and dataset—had significant negative effects on network performance, while we were able to increase the magnitude of $\alpha$ all the way to $-20$ with minimal effect on network performance, so long as we did not regularize until epoch 5. Regularizing against class selectivity in early epochs also degrades performance of smaller versions of ResNets i.e., ResNet18 and ResNet34 (Appendix \ref{rn18_34_reg}).

The model never fully recovers the performance if the regularization is kept on, but if the regularization is turned off after the first few epochs of training then the network is slowly able to recover from the perturbation after 20+ epochs of training (Appendix \ref{appRecover}).

% \begin{figure}[t]
%   \centering
%   \includegraphics[width=0.9\linewidth]{Figures/reg_a20_comb.pdf}
%   \caption{\textbf{Regularization against selectivity.} \textbf{(a)} Class selectivity indices throughout training for three models, respectively unregularized, regularized from epoch 0 onward, and regularized from epoch 5 onward. The effect of regularization on class-selectivity is clearly visible in the second and third plots. Note that only early and intermediate modules are regularized. The latter module (module 7) is not regularized against selectivity. Error shades indicate the 95\% CI of the mean, calculated over all the bottleneck layers for each module.
%   \textbf{(b)} Train and Validation accuracies for each model throughout training. The model regularized against selectivity from epoch 5 onward performs as well as the original unregularized model. The model regularized against selectivity from epoch 0 onward performs significantly worse than the other two models. This result shows the importance of class selectivity in the early epochs of training. }
%   \label{reg_a20}
% \end{figure}

\subsection{Class Selectivity is most important during the first epoch of training}\label{mainE1}

\begin{figure}
    \centering
    \begin{subfigure}{0.45\linewidth}
    \includegraphics[width=\linewidth]{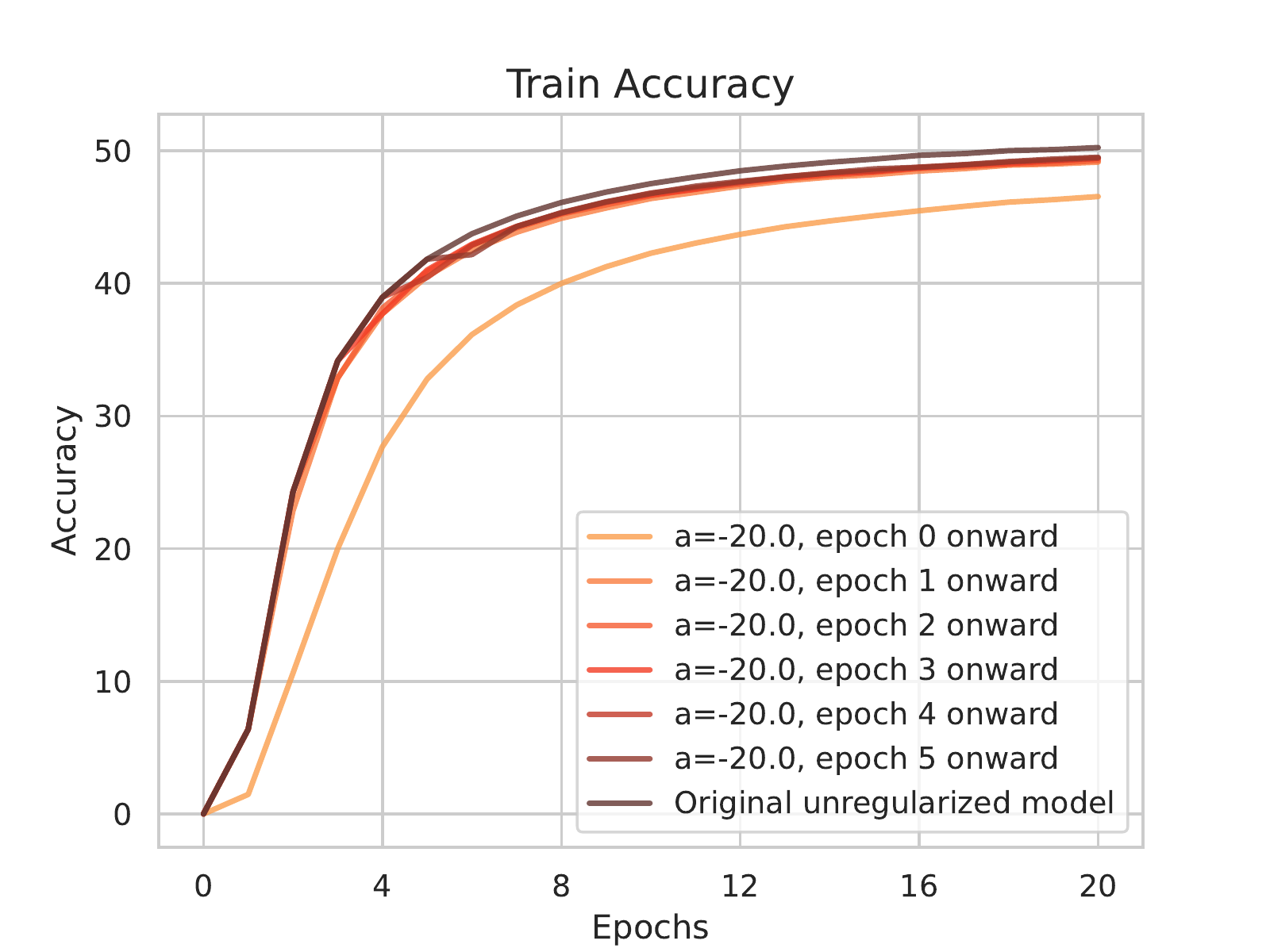}
    \end{subfigure}
    \begin{subfigure}{0.45\linewidth}
    \includegraphics[width=\linewidth]{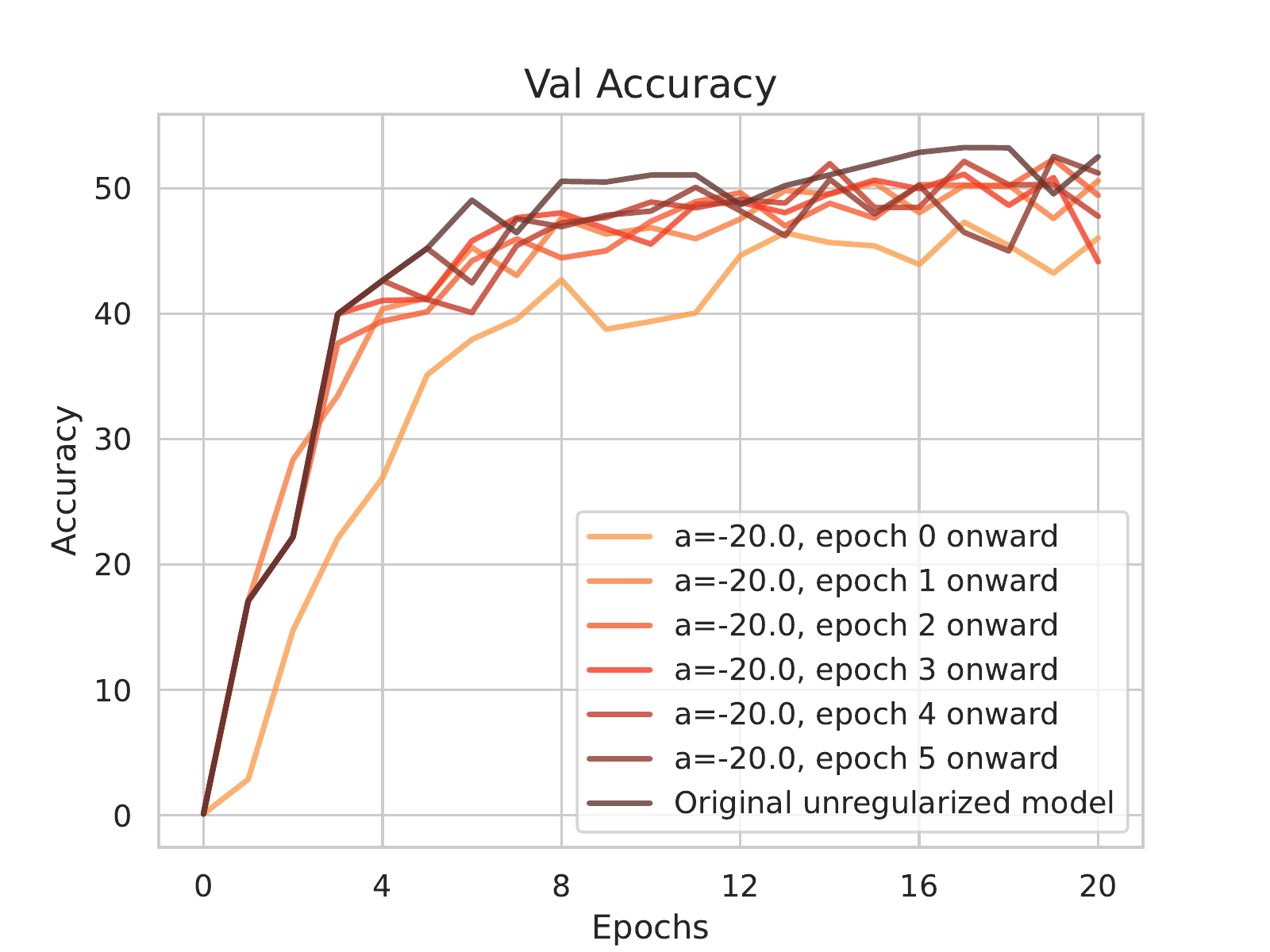}
    \end{subfigure}
\caption{\textbf{Effect of regularization against selectivity when starting from different epochs.} Only the model regularized from epoch 0 onward performs poorly. Models which were regularized starting from epoch 1-5, all performed similarly. This suggests that the critical phase during which class-selectivity is important is restricted to the very first epoch of training.}
\label{reg_epochs}
\end{figure}

In the previous set of experiments, we regularized against class-selectivity from epoch 5 onward, which starts after the spike in selectivity. But the next question which arises is what happens if we regularize in-between epochs 0 and 5? Fig \ref{reg_epochs} shows that learning is only impaired if the model is regularized from epoch 0 onward. Models regularized from epoch 1-5 onward performed similarly to the unregularized model, suggesting that the most critical phase for class selectivity happens during the very first epoch of training. 

\subsection{Regularizing in favor of class selectivity does not improve learning}
As class-selective neurons are crucial to learning in the early epochs, could increasing the selectivity in those early epochs further improve learning? To answer this question, we regularized \emph{in favor} of class-selectivity (i.e. with  positive values of $\alpha$) during those early epochs. We found that increasing the selectivity in those early epochs did not improve performance further (Fig \ref{regP1}, Fig \ref{regP2}). Therefore, both an increase and a decrease in selectivity is harmful to learning, suggesting that the network roughly learns the "right" amount of selectivity for optimal learning. More details on the regularization for selectivity experiments can be found in Appendix \ref{appReg2}.

\subsection{Effect of class selectivity regularization on the balance of class representation}

We next explored how the regularization in favor and against class selectivity affects the balance of class representation in the predictions of the network. We found that, both increasing and decreasing class selectivity in early and intermediate modules resulted in a poorer diversity of class representation in the predictions of the network (Appendix \ref{appBal}). This effect lasted for the few first epochs of training, after which it was no longer visible. We conclude that tampering with class selectivity in early and intermediate layers perturbs the correct learning of the network by over-representing certain classes at the prediction stage during a period which is apparently critical for the correct training of the network.

\section{Discussion}

Previous experiments have shown that class selectivity in individual units may not be necessary for high accuracy converged solutions \citep{s.2018on,amjad_understanding_2018,donnelly_interpretability_2019,leavitt2020selectivity,kanda_deleting_2020,leavitt_linking_2021}. We sought to determine \textit{why} class selectivity is learned if it is not necessary. In a series of experiments examining the training dynamics of ResNet50 trained on ImageNet, we found that class selectivity in early and intermediate layers is actually necessary during a critical period early in training in order for the network to learn high accuracy converged solutions. Specifically, we observed that class selectivity rapidly emerges within the first epoch of training, before receding over the next 2-3 epochs in early and intermediate layers. Ablating class-selective neurons during this critical period has a much greater impact on accuracy than ablating class-selective neurons at the end of training. Furthermore, we used \cite{leavitt2020selectivity}'s selectivity regularizer and found that strongly regularizing against class selectivity—to a degree that \cite{leavitt2020selectivity} found had catastrophic effects on network performance—had a negligible effect on network performance \textit{if regularization doesn't occur in the initial epochs}. We also found that the representations of early and latter layers are much more similar during the early critical period of training compared to later in training, indicating that selectivity in early layers may be leveraged by latter layers in order to solve the classification problem early in training. Taken together, our results show that class-selective neurons in early and intermediate layers of deep networks are a vestige of their emergence during a critical period early in training, during which they are necessary for successful training of the network.

One limitation of the present work is that it is limited to ResNet-50 trained on ImageNet. While \cite{leavitt2020selectivity} found that their results were consistent across a breadth of CNN architectures and image classification datasets, it is possible that our findings may not generalize to networks trained on NLP tasks, in which single neuron selectivity is also a topic of interest.

Our observations add to the list of phenomena occurring during the critical phase early in training (see related work section \ref{RWork}), and prompts more investigations regarding this phase. In particular, it would be interesting to understand \emph{why} the emergence of class-selectivity is necessary in this early phase of training, and whether this phenomenon connects to other critical-period phenomena, such as the emergence of lottery ticket subnetworks \citep{frankle_lottery_2018, paul2022lottery}, and critical learning flexibility \citep{achille_critical_2018}.

\bibliography{main}
\bibliographystyle{tmlr}

\appendix
\section{Appendix}

\begin{figure}[h]
     %\vspace{1cm}
     \centering
     \begin{subfigure}[b]{0.8\linewidth}
         \centering
         \includegraphics[width=\linewidth]{Figures/resnet_layers.pdf}
         \caption{High-level modular design of a ResNet-50.}
         \label{app_r50_overview}
     \end{subfigure}
     \hfill
     \begin{subfigure}[b]{0.3\linewidth}
         \centering
         \includegraphics[width=\linewidth]{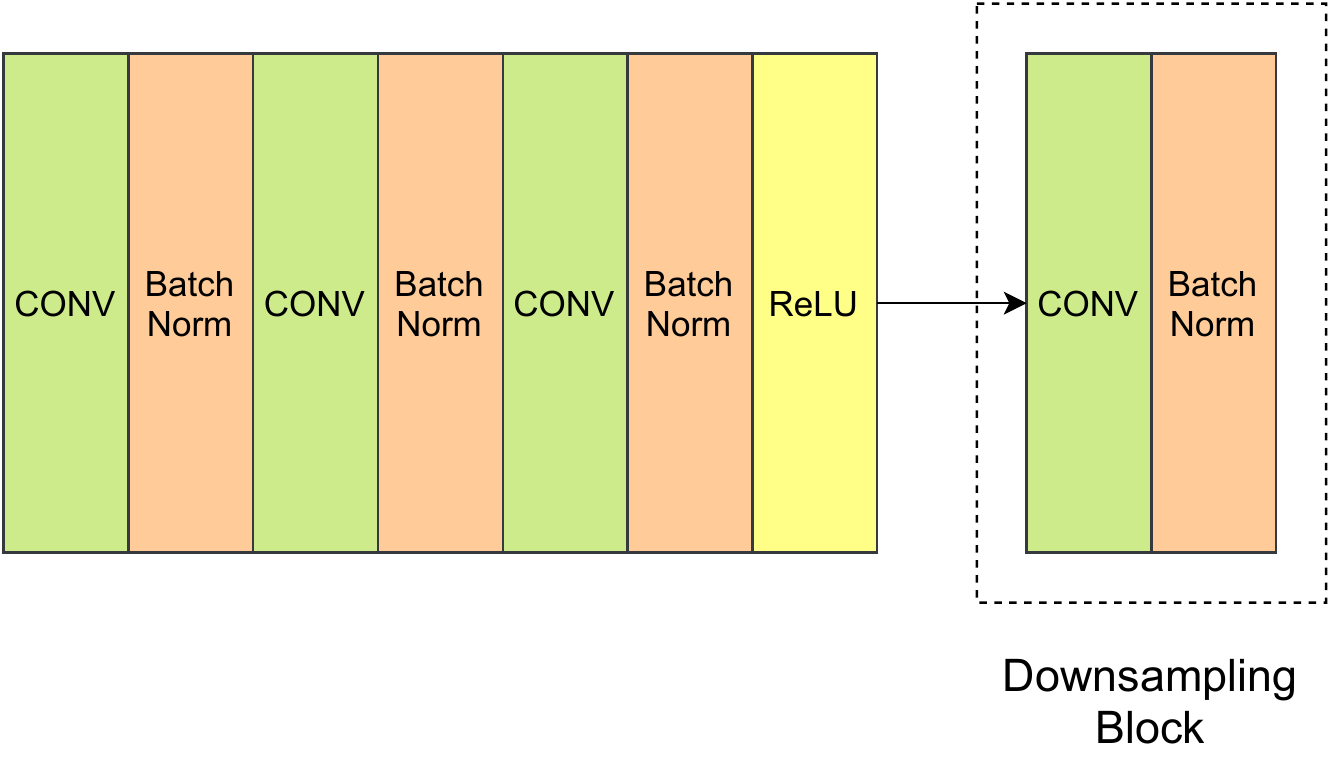}
         \caption{Structure of the first bottleneck layer inside every module (conv block).}
         \label{app_bn1}
     \end{subfigure}
     \hspace{0.1\linewidth}
     \begin{subfigure}[b]{0.3\linewidth}
         \centering
         \includegraphics[width=\linewidth]{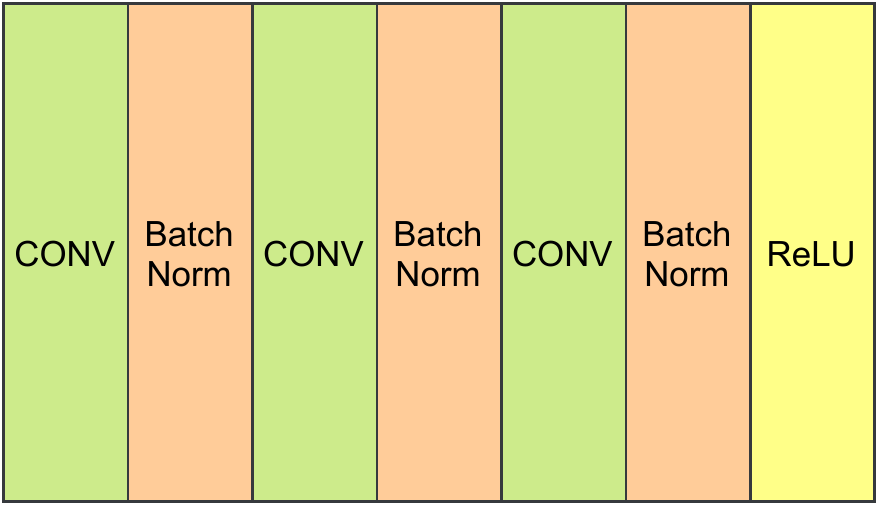}
         \caption{Structure of remaining bottleneck layers inside every module (identity block).}
         \label{app_bn2}
     \end{subfigure}
     \caption{\textbf{Structure of ResNet-50.} \textbf{(a)} We define the early and intermediate layers as the layers present in the early and intermediate modules. \textbf{(b, c)} Each of these modules are made of multiple bottleneck layers which are also commonly known as conv blocks and identity blocks in ResNets. }
     \label{app_r50_structure}
\end{figure}

\subsection{Structure of ResNet-50}\label{appStruc}
We define early, intermediate, and latter layers as the layers present in the early, intermediate, and latter modules respectively. (Fig \ref{app_r50_overview}). These modules are made up of bottleneck layers (Fig \ref{app_bn1}, \ref{app_bn2}). We calculate class selectivity index on individual channels extracted after the ReLU layer of each bottleneck layer.

\subsection{Model Training}\label{appTrain}
We trained 10 instances of ResNet-50 on ImageNet using standard training procedure using PyTorch \citep{paszke2019pytorch}. All instances were trained for 90 epochs with a batch size of 256, learning rate of 0.1, weight decay of 1e-4, and momentum of 0.9.

The class selectivity indices were calculated over the validation set of 50k images for every epoch from epoch 0 to epoch 90. All plots were generated using Seaborn Library \citep{waskom2021seaborn}. Accuracies shown in all plots are top-1 ImageNet training/validation accuracy. Error shades in all plots represent the 95\% confidence interval (CI) of the mean. 

\subsection{Class-Selective Ablations on a fully-trained network are less damaging than Random Ablations}\label{appAB} 
\begin{figure}[!hbt]
  \centering
  \includegraphics[width=0.65\linewidth]{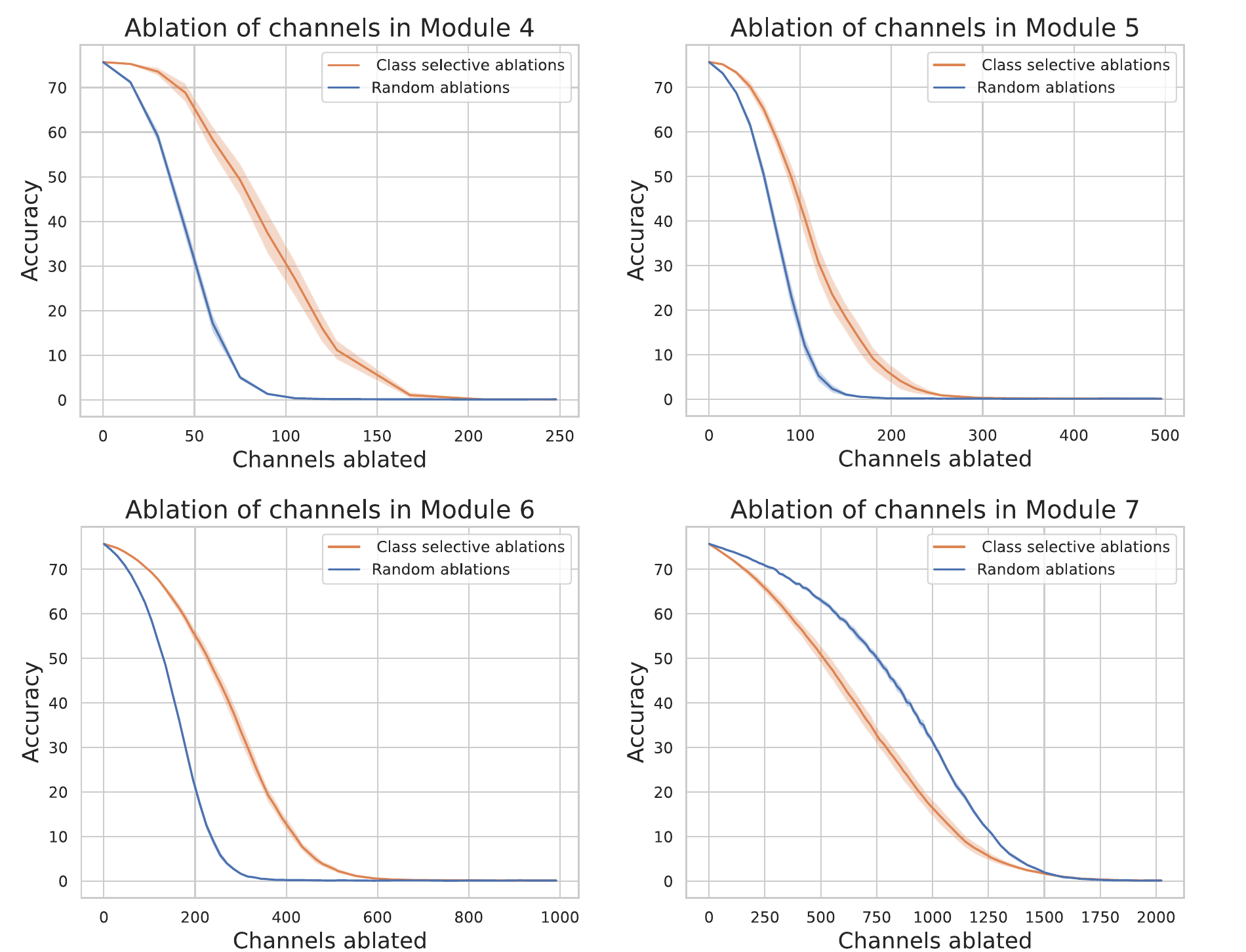}
  \caption{\textbf{Abation of channels in a fully-trained ResNet-50.} Surprisingly, ablating channels in a random order is more damaging to the network accuracy than ablating the most class-selective channels for modules 4, 5, 6. However, this trend is reversed for module 7. Error shades indicate the 95\% CI of the mean, calculated over 10 networks trained from different random initial conditions.}
  \label{avg_ab}
\end{figure}

We find that in the early and intermediate modules (modules 4, 5, 6), an ablation of units ordered by their class-selectivity rank (i.e. most class-selective units ablated first) affects the network accuracy less than a control experiment where we ablate channels in a random order (Fig \ref{avg_ab}), consistent with the findings of \cite{s.2018on}. This result is paradoxical, as it seems to indicate that although class-selective channels emerge during training in early modules, these channels do not support the decision of the fully-trained network. In the latter module (module 7), we find that class-selective channels do support the decision of the network more than random channels (Fig \ref{avg_ab}), consistent with our intuition (i.e. the layers near the end of the network do need to be class-selective for good performance) and previous findings by \cite{s.2018on}.

\begin{figure}
  \centering
  \includegraphics[width=0.75\linewidth]{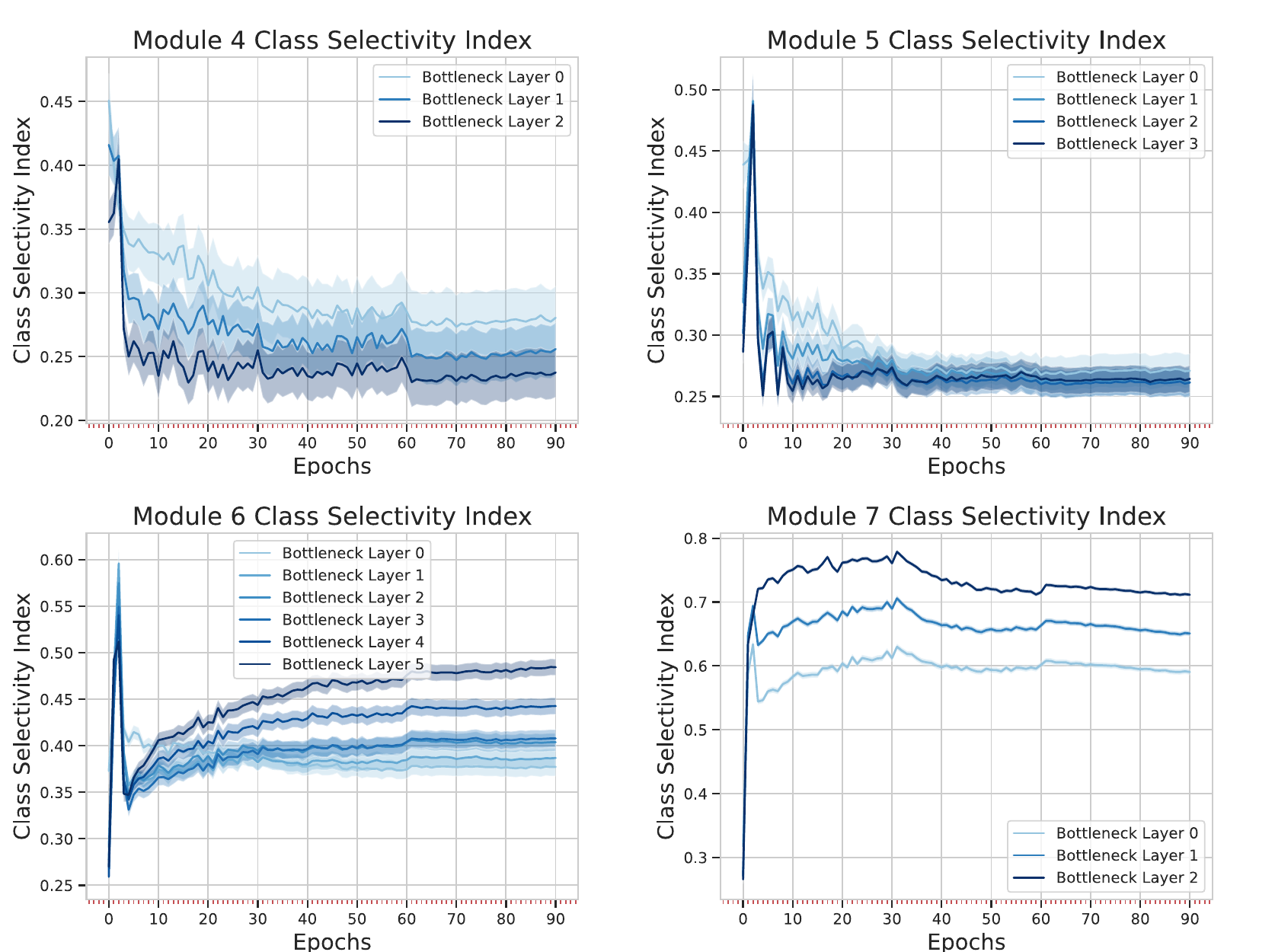}
\caption{\textbf{Evolution of class selectivity throughout training across bottleneck layers.}  Error shades indicate the 95\% CI of the mean, calculated over the channels of each bottleneck layer.}
\label{soc_bot}
\end{figure}

\subsection{Class Selectivity Index Across Bottleneck Layers}\label{appBott}

Fig \ref{soc_bot} shows the evolution of class selectivity throughout training for every bottleneck layer inside each module. An interesting observation here is that for module 6 and 7, the bottleneck layers are increasingly class-selective with network depth i.e, the latter bottleneck layers are more selective than the earlier ones inside those modules. For module 5, they exhibit roughly the same amount of selectivity later in training. However, for module 4, this trend is reversed and the bottleneck layers are decreasingly class-selective with network depth.

\begin{figure}
 \begin{subfigure}{\linewidth}
    \centering
    \includegraphics[width=\linewidth]{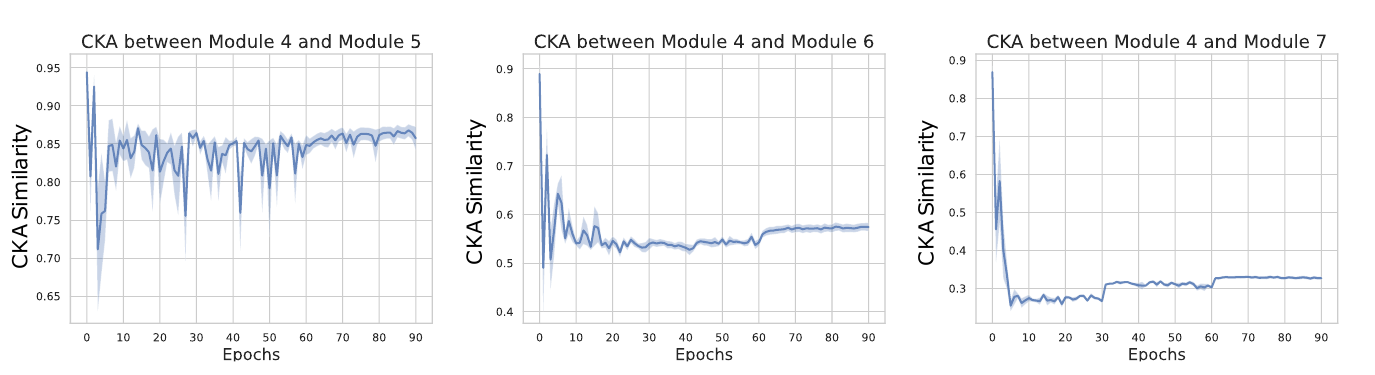}
    \caption{}
    \label{cka_m4}
 \end{subfigure}
 \hfill 
 \begin{subfigure}{\linewidth}
    \centering
    \includegraphics[width=\linewidth]{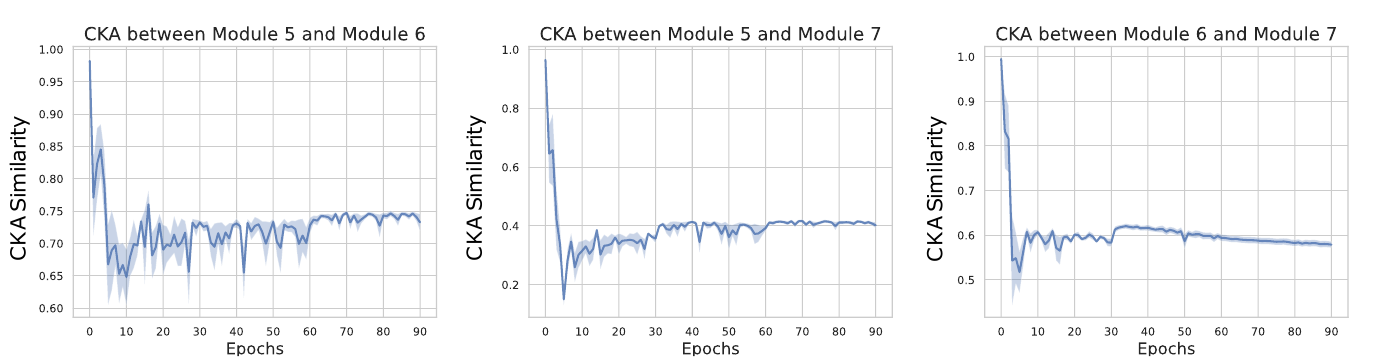}
    \caption{}
    \label{cka_m56}
 \end{subfigure}
\hfill
\begin{subfigure}{\linewidth}
    \centering
    \includegraphics[width=0.8\linewidth]{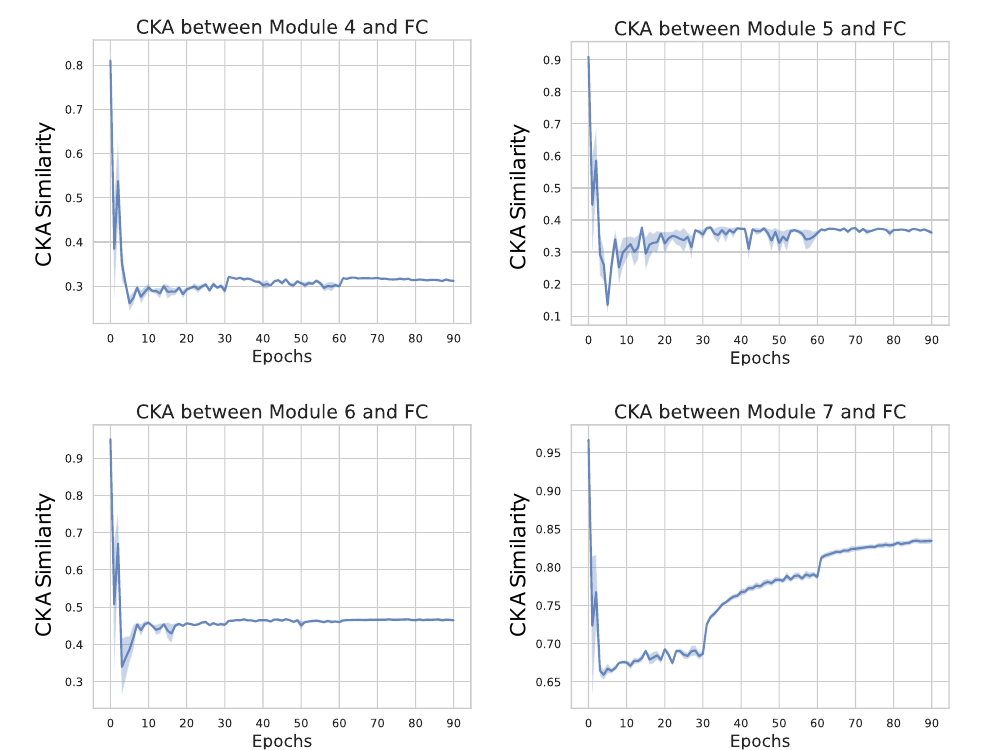}
    \caption{}
    \label{cka_fc}
\end{subfigure}
\caption{\textbf{CKA Analysis between different modules.} The representational similarity across all cases is high during the first few epochs after which it recedes to a low level for the rest of the training period. The high similarity in the initial epochs might explain the emergence of class-selective neurons as high similarity suggests the network is learning a crude, joint representation which causes the emergence of these class-selective neurons. Error shades indicate the 95\% CI of the mean, calculated over the 6 different networks trained from different random initial conditions.}
\label{cka_all}
\end{figure}

\subsection{Centered Kernel Alignment (CKA) Analysis}\label{appCKA}
To check the representational similarity between the modules at different stages of learning we used the Centered Kernel Alignment (CKA) \citep{kornblith2019similarity, nguyen2020wide} metric. To perform the CKA analysis we used the torch\textunderscore cka library \citep{subramanian2021torchcka}. The CKA similarity across all cases is high during the early epochs of training (Fig \ref{cka_all}) which suggests that the network is jointly learning to be class-selective, with neurons from latter layers relying on the class-selective features from early and intermediate layers to produce crude shortcut strategies. 

Fig \ref{cka_m4} shows that module 4 has a high representation similarity with module 5, 6, 7 in the first few epochs which then recedes to a low level for the rest of the training period. After the similarity level recedes, consistent with our intuition, the similarity decreases with increasing network depth i.e., after the first few epochs, module 4 is most similar to module 5, then to module 6, and then module 7. The remaining cases show a similar trend (Fig \ref{cka_m56}, Fig \ref{cka_fc}).

\subsection{Regularization Experiments}\label{appReg}

\begin{figure}
  \centering
  \includegraphics[width=0.8\linewidth]{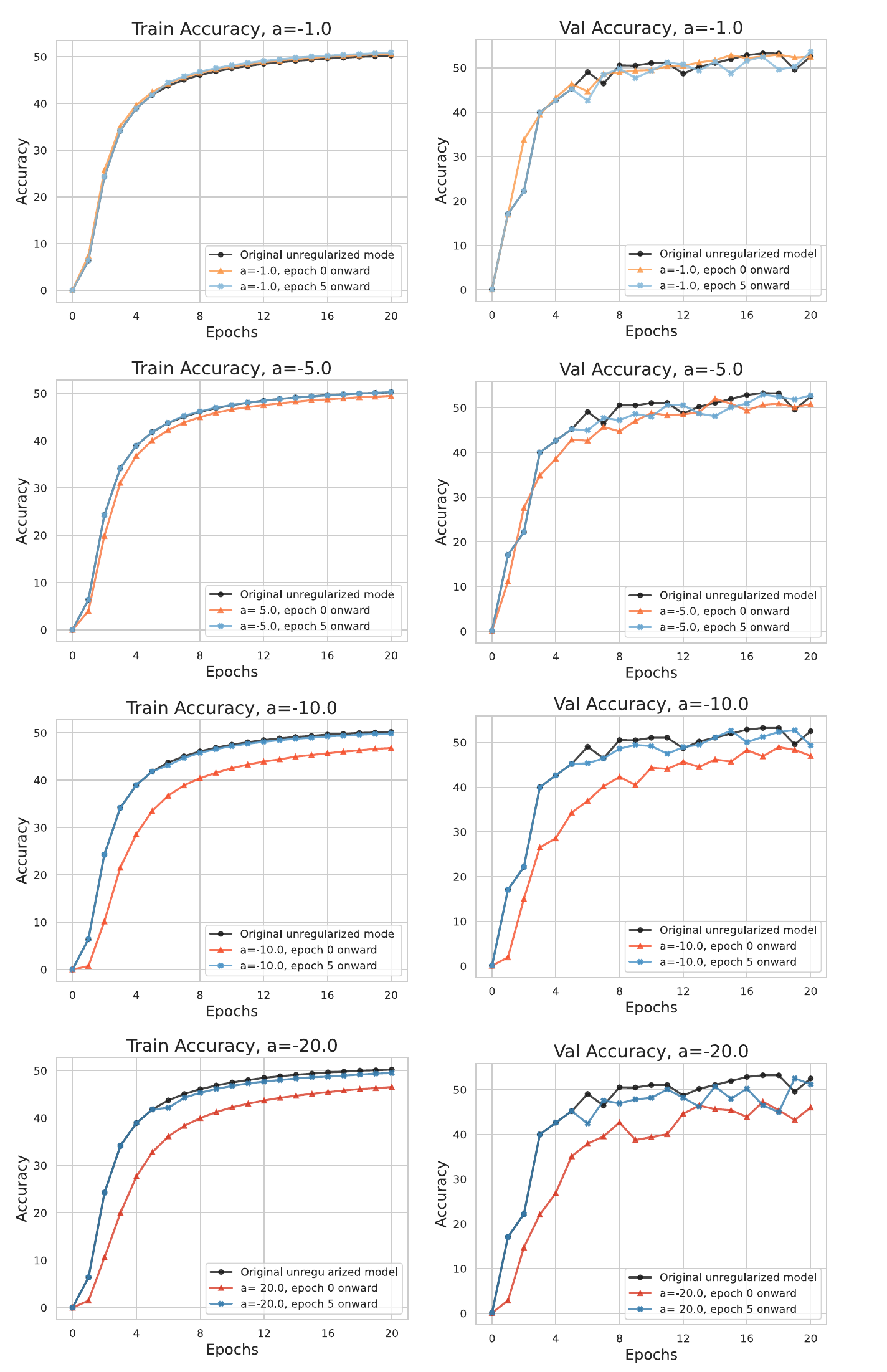}
\caption{\textbf{Regularizing against selectivity for different values of $\bm{\alpha}$ throughout training.}}
\label{reg_neg}
\end{figure}

\begin{figure}
  \centering
  \includegraphics[width=0.8\linewidth]{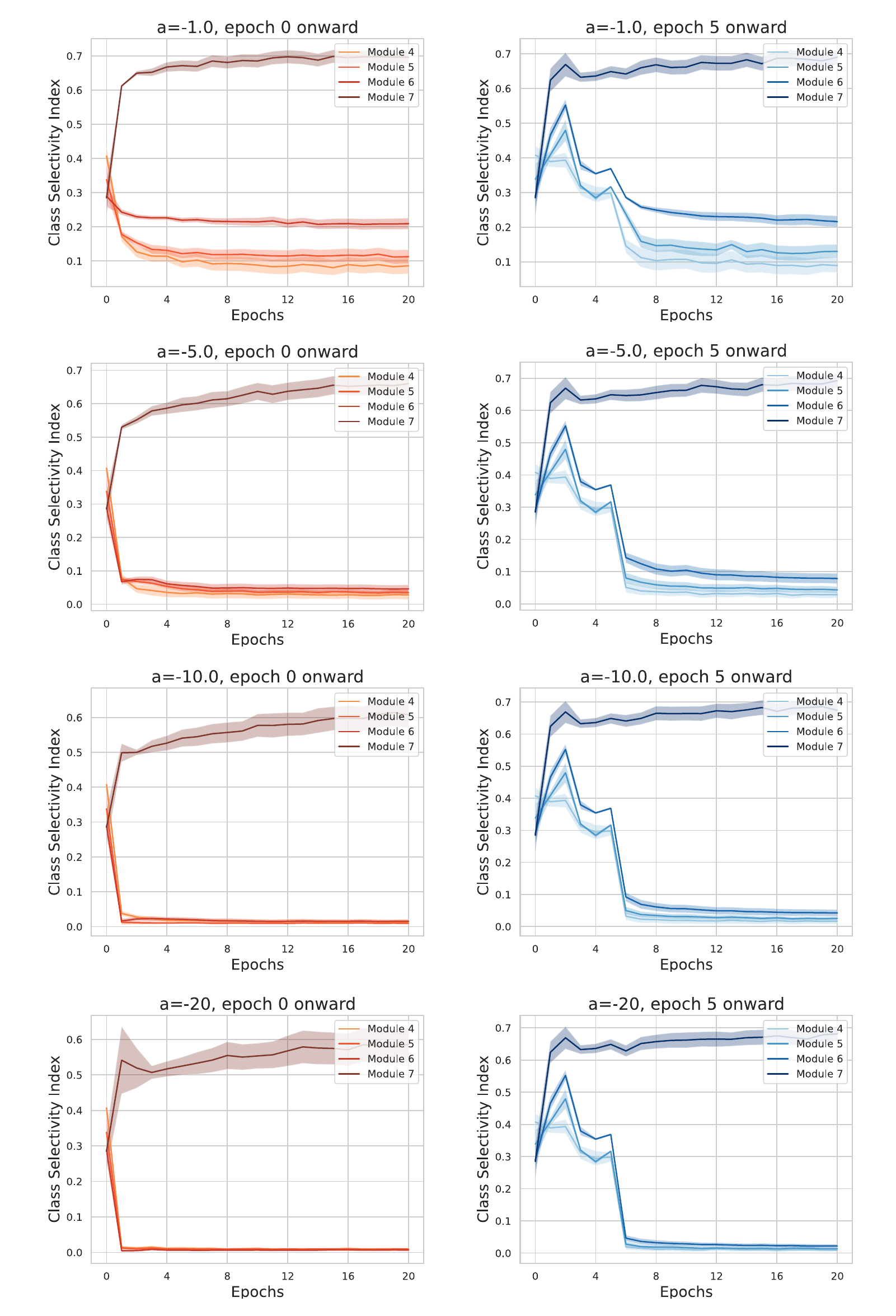}
  \caption{\textbf{Effect of regularization on selectivity indices across modules for different values of $\bm{\alpha}$ throughout training.}  Error shades indicate the 95\% CI of the mean, calculated over the bottleneck layers of each module.}
  \label{soc_neg}
\end{figure}

\subsubsection{Regularizing against selectivity}\label{appReg1}
We conduct a set of regularization experiments where we regularize against selectivity for different values of $\alpha$. We want to test two scenarios: Regularizing from epoch 0 onward VS regularizing from epoch 5 onward (epoch 5 is approximately after the spike in selectivity). We only regularize the early and intermediate modules (i.e. module 4, 5, 6) as we are interested in understanding the emergence of class selectivity in those modules. Also, regularizing only the early and intermediate modules allows us to use a higher values of $\alpha$ without squashing the network accuracy completely. The results for different values of $\alpha$ are shown in Fig \ref{reg_neg}. The following main observations can be made: 
\begin{itemize}
    \item For $\alpha$ = -1.0, there isn't much difference between the unregularized model and the models regularized from epoch 0 and epoch 5 onward. 
    \item For $\alpha$ = -5.0, the model regularized from epoch 0 onward performed worse than the one regularized from epoch 5 onward. 
    \item For $\alpha$ = -10.0, -20.0, the models regularized from epoch 0 onward performed \textit{significantly} worse than the ones regularized from epoch 5 onward. 

\end{itemize}
These results indicate that selectivity is important for learning in those early epochs and suppressing it by a significant amount impairs learning. 

\subsubsection{Regularizing for selectivity}\label{appReg2}
\begin{figure}[!hbt]
  \centering
  \includegraphics[width=0.8\linewidth]{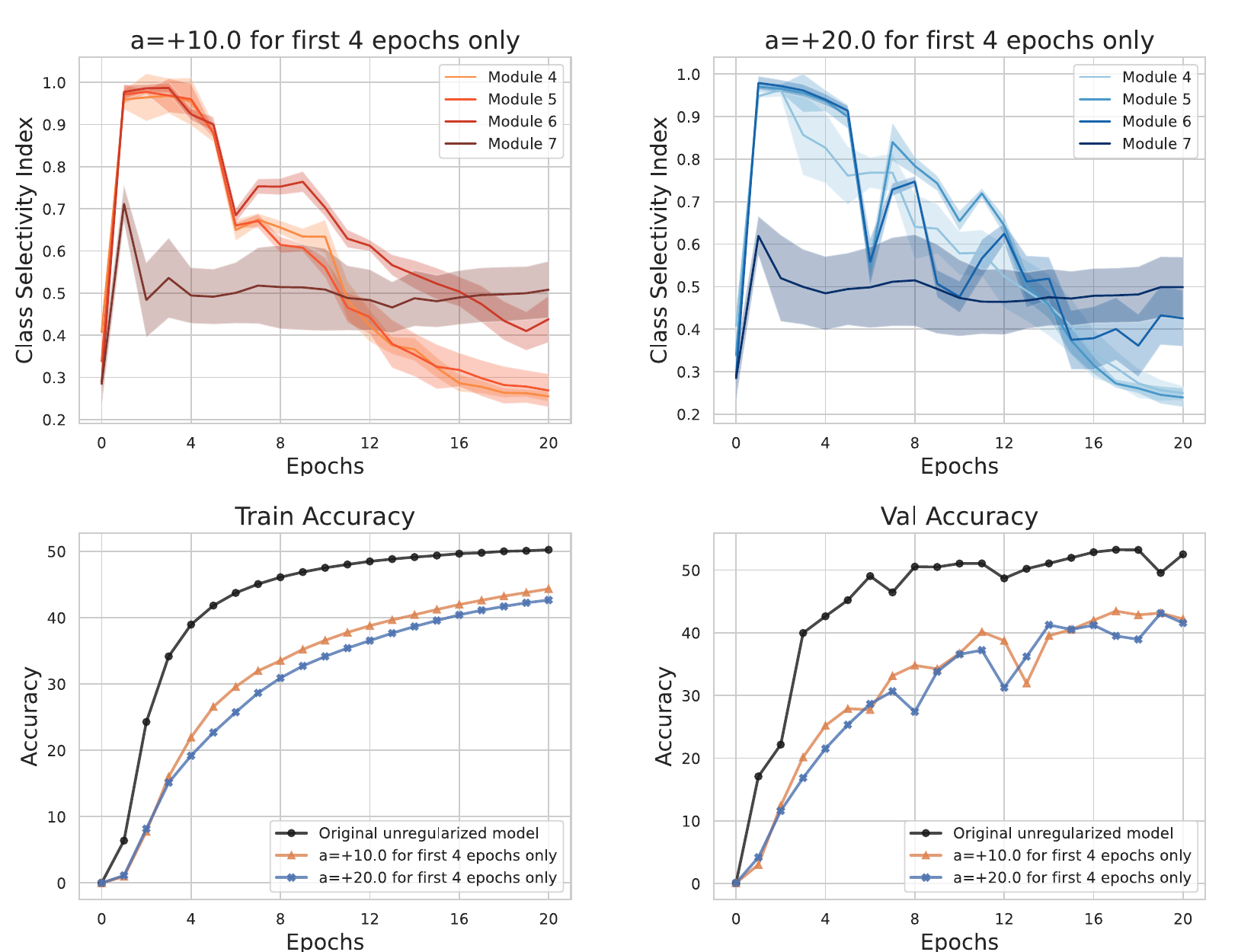}
  \caption{\textbf{Regularizing for selectivity with large values of $\bm{\alpha}$}. Models regularized for selectivity with large $\alpha$ in the early epochs are harmful to performance. The regularization was performed on the early and intermediate modules (module 4, 5, 6) for first 4 epochs of training.  Error shades indicate the 95\% CI of the mean, calculated over the bottleneck layers of each module.}
  \label{regP1}
\end{figure}

If selectivity is important for learning in the early epochs, then does increasing selectivity further in early epochs improve learning? To test this, we conducted experiments in which we regularized for selectivity with different values of $\alpha$. First, we tried regularizing the early and intermediate modules with large values of $\alpha$ (+10, +20) for the first four epochs, after which the regularizer was turned off. The latter module (module 7) was not regularized for selectivity. We found that in both cases, the model performed much worse as compared to the original unregularized model (Fig \ref{regP1}). The large $\alpha$ causes the selectivity index to go close to 1.0. Also, as module 7 was not regularized, its selectivity index ends up being lower than the other modules (Fig \ref{regP1}). 

Next, we re-did the experiment with smaller values of $\alpha$ (+1, +5) and we also decided to regularize module 7 for selectivity along with the early and intermediate modules. This time, we tried the experiment on two scenarios:
\begin{itemize}
    \item Turning the regularizer on for first four epochs for all modules with $\alpha$ = +1, +5. 
    \item  Turning the regularizer on only for epoch 1 (as it is the most important (Section \ref{mainE1})) for all modules with $\alpha$ = +1, +5. 
\end{itemize}

\begin{figure}[!hbt]
  \centering
  \includegraphics[width=\linewidth]{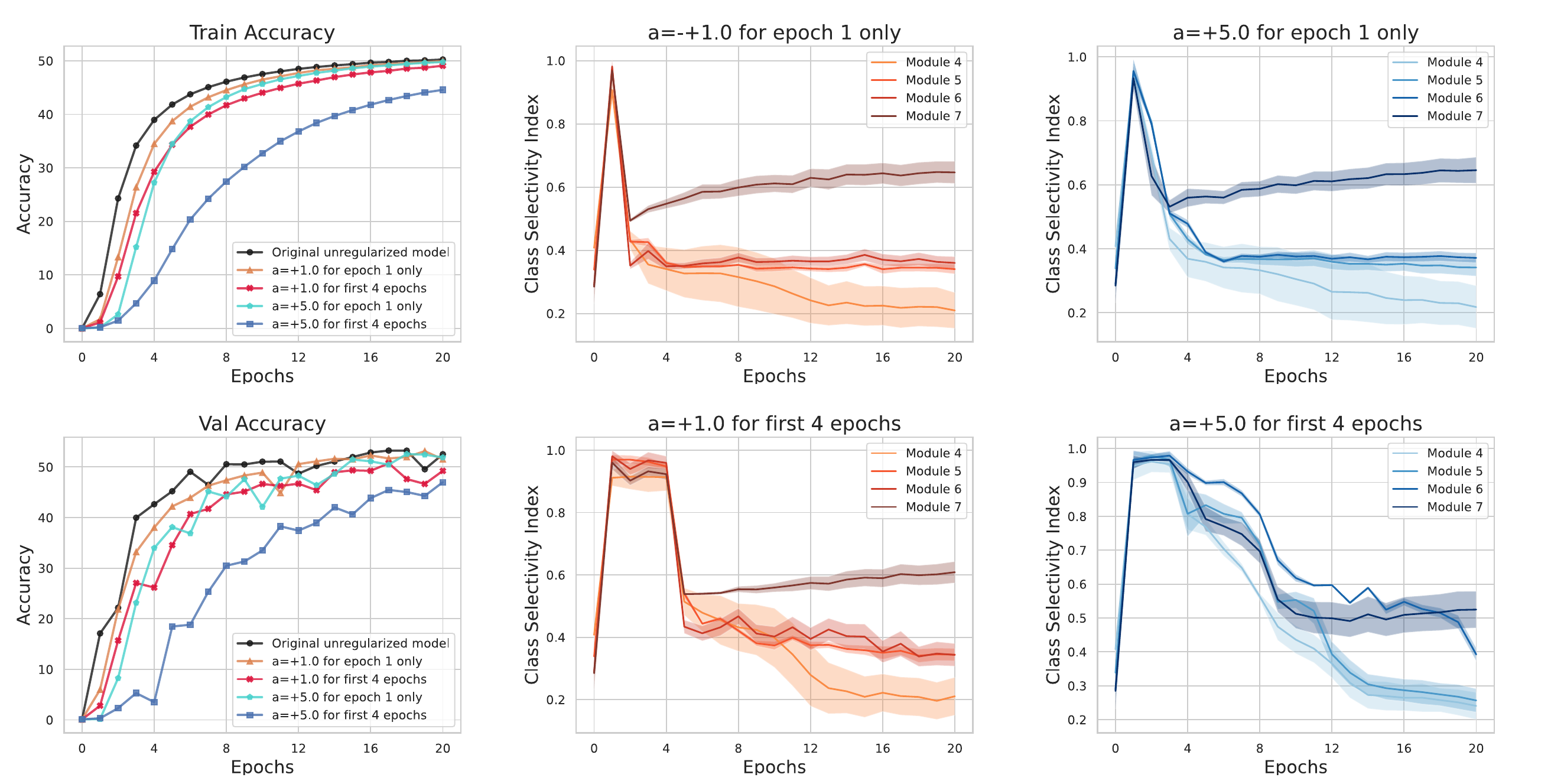}
  \caption{\textbf{Regularizing for selectivity with smaller values of $\bm{\alpha}$. } Even models regularized with smaller values of $\alpha$ do not help in increasing the performance. The regularization was performed on all modules for two cases: Over first four epochs and only over the first epoch.  Error shades indicate the 95\% CI of the mean, calculated over the bottleneck layers of each module.}
  \label{regP2}
\end{figure}

In all cases, the increase in selectivity was harmful to performance (Fig \ref{regP2}). So overall, the results show that increasing selectivity does not improve performance. Therefore, both an increase and a decrease in selectivity is harmful to learning. 

\subsubsection{Can the model eventually recover its performance from the effects of regularization?}\label{appRecover}

\begin{figure}
    \centering
    \begin{subfigure}{0.45\linewidth}
    \includegraphics[width=\linewidth]{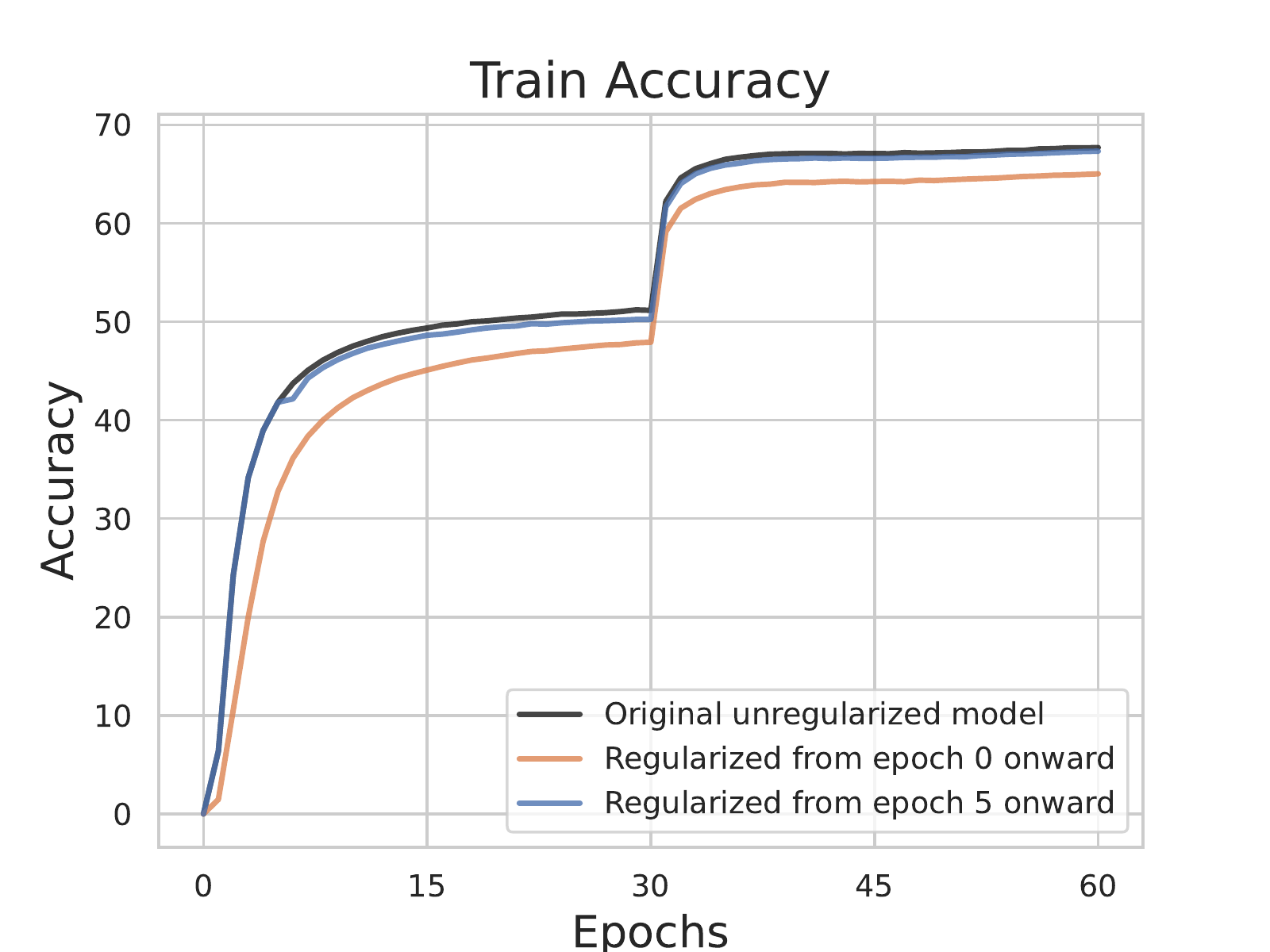}
    \caption{}
    \label{regRecover1}
    \end{subfigure}
    \begin{subfigure}{0.45\linewidth}
    \includegraphics[width=\linewidth]{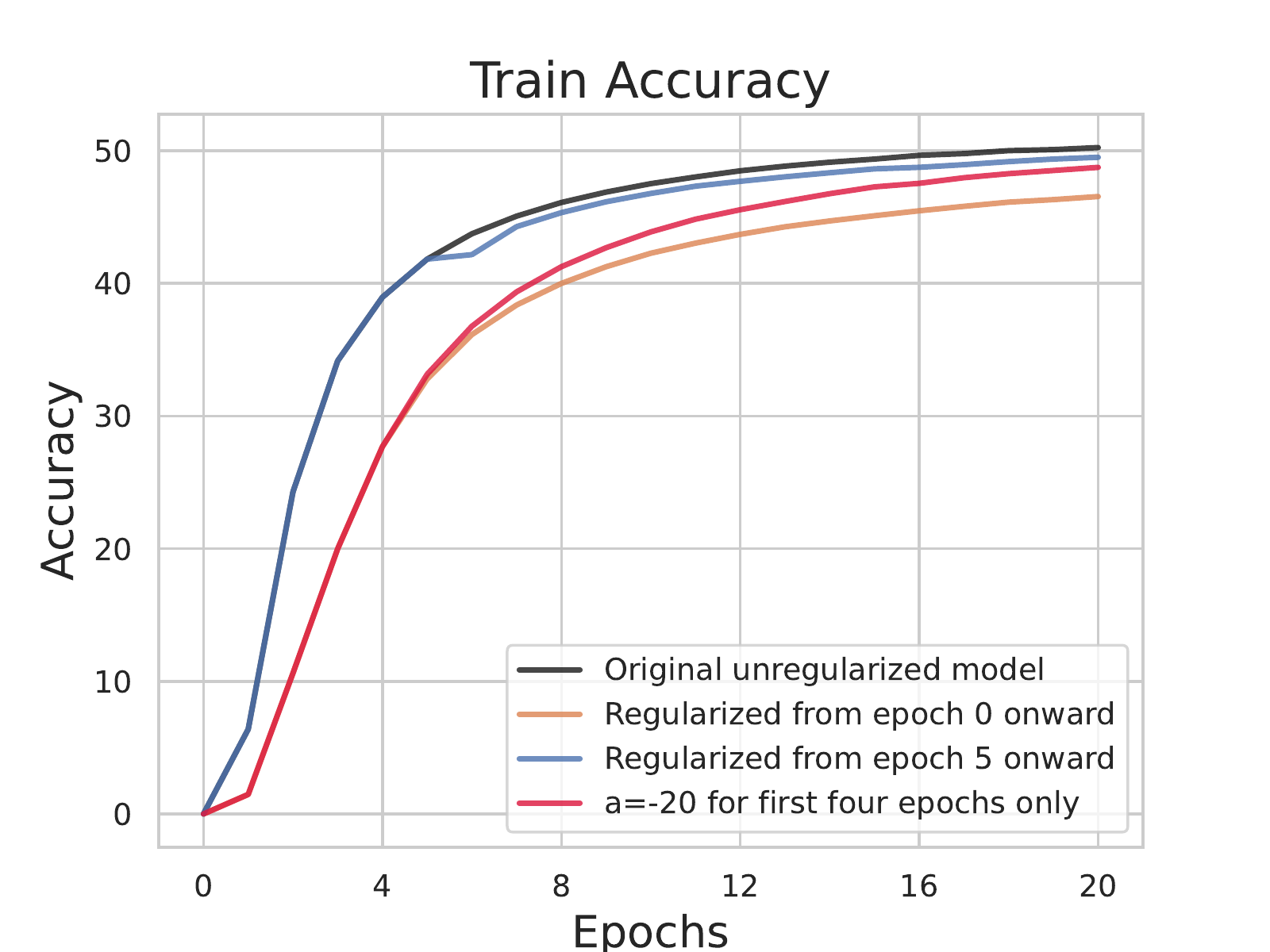}
    \caption{}
    \label{regRecover2}
    \end{subfigure}
\caption{\textbf{Can the model recover from the effects of regularization?} \textbf{(a)} The models are trained longer with the regularizer kept on at $\alpha$ = -20, but the model trained from epoch 0 is never able to recover. \textbf{(b)} If the regularizer is only kept on for only first four epochs then the model manages to recover from the effects of regularization.}
\label{regRecover}
\end{figure}

The regularization experiments conducted in Section \ref{mainReg} and Appendix \ref{appReg1} are done till epoch 20. So, this naturally leads to the following questions: 
\paragraph{Does the model eventually recover from the effects of regularization if it is trained for more epochs?} 
We trained the models regularized from epoch 0 onward and the one regularized from epoch 5 onward for 60 epochs while keeping the regularizer on with $\alpha$ = -20. The model regularized from epoch 0 onward is never able to recover the lost performance (Fig \ref{regRecover1}). 

\paragraph{Does the model eventually recover from the effects of regularization if the regularization is turned off after the early epochs?}

Next, we regularized a third model for only the first four epochs with $\alpha$ = -20 after which the regularizer was turned off. This model was able to eventually recover the lost performance by epoch 20 (Fig \ref{regRecover2}). 

These two experiments confirm that class selectivity plays an important role towards model's performance. 

\subsubsection{Effect of regularization on balance of class representation}\label{appBal}

\begin{figure}[h]
  \centering
  \includegraphics[width=0.6\linewidth]{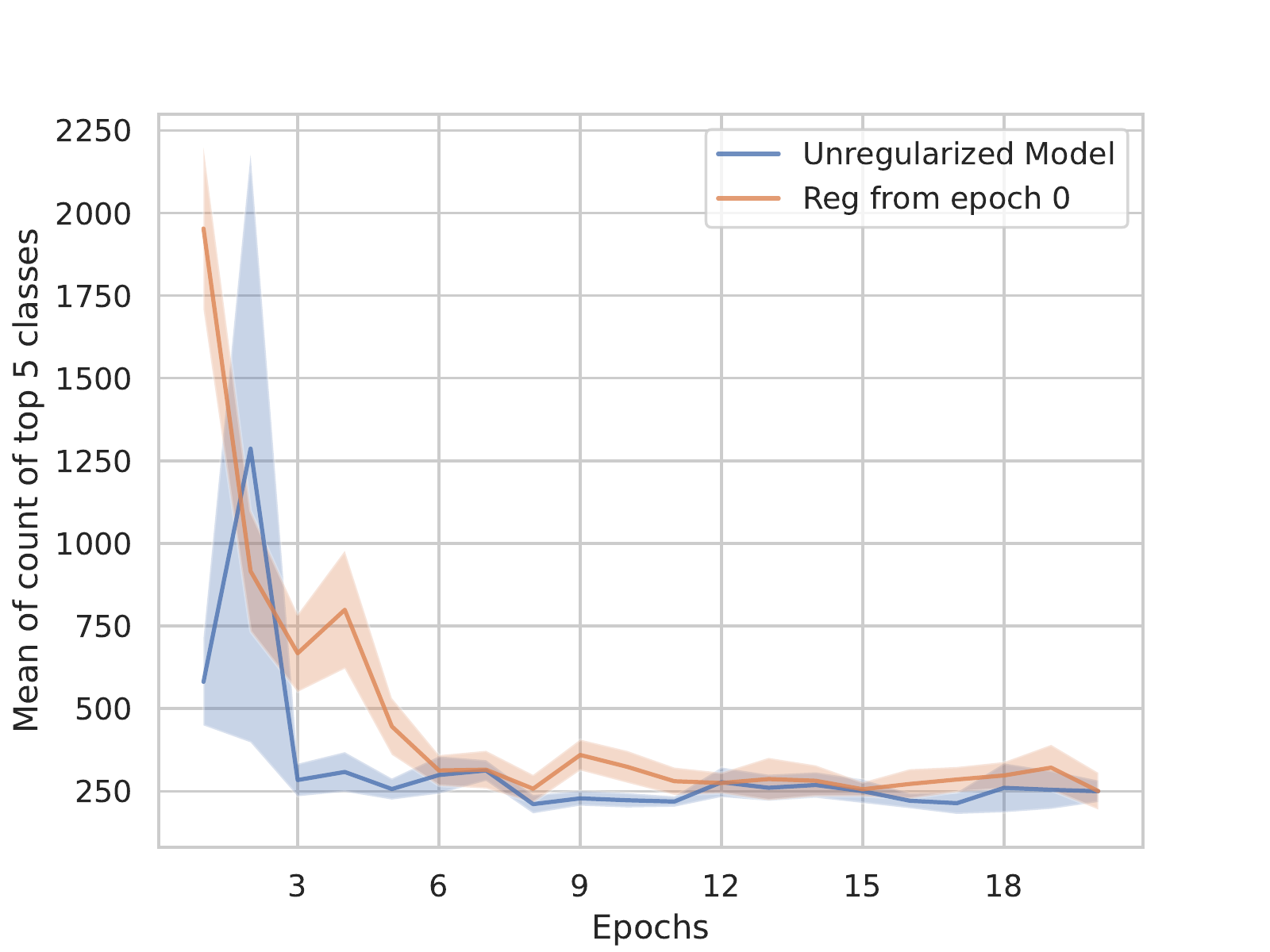}
  \caption{\textbf{Class representation over epochs. } The figure shows the mean of top-5 class counts, where the class counts are counted as the classes predicted by the model over the validation set in each epoch. The model regularized from epoch 0 over-represents classes in the early epochs.  Error shades indicate the 95\% CI of the mean, calculated over the top-5 mean count of two different networks trained from different random initial conditions.}
  \label{regBal}
\end{figure}

What happens to the class representation when we regularize against selectivity? To test this, we kept a count of each class predicted by the model on the validation set for each epoch, and then took the mean of top-5 class counts for each epoch. One would expect that if the model is regularized against selectivity then classes won't be over-represented i.e., the mean of top-5 class counts should be low for each epoch. However, surprisingly, the mean of top-5 class counts is actually higher in case of the model regularized from epoch 0 when compared to the unregularized model (Fig \ref{regBal}). So even if the individual neurons aren't class-selective, the model ends up over-representing certain classes in the early epochs. We think this might be because if neurons are not class-selective at all, then the model might have difficulty learning in the early epochs and so the model as a whole ends up overfitting to a few classes. 

Similarly, when the model is regularized in favor of selectivity, the classes again end up being over-represented. This suggests that tampering with the selectivity in either direction perturbs the correct learning by over-representing classes. 

\subsection{Analyzing class selectivity at a sub-epoch resolution}

\begin{figure}
    \centering
    \begin{subfigure}{0.45\linewidth}
    \includegraphics[width=\linewidth]{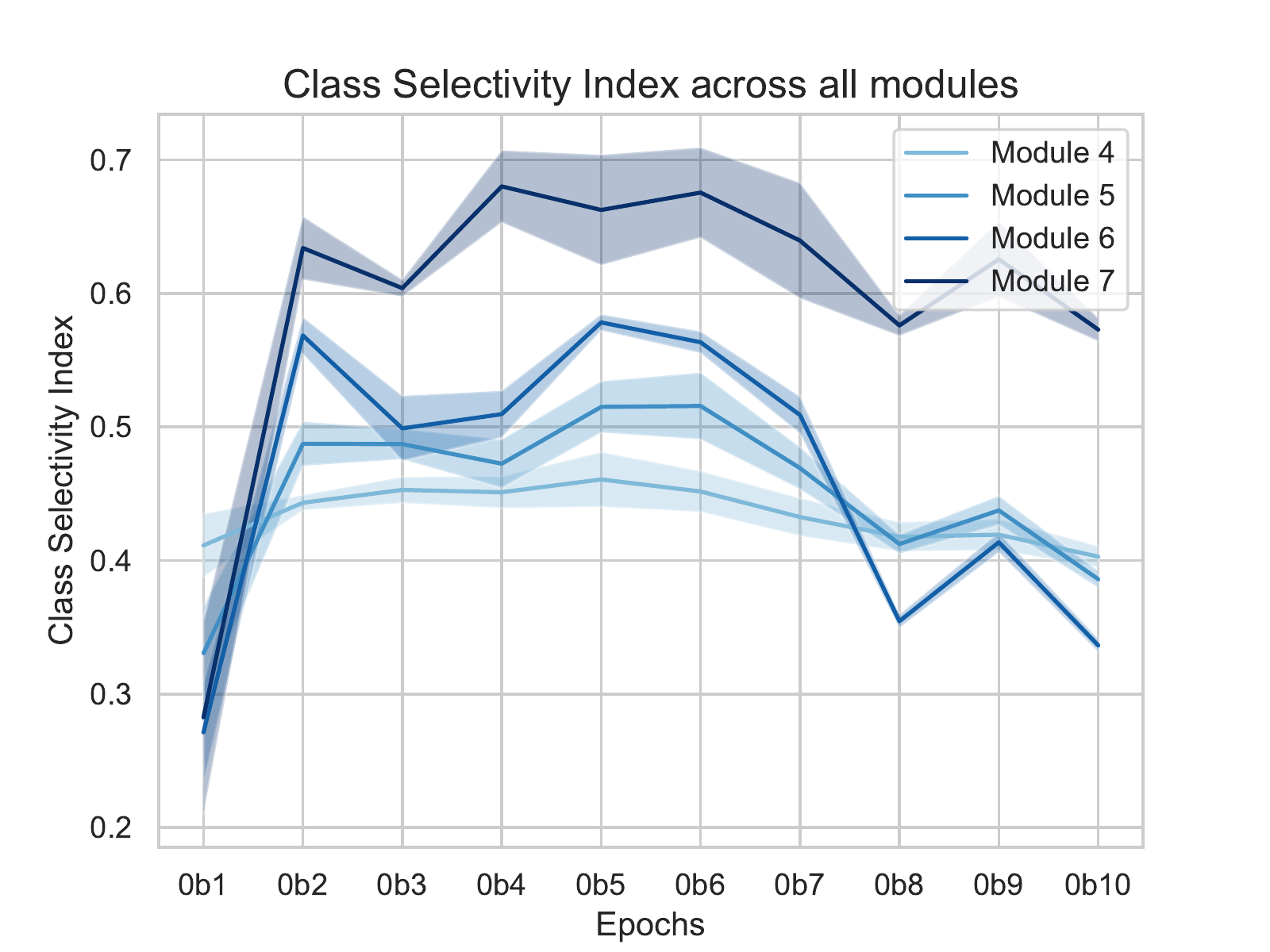}
    \caption{}
    \label{se00}
    \end{subfigure}
    \begin{subfigure}{0.45\linewidth}
    \includegraphics[width=\linewidth]{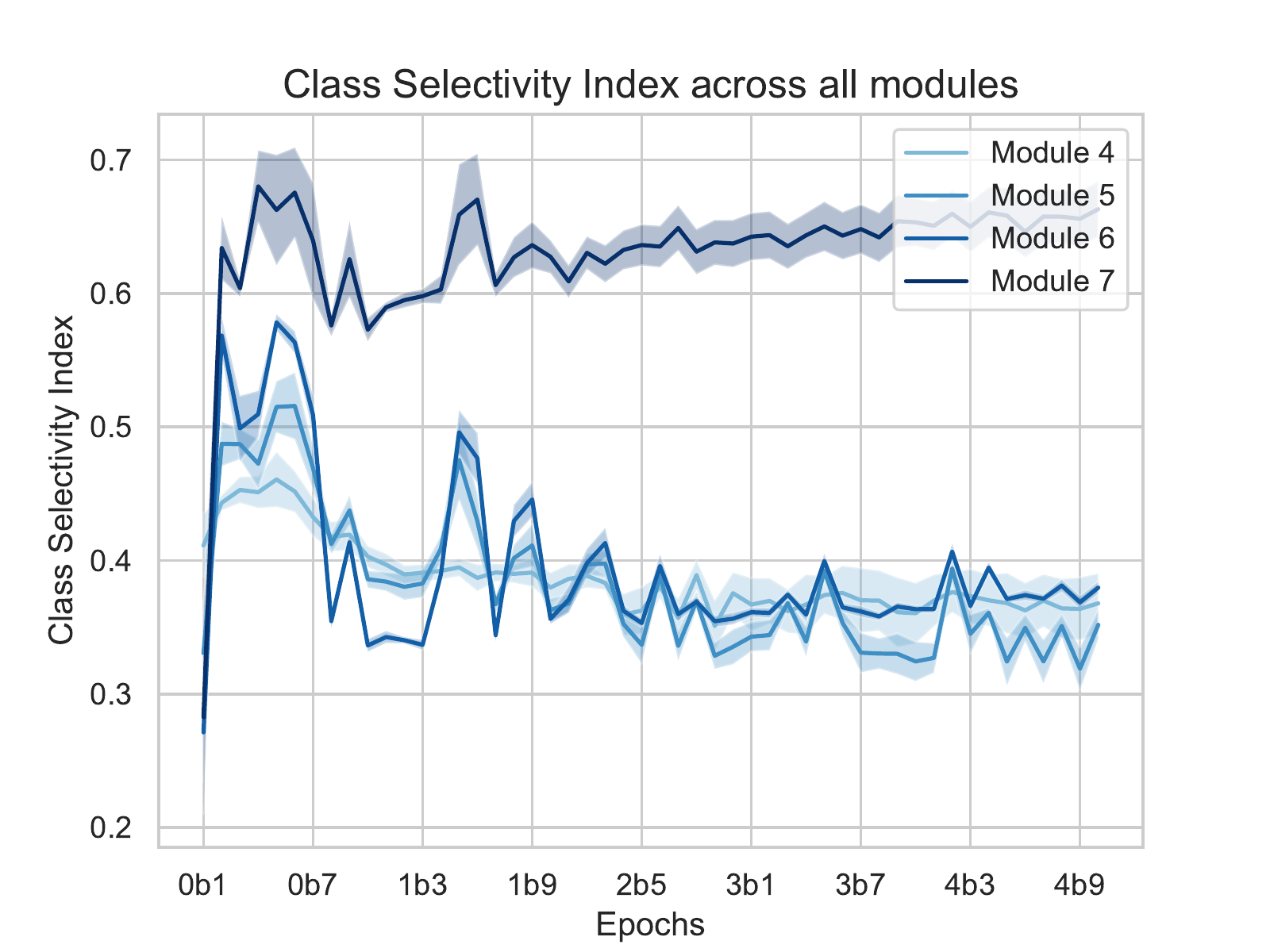}
    \caption{}
    \label{se04}
    \end{subfigure}
\caption{\textbf{Evolution of class selectivity at a sub-epoch resolution.} The model is saved every 1000 batches of training within an epoch and the labels on the x-axis are of the form (e)b(i) where e represents the epoch number and i represents the batch number (in multiples of 1000). 
\textbf{(a)} The selectivity index rises for all modules after training the model on first 2000 batches. \textbf{(b)} Selectivity is most prominent during the first few thousand batches after which it starts to settle down. }
\label{subepoch}
\end{figure}

To understand how soon class selectivity rises during the first epoch of training, we analyzed selectivity at a sub-epoch resolution (Fig \ref{subepoch}). The model was saved 10 times within each epoch, i.e, the model was saved after every 1000 batches of training with a batch size of 128, for the first few epochs. 
Fig \ref{se00} shows the evolution of selectivity during the first epoch of training. We can see that the selectivity rises after training the model on the first 2000 batches (Point 0b2 in Fig \ref{se00}). Fig \ref{se04} shows the evolution of selectivity at a sub-epoch resolution from epoch 0 to epoch 4. We can observe that the selectivity is most prominent during the first few thousand batches of training after which it starts to settle down.

\subsection{Class Selectivity in smaller variants of ResNets}
In this section, we investigate whether smaller variants of ResNet (i.e. ResNet18/34) also exhibit similar class selectivity characteristics as a ResNet50. The ResNet18 and ResNet34 variants can also be divided into four modules like the ResNet50 (Fig \ref{app_r50_structure}). Both ResNet18 and ResNet34 have fewer layers and also fewer channels in each layer compared to a ResNet50.  

\subsubsection{Evolution of Class Selectivity in all modules}

\begin{figure}
    \centering
    \begin{subfigure}{0.45\linewidth}
    \includegraphics[width=\linewidth]{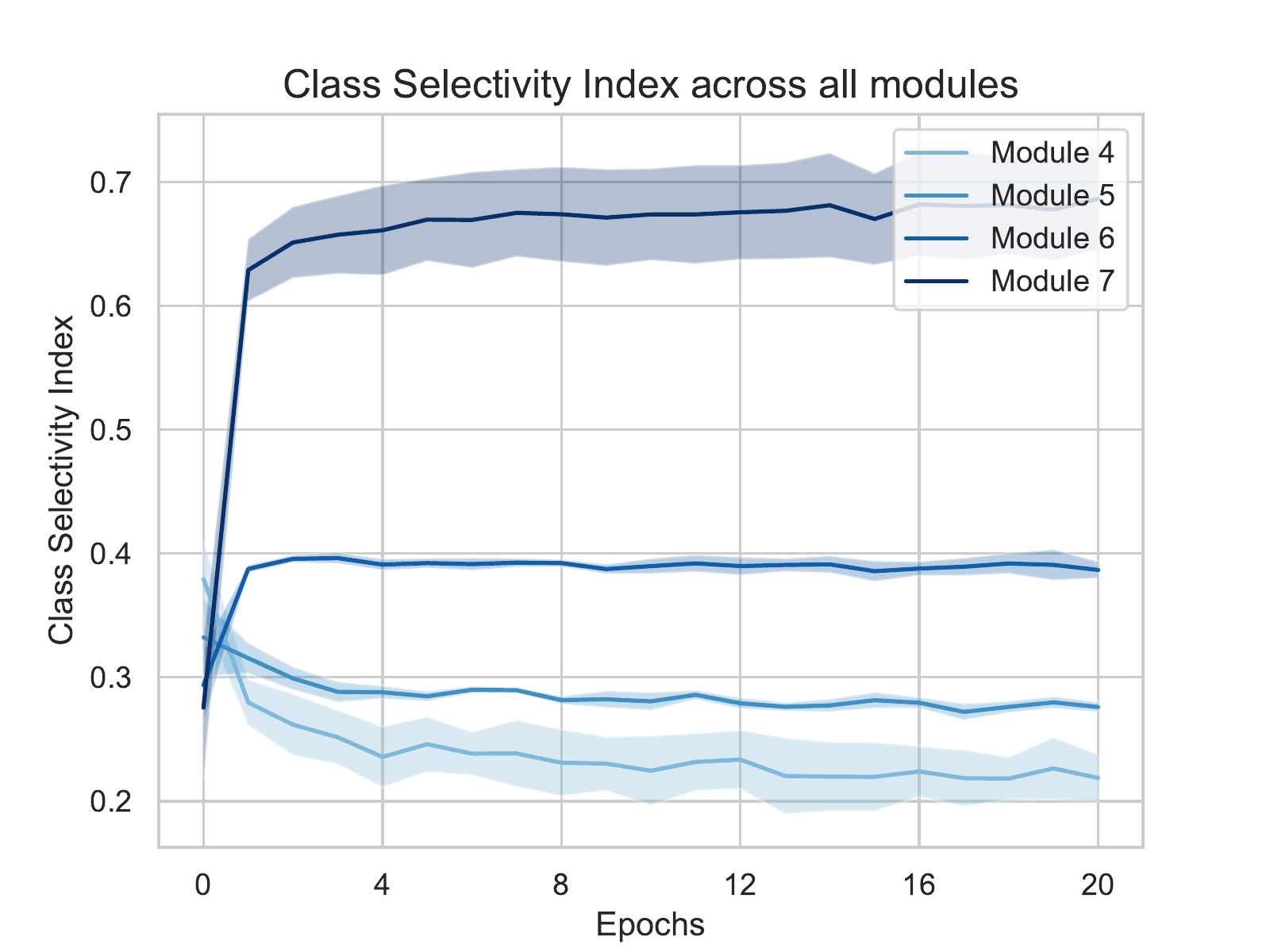}
    \caption{ResNet18}
    \label{rn18}
    \end{subfigure}
    \begin{subfigure}{0.45\linewidth}
    \includegraphics[width=\linewidth]{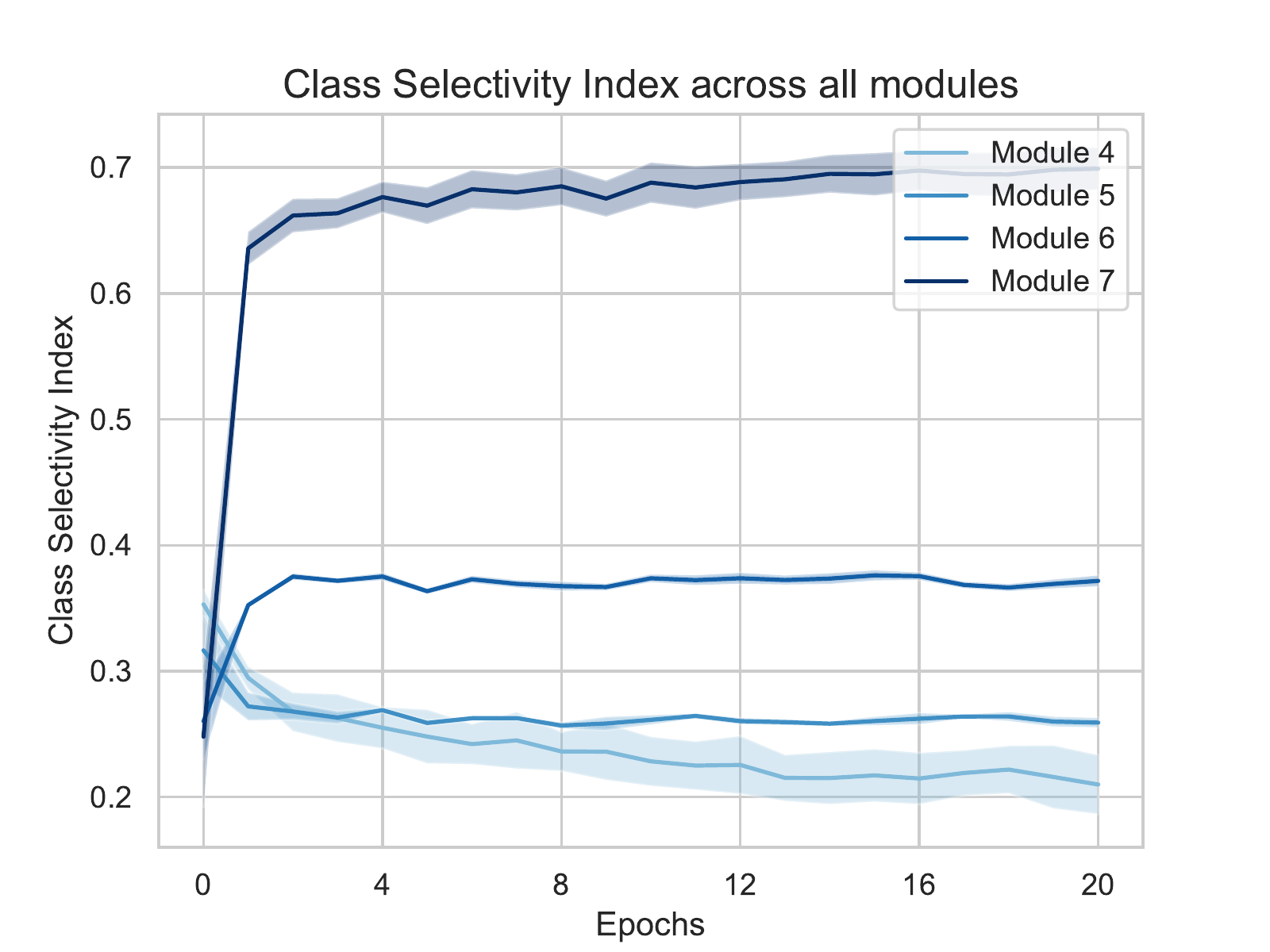}
    \caption{ResNet34}
    \label{rn34}
    \end{subfigure}
\caption{\textbf{Evolution of class selectivity throughout training in all modules of ResNet18 and ResNet34}  }
\label{rn_small}
\end{figure}

In both ResNet18 and ResNet34, the rise in selectivity in early epochs can only be observed for module 6 and 7 (Fig \ref{rn_small}). The selectivity values then stabilize across all modules just like what was observed in the case of ResNet50. Another interesting observation is that the stabilized class selectivity values for each module are approximately the same across all three ResNet architectures, i.e, for module 7 it is $\approx$ 0.7, for module 6 it is $\approx$ 0.4, for module 5 and 4 it is  $\approx$ between 0.2-0.3. 

The lower rise in selectivity could possibly be due to smaller depth (fewer layers) and also significantly fewer channels in each layer. For comparison, in a ResNet50 there are 256, 512, 1024, 2048 channels in each layer of module 4, 5, 6 and 7 respectively while in ResNet18/34 there are only 64, 128, 256, 512 channels in each layer of module 4, 5, 6 and 7 respectively. 

\subsubsection{Evolution of class selectivity in bottleneck layers}

\begin{figure}
\centering
 \begin{subfigure}{0.45\textwidth}
     \includegraphics[width=\textwidth]{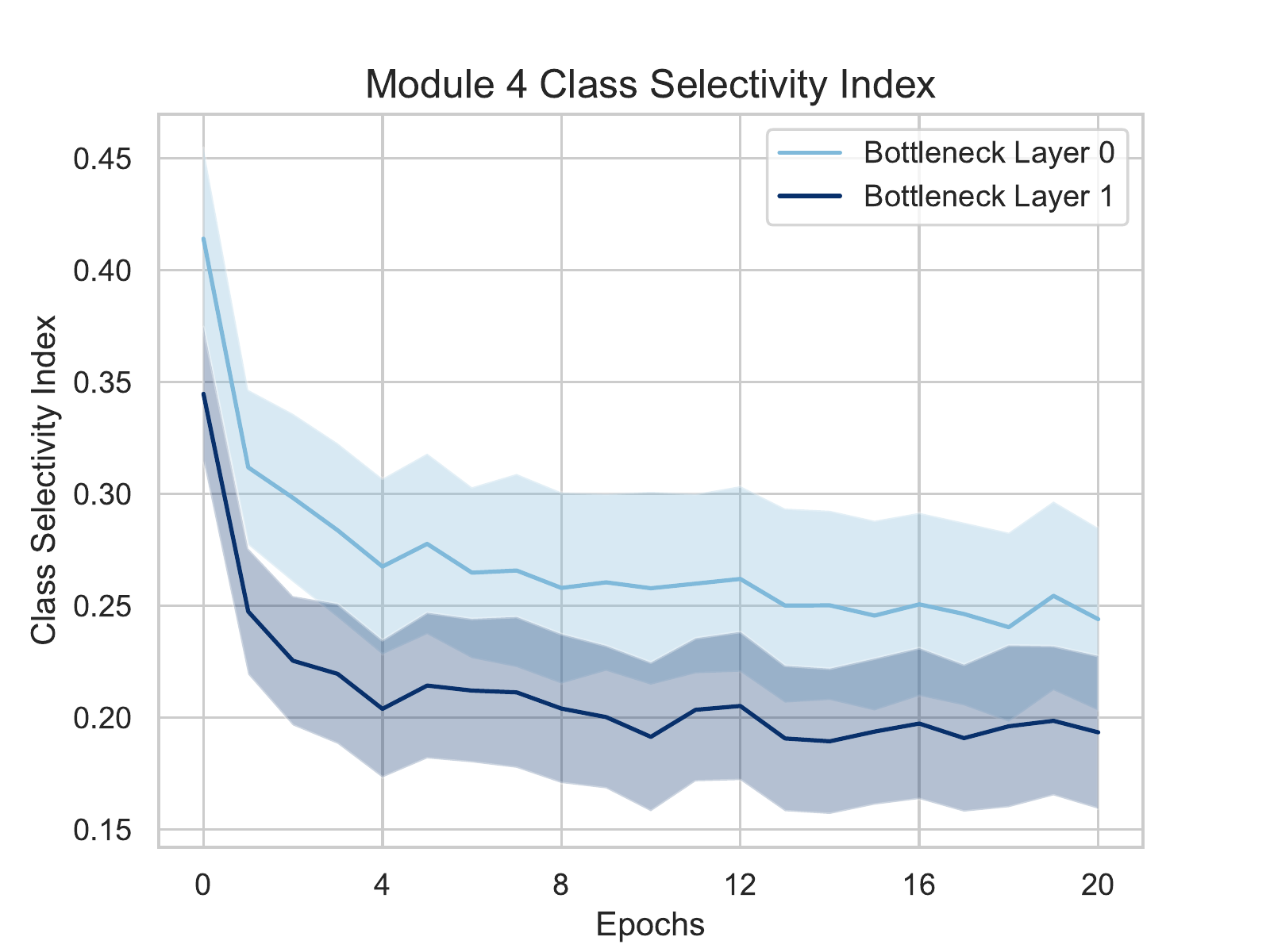}
 \end{subfigure}
 \begin{subfigure}{0.45\textwidth}
     \includegraphics[width=\textwidth]{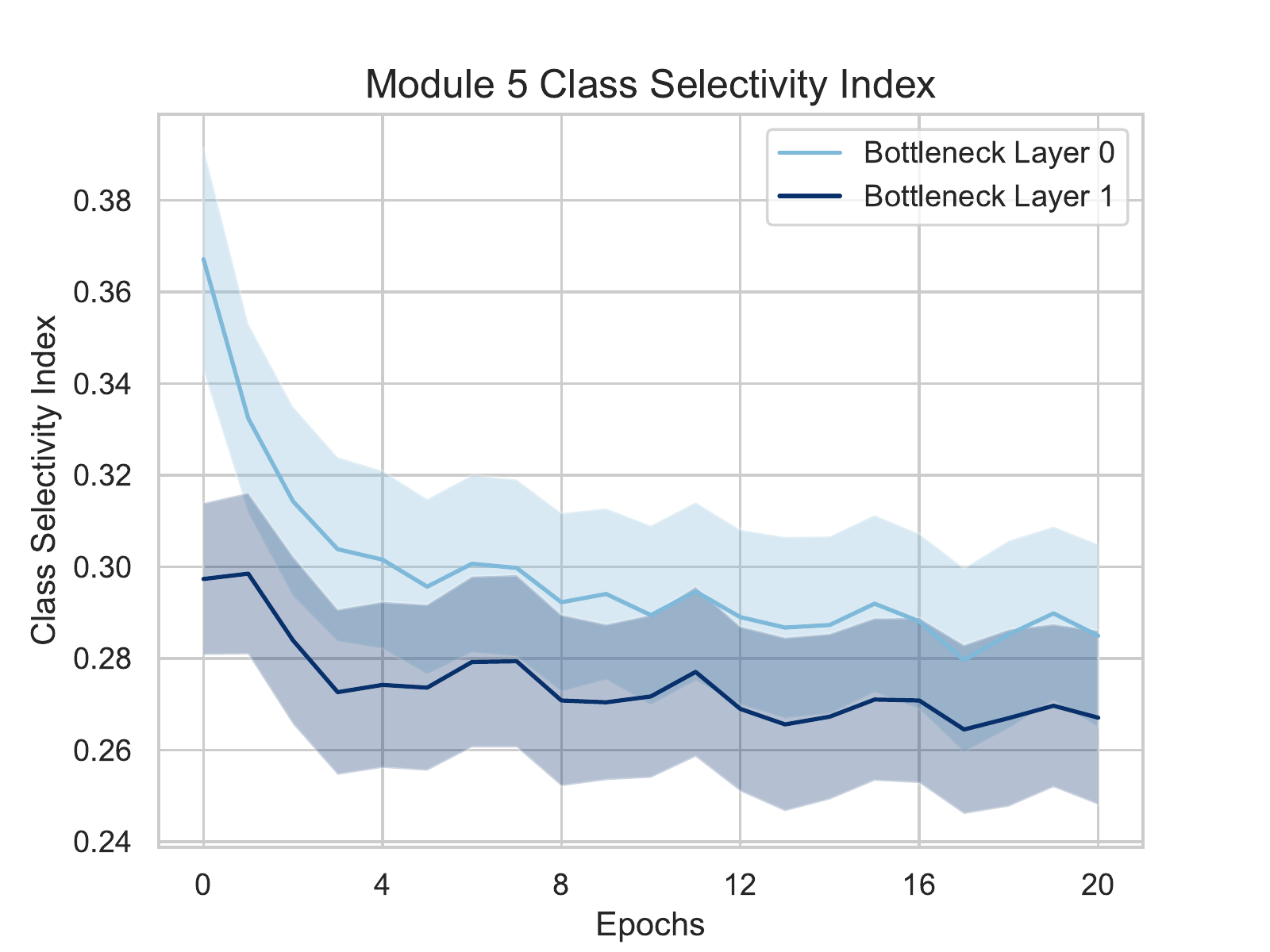}
 \end{subfigure}
 \medskip
 \begin{subfigure}{0.45\textwidth}
     \includegraphics[width=\textwidth]{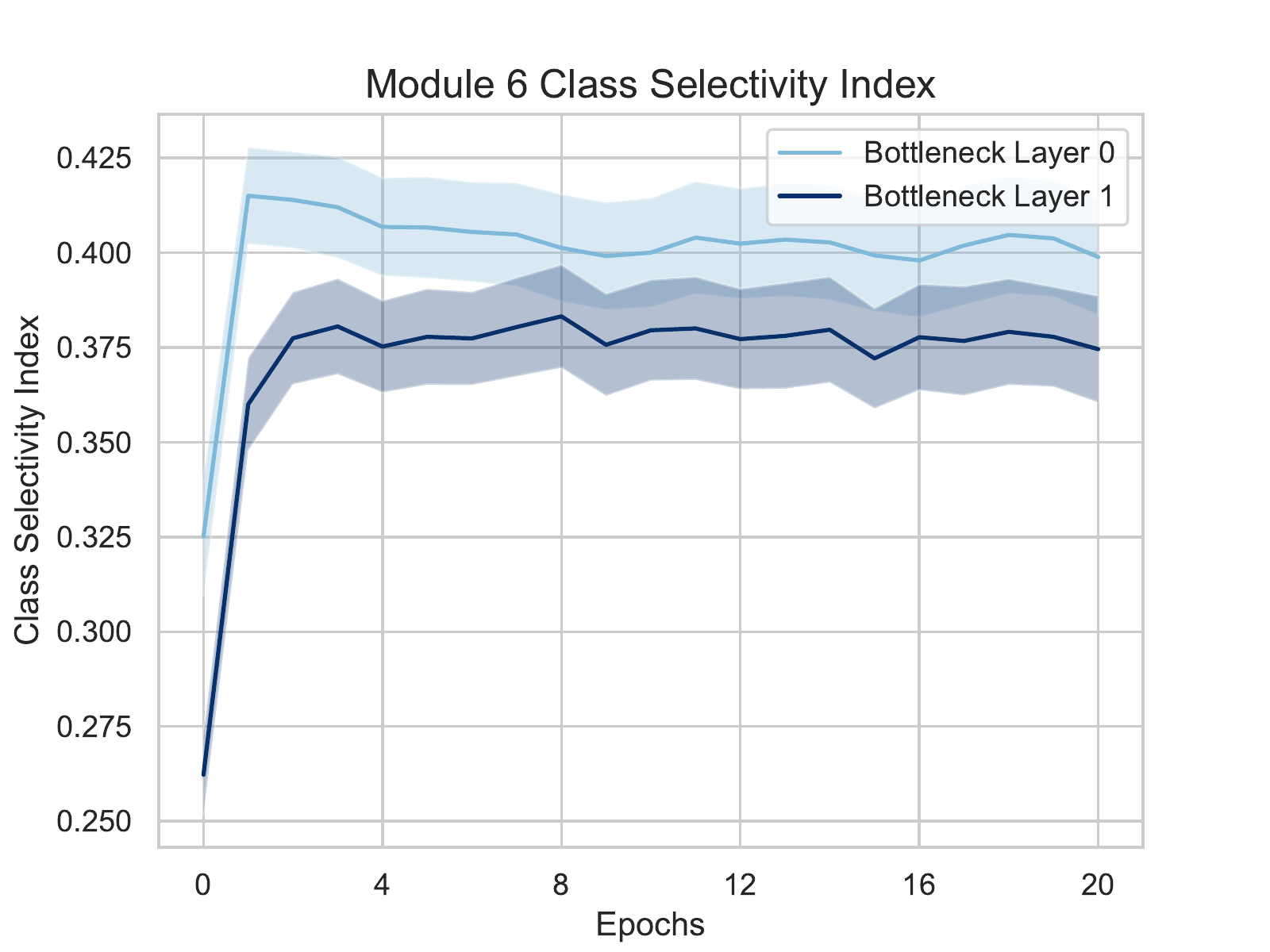}
 \end{subfigure}
 \begin{subfigure}{0.45\textwidth}
     \includegraphics[width=\textwidth]{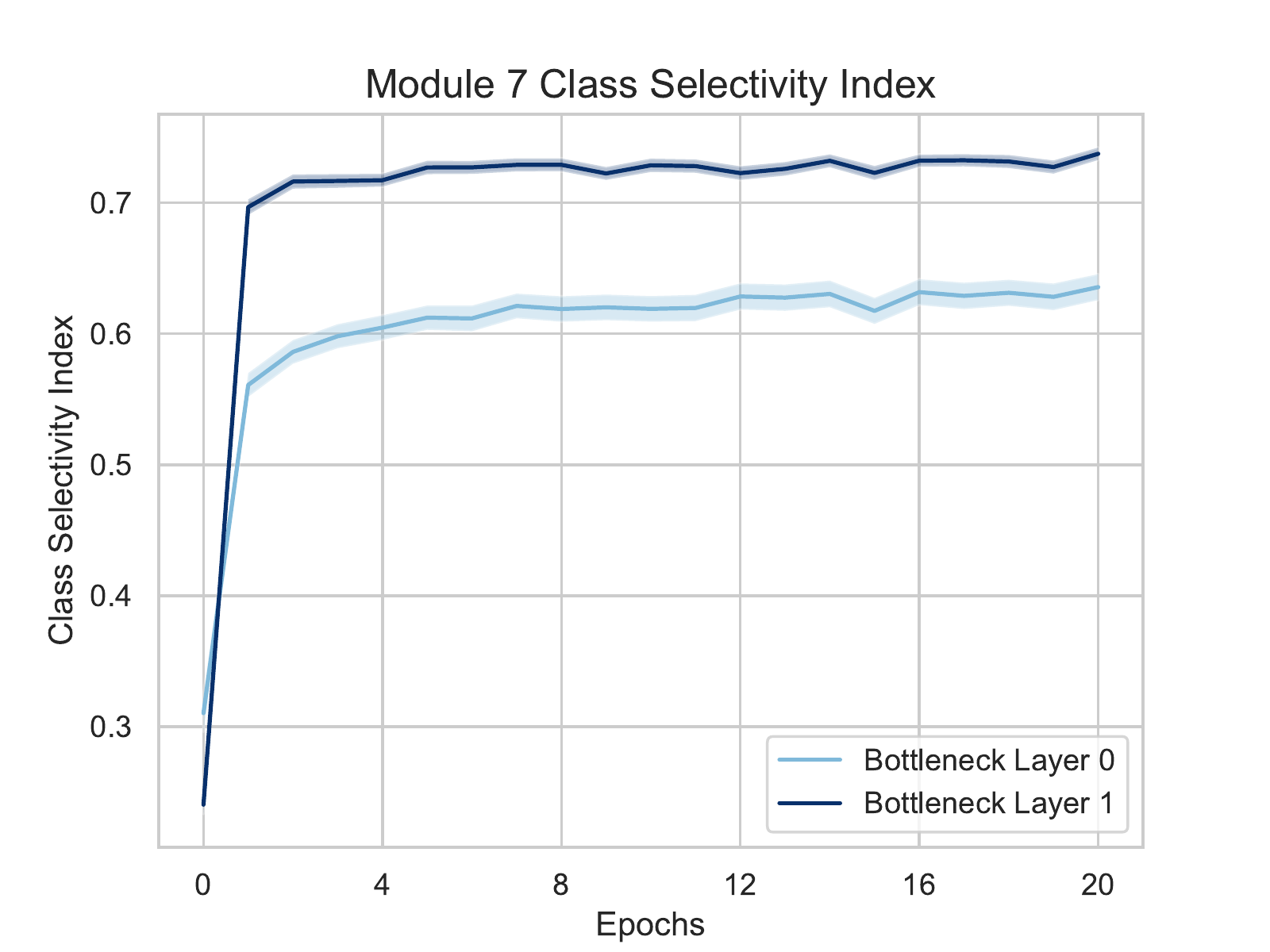}
 \end{subfigure}

 \caption{\textbf{Evolution of class selectivity throughout training across bottleneck layers for ResNet18}} 
 \label{rn18_bn}
\end{figure}

\begin{figure}
\centering
 \begin{subfigure}{0.45\textwidth}
     \includegraphics[width=\textwidth]{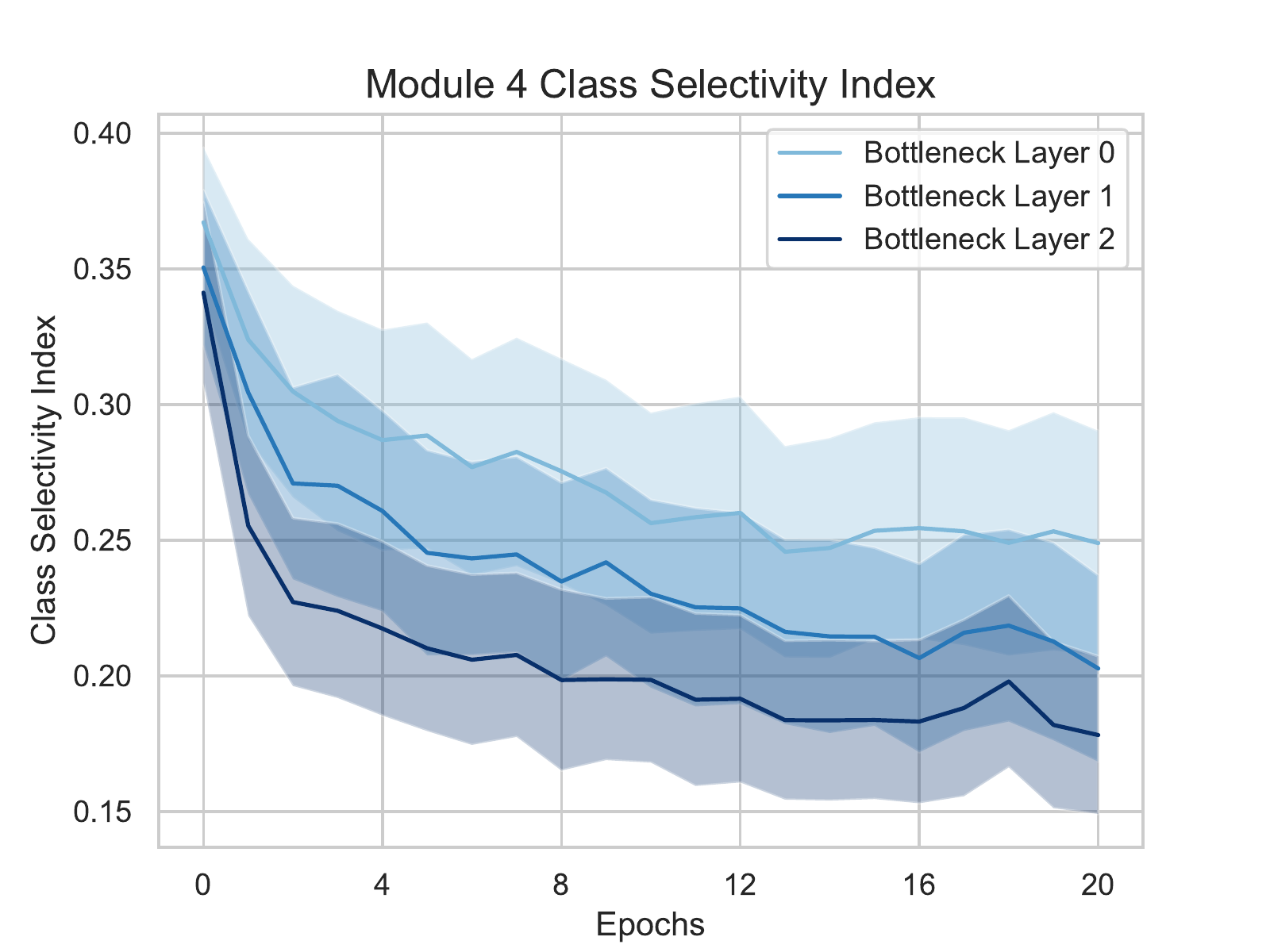}
 \end{subfigure}
 \begin{subfigure}{0.45\textwidth}
     \includegraphics[width=\textwidth]{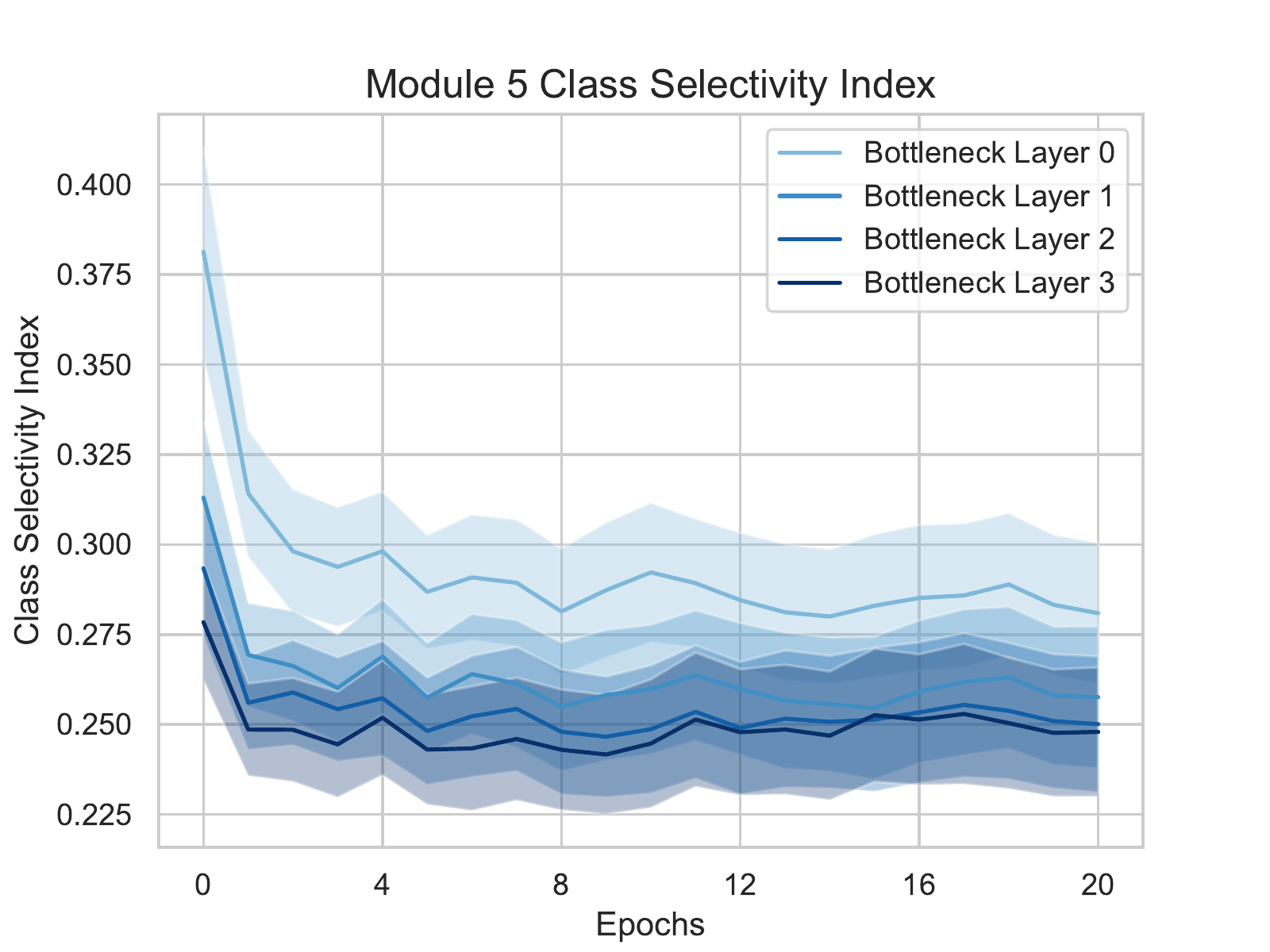}
 \end{subfigure}
 \medskip
 \begin{subfigure}{0.45\textwidth}
     \includegraphics[width=\textwidth]{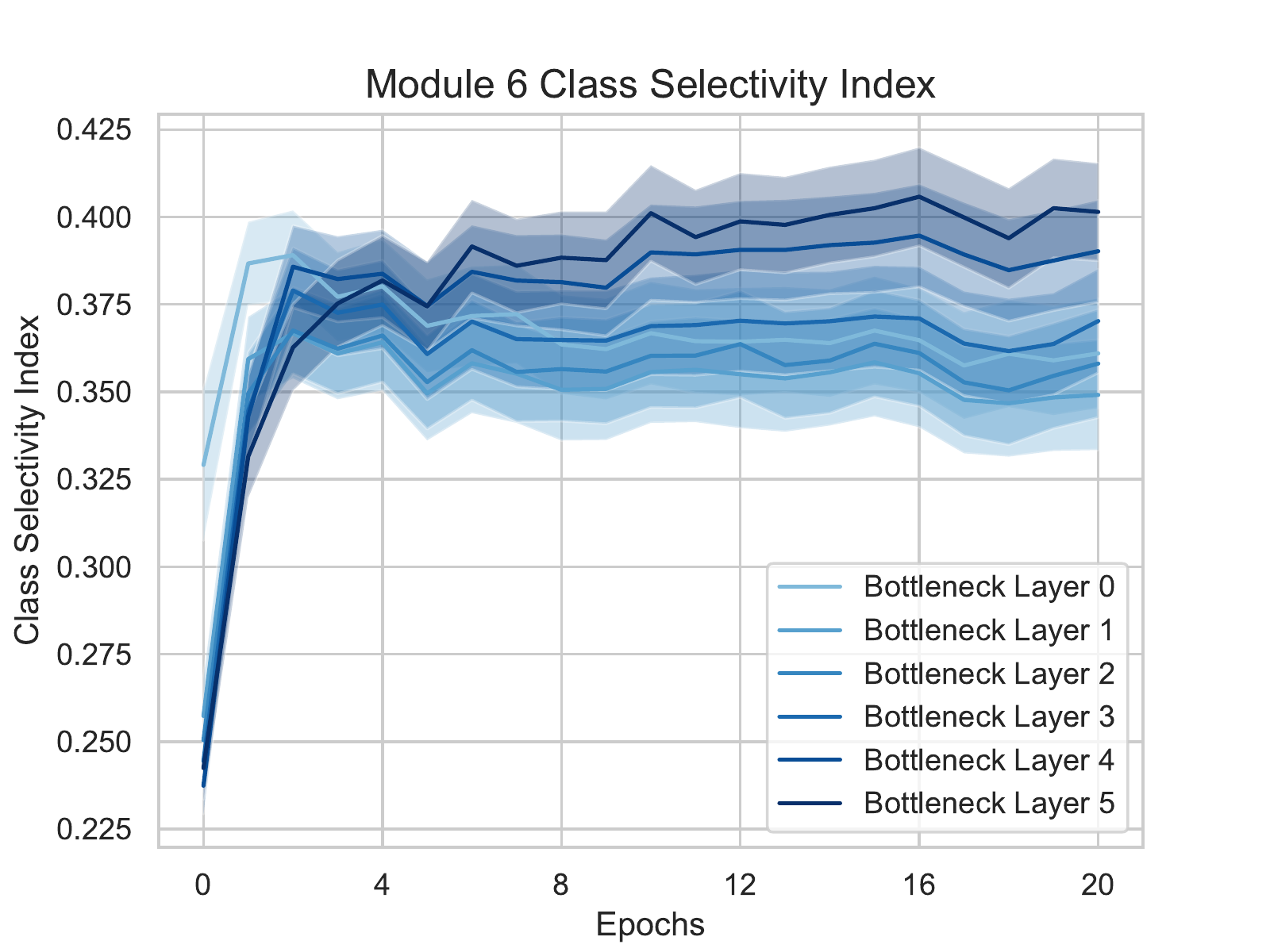}
 \end{subfigure}
 \begin{subfigure}{0.45\textwidth}
     \includegraphics[width=\textwidth]{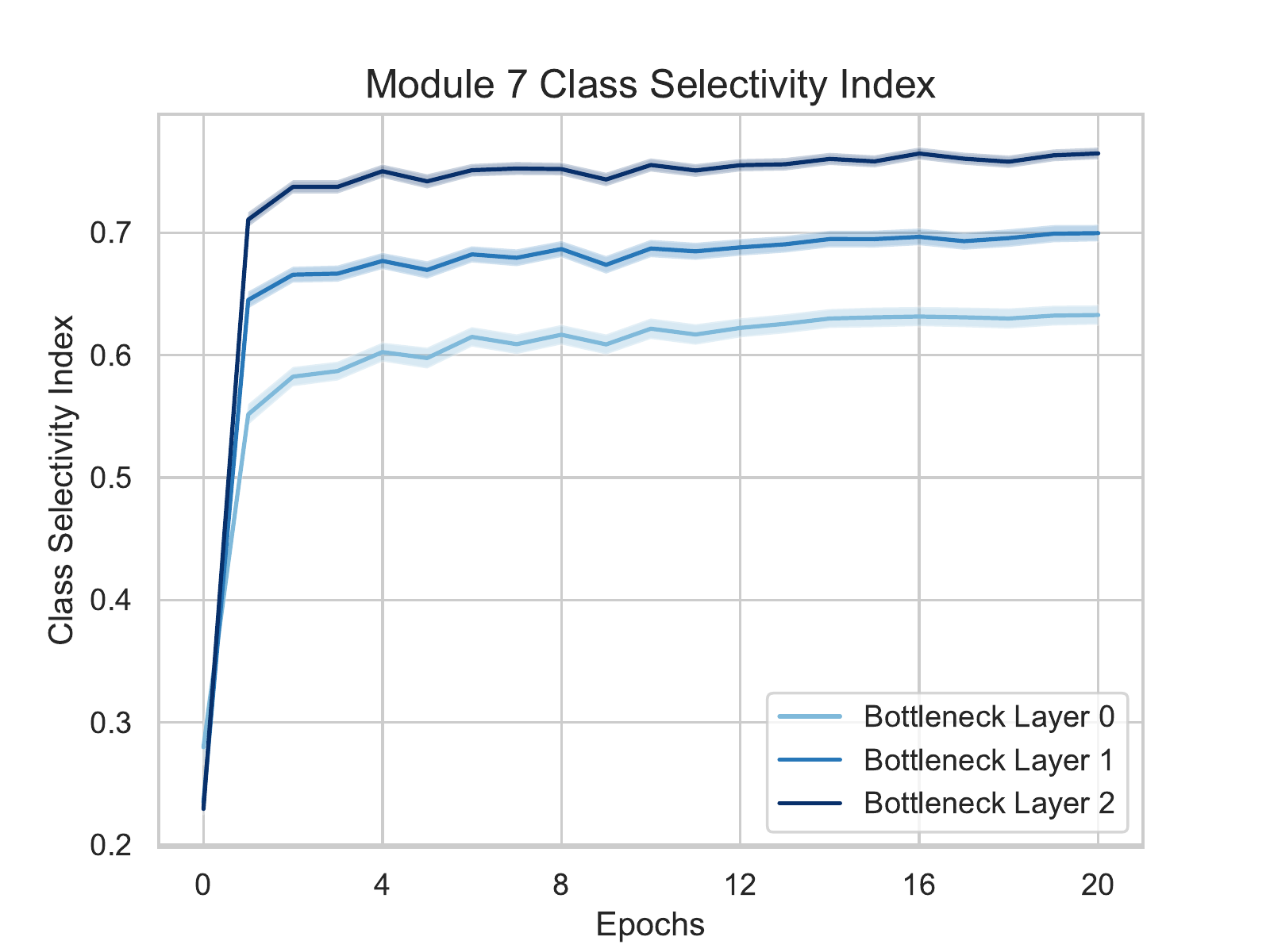}
 \end{subfigure}

 \caption{\textbf{Evolution of class selectivity throughout training across bottleneck layers for ResNet34}} 
 \label{rn34_bn}
\end{figure}

The evolution of selectivity in bottleneck layers of ResNet18 and ResNet34 (Fig \ref{rn18_bn}, Fig \ref{rn34_bn}) show a similar trend to that of ResNet50 bottleneck layers (Fig \ref{soc_bot}). For ResNet34, modules 6 and 7 are increasingly class-selective with network depth and modules 4 and 5 are decreasingly class-selective with network depth (Fig \ref{rn34_bn}). The same effect can be observed for ResNet18 except for module 6 which is decreasingly class-selective with network depth (Fig \ref{rn18_bn}). 

\subsubsection{Regularizing against selectivity}\label{rn18_34_reg}
\begin{figure}
\centering
\begin{subfigure}{0.45\textwidth}
     \includegraphics[width=\textwidth]{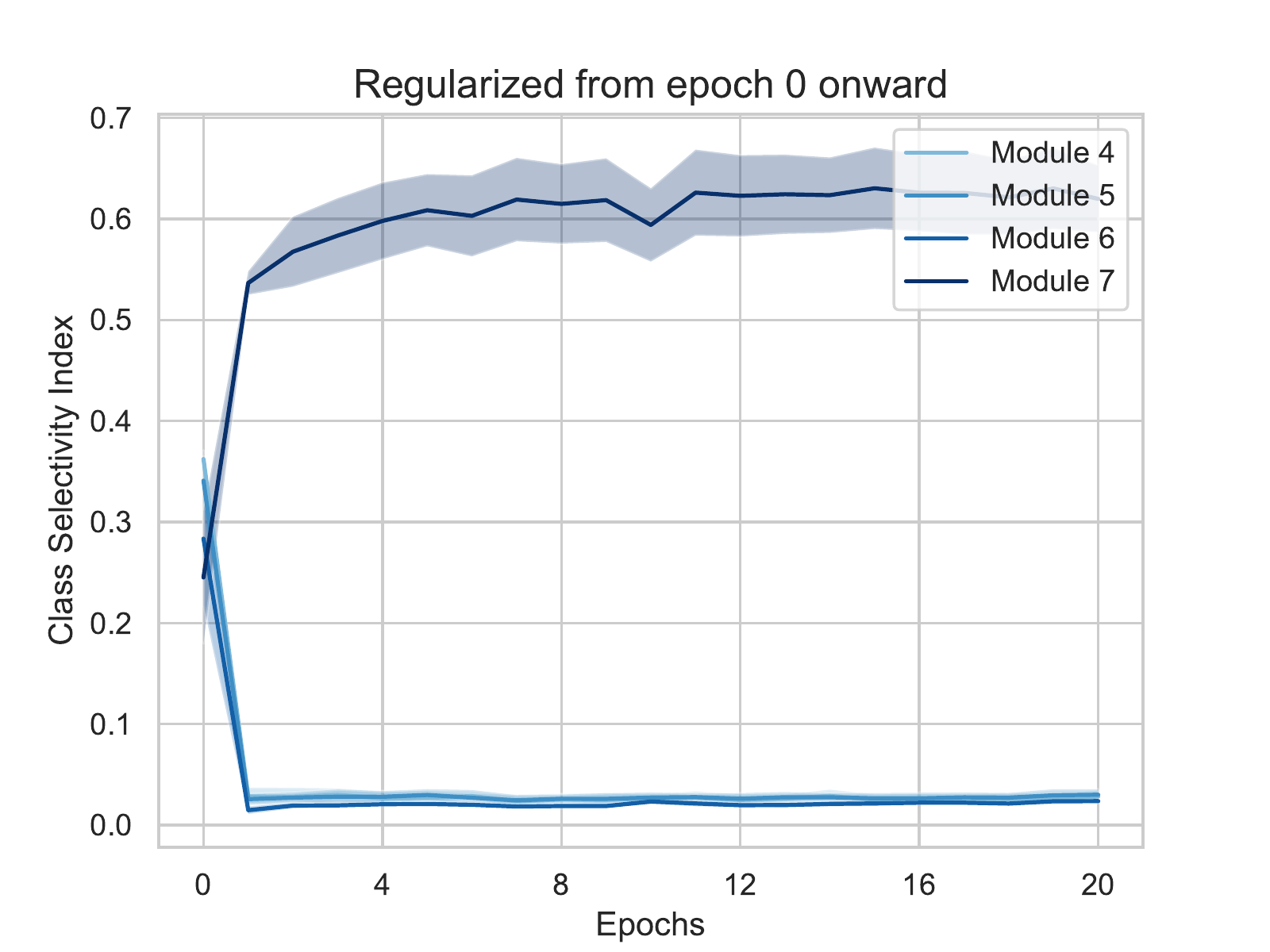}
 \end{subfigure}
 \begin{subfigure}{0.45\textwidth}
     \includegraphics[width=\textwidth]{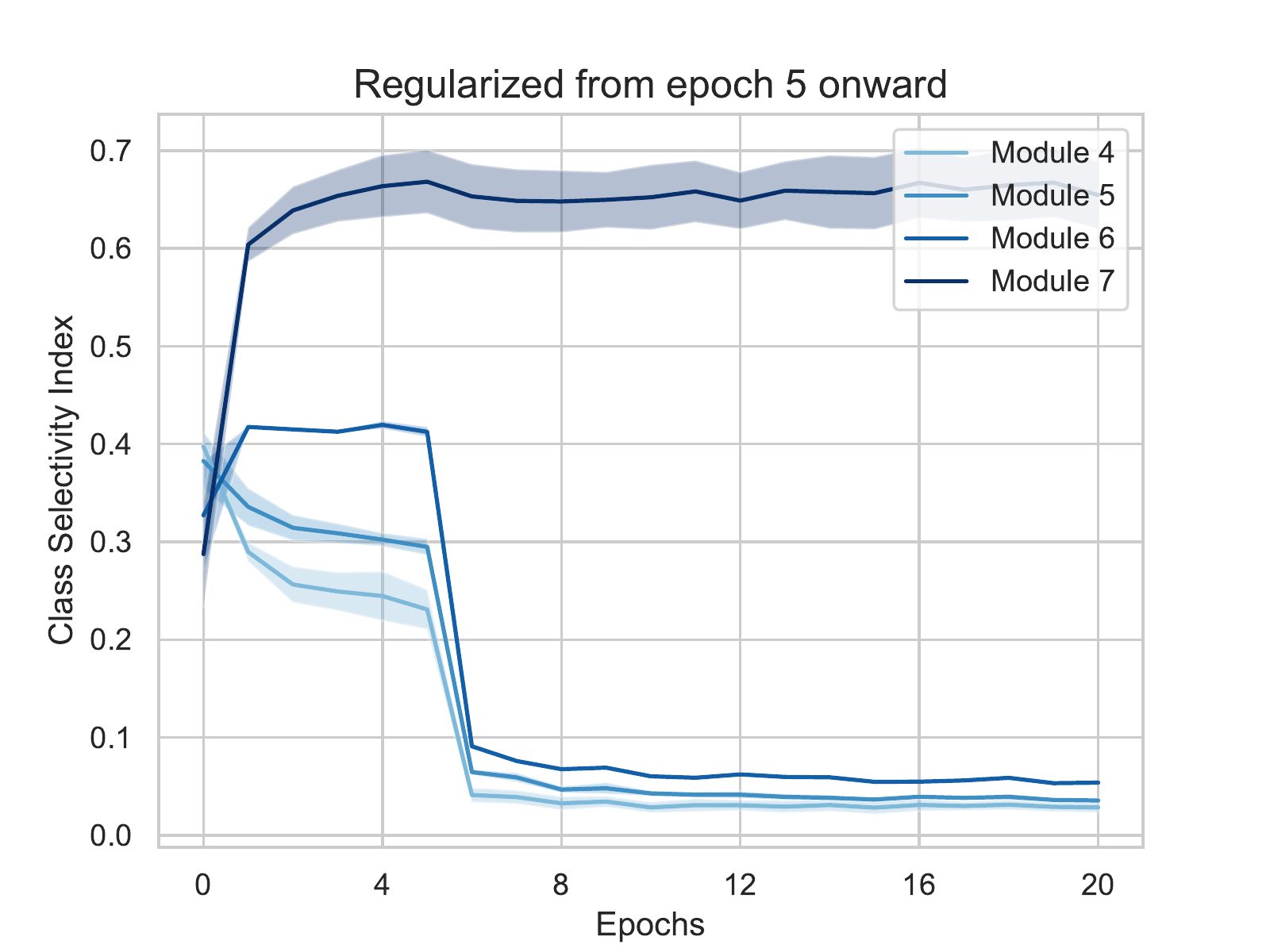}
 \end{subfigure}
 \begin{subfigure}{0.45\textwidth}
     \includegraphics[width=\textwidth]{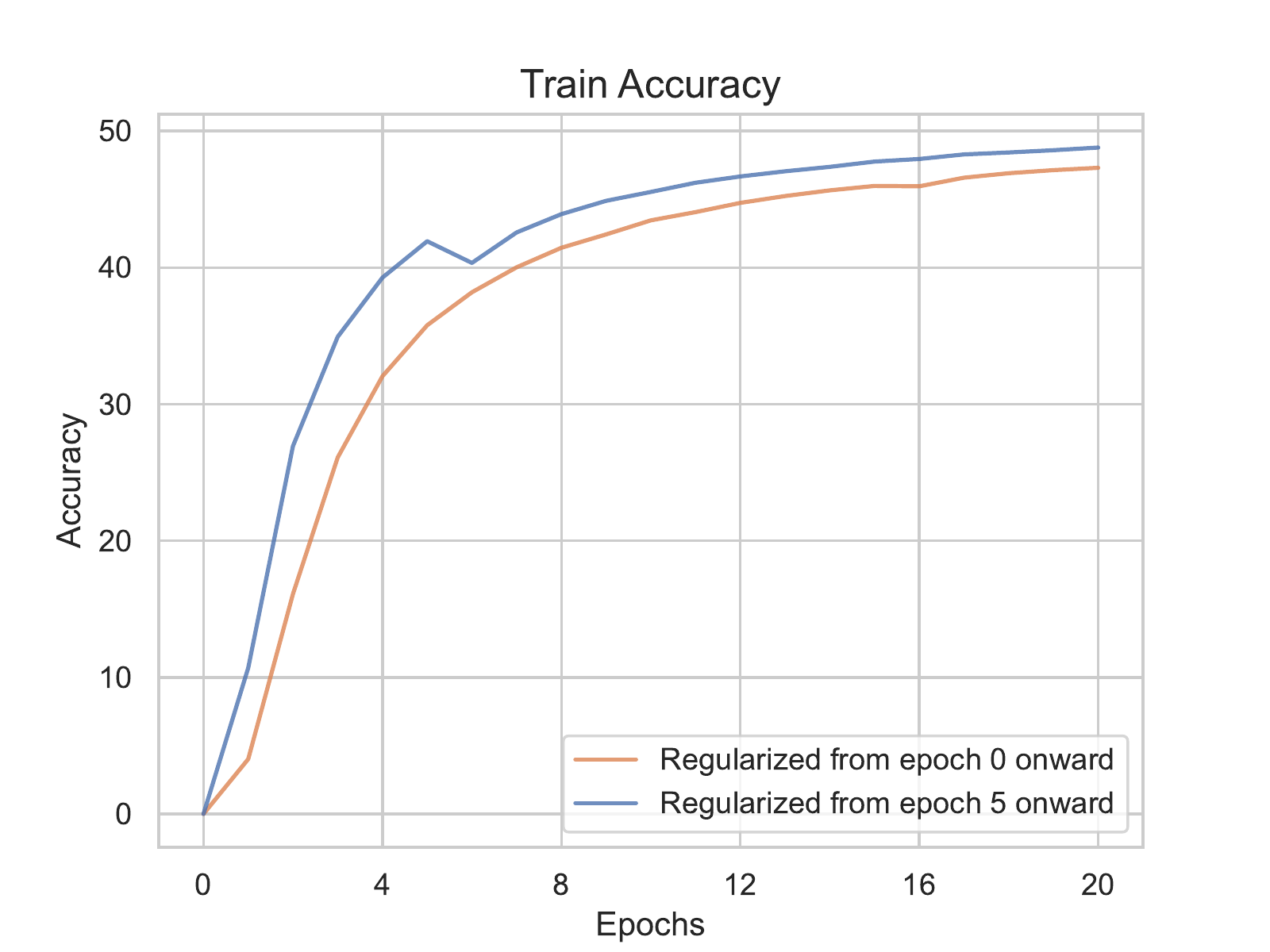}
\end{subfigure}
\medskip
\begin{subfigure}{0.45\textwidth}
     \includegraphics[width=\textwidth]{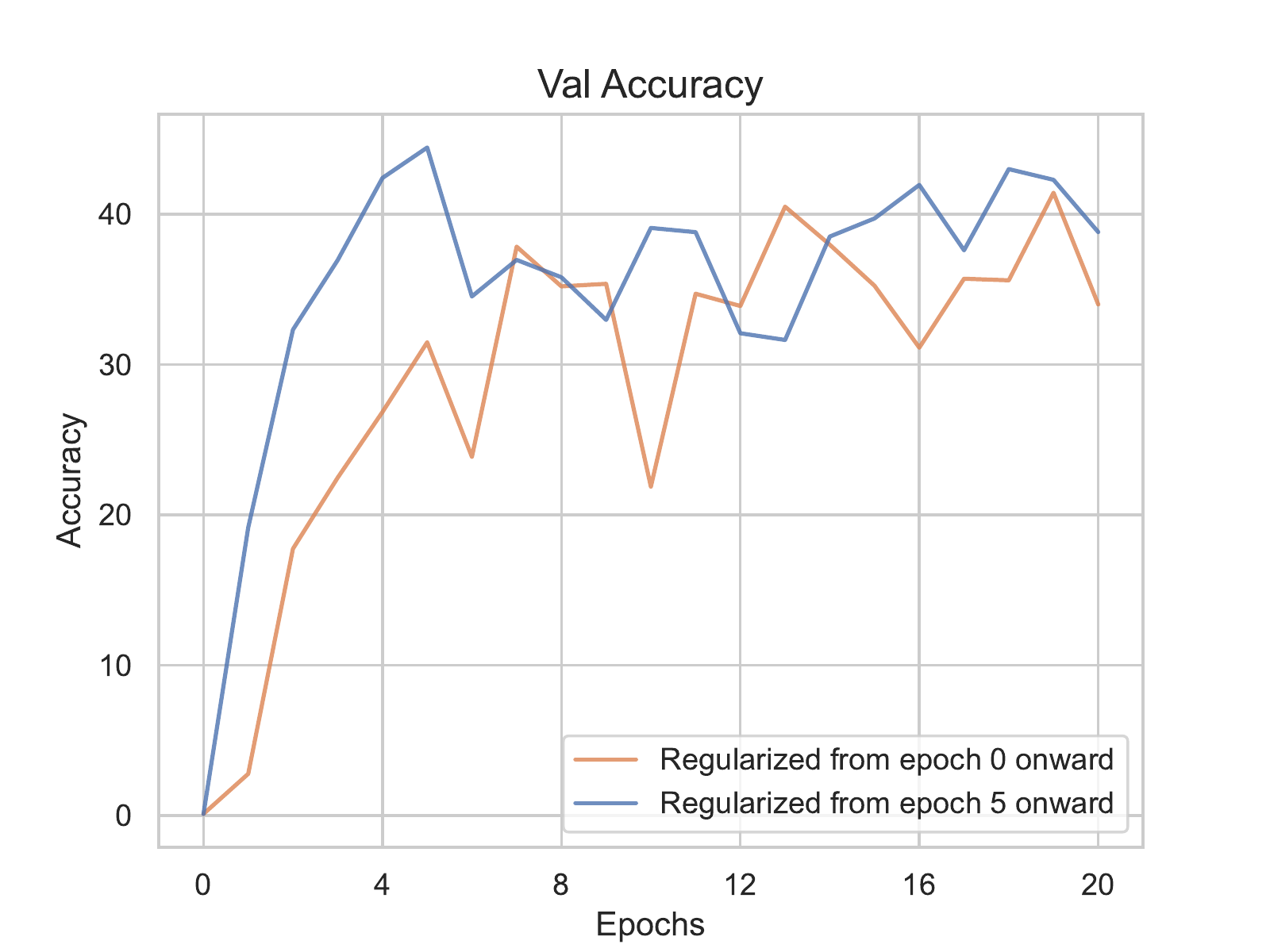}
 \end{subfigure}
 
 \caption{\textbf{Regularization against selectivity for ResNet18.} The regularization result observed for ResNet50 holds true for ResNet18. $\alpha$=-20 was used to regularize the models. The model regularized from epoch 0 onward performs worse than the model regularized from epoch 5 onward.} 
 \label{rn18_reg}
\end{figure}

\begin{figure}
\centering
\begin{subfigure}{0.45\textwidth}
     \includegraphics[width=\textwidth]{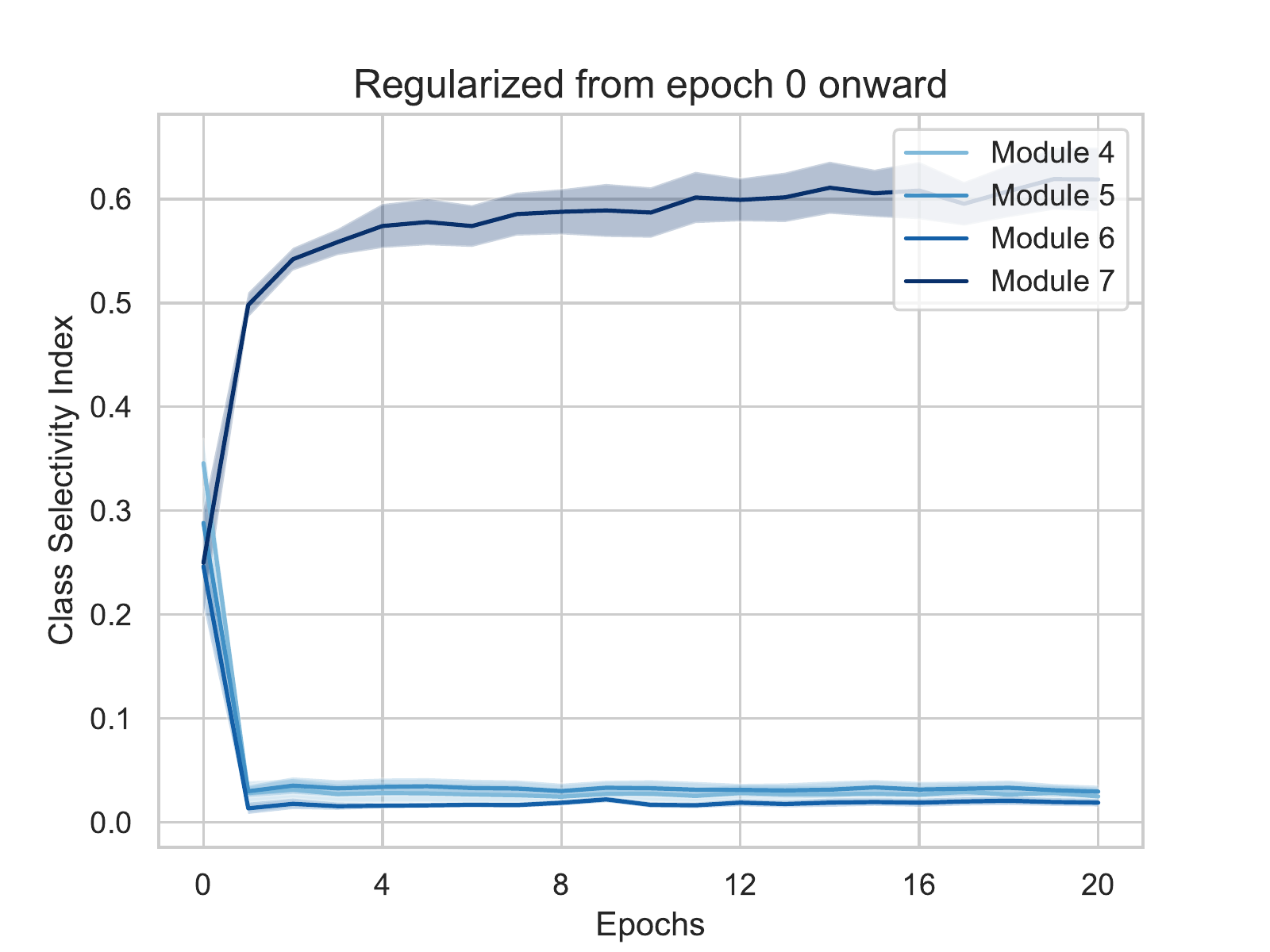}
 \end{subfigure}
 \begin{subfigure}{0.45\textwidth}
     \includegraphics[width=\textwidth]{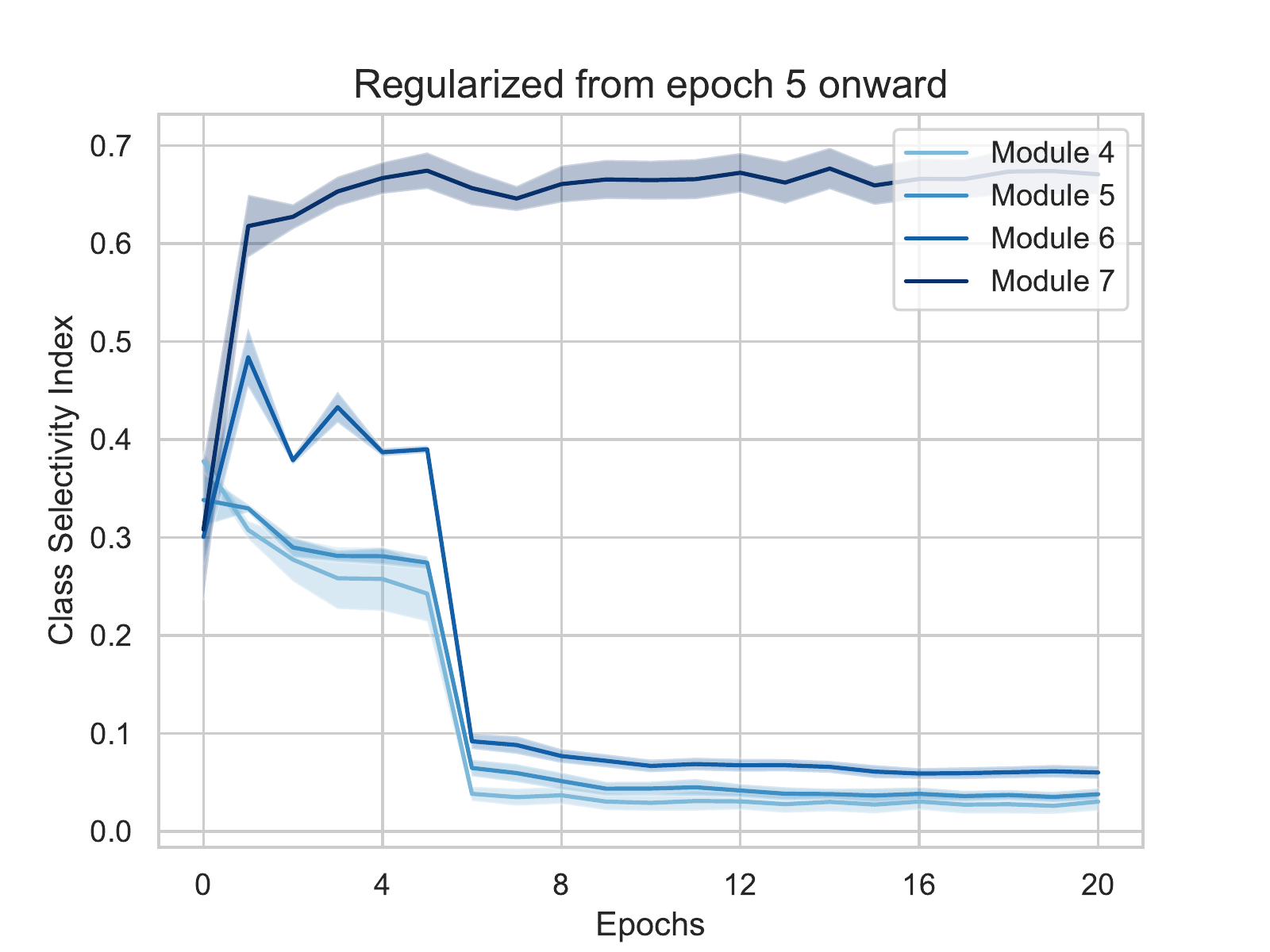}
 \end{subfigure}
 \begin{subfigure}{0.45\textwidth}
     \includegraphics[width=\textwidth]{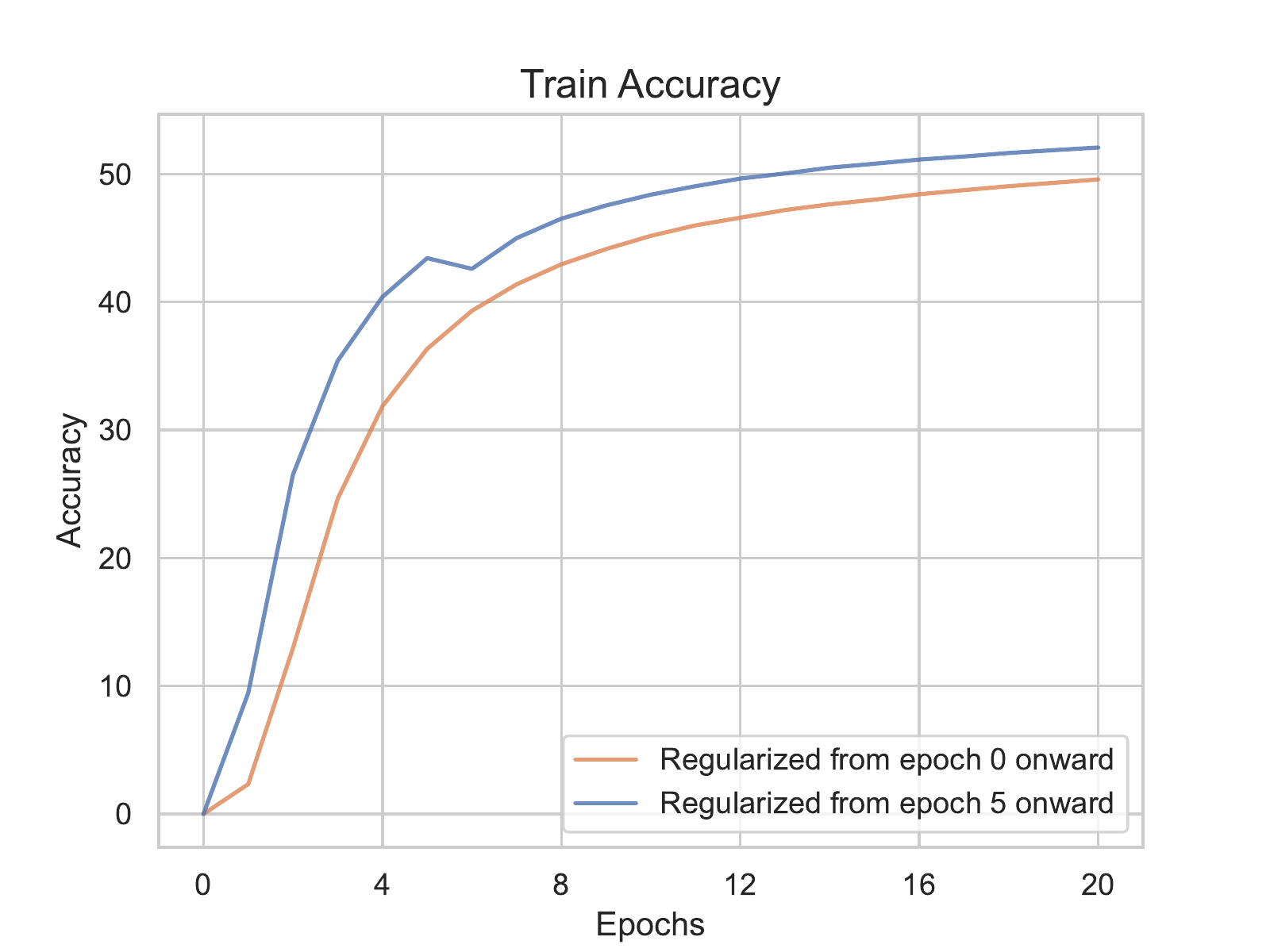}
\end{subfigure}
\medskip
\begin{subfigure}{0.45\textwidth}
     \includegraphics[width=\textwidth]{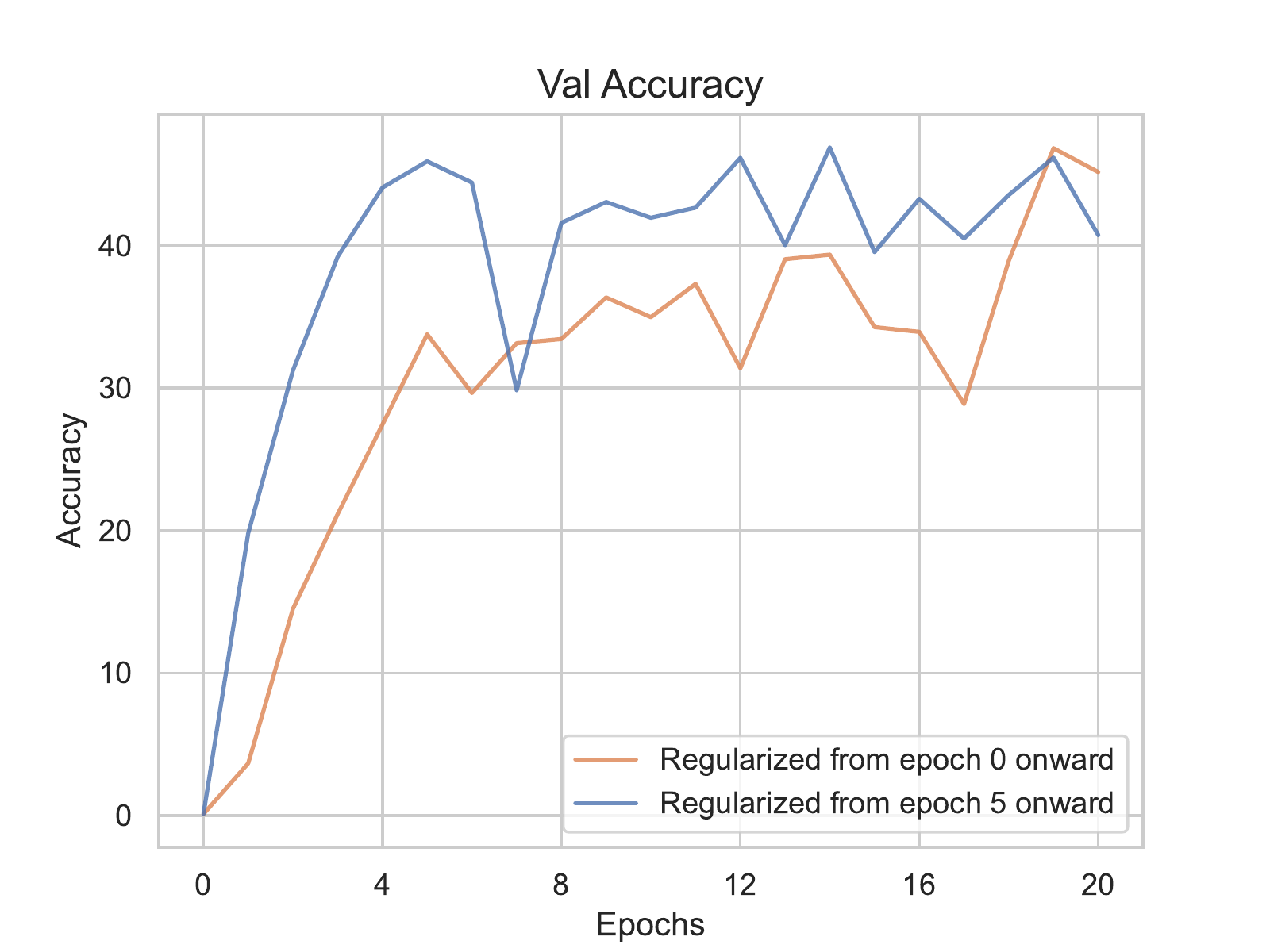}
 \end{subfigure}
 
 \caption{\textbf{Regularization against selectivity for ResNet34.} The regularization result observed for ResNet50 also holds true for ResNet34. $\alpha$=-20 was used to regularize the models. The model regularized from epoch 0 onward performs worse than the model regularized from epoch 5 onward.} 
 \label{rn34_reg}
\end{figure}

Both ResNet18 and ResNet34 were regularized against selectivity with $\alpha=-20$. Interestingly, both the models show a similar regularization result to that of ResNet50. For both ResNet18 and ResNet34, the models regularized from epoch 0 onward performed worse than the models regularized from epoch 5 onward (Fig \ref{rn18_reg}, \ref{rn34_reg}). This indicates that class selectivity is important during the early epochs of training even in the case of smaller variants of ResNet.

\end{document}

%% file: math_commands.tex
\usepackage{amsmath,amsfonts,bm}

\def\eqref#1{equation~\ref{#1}}
\def\1{\bm{1}}

\DeclareMathAlphabet{\mathsfit}{\encodingdefault}{\sfdefault}{m}{sl}
\SetMathAlphabet{\mathsfit}{bold}{\encodingdefault}{\sfdefault}{bx}{n}